\documentclass[twoside]{article}

\usepackage{microtype}
\usepackage{graphicx}
\usepackage{subfigure}
\usepackage{booktabs} 

\usepackage{amsmath,amsfonts,latexsym}
\usepackage{enumitem,multirow,supertabular,amssymb,setspace,tabularx ,booktabs,inputenc,comment,booktabs
	,multicol}
\usepackage{lipsum}
\usepackage{parskip}
\usepackage{xspace}
\usepackage{epsfig}
\usepackage{epstopdf}
\usepackage{setspace}
\usepackage{amsthm}
\usepackage{comment}
\usepackage{arydshln}
\usepackage{wrapfig}

\graphicspath{
	{../../src/images/}
	{../../src_mpi/}
	{../../src_mpi/logs/}
	{./images/},
}

\newcommand{\R}{\mathbb{R}}
\newcommand{\tnorm}[1]{\lVert{#1}\rVert^2}
\newcommand{\norm}[1]{\lVert{#1}\rVert}

\DeclareMathOperator{\E}{\mathbb{E}}

\usepackage{algorithm}
\usepackage{algorithmic}
\newtheorem{theorem}{Theorem}
\newtheorem{assumption}{Assumption}
\newtheorem{definition}{Definition}
\newtheorem{lemma}{Lemma}
\newtheorem{corollary}{Corollary}

\usepackage[colorinlistoftodos,prependcaption,textsize=tiny]{todonotes}

\usepackage[colorlinks]{hyperref}
\newcommand\myshade{85}
\colorlet{mylinkcolor}{red}
\colorlet{mycitecolor}{blue}
\colorlet{myurlcolor}{red}

\hypersetup{
  linkcolor  = mylinkcolor!\myshade!black,
  citecolor  = mycitecolor!\myshade!black,
  urlcolor   = myurlcolor!\myshade!black,
  colorlinks = true,
}

\usepackage[accepted]{aistats2020}
\begin{document}

%
\runningtitle{Efficient Distributed Hessian Free Algorithm for Large-scale ERM via Accumulating Sample Strategy}

%

\runningauthor{M. Jahani, X. He, C. Ma, A. Mokhtari, D. Mudigere, A. Ribeiro, and M. Tak{\'a}{\v{c}}ˇ}

\twocolumn[

\aistatstitle{Efficient Distributed Hessian Free Algorithm for Large-scale Empirical Risk Minimization via Accumulating Sample Strategy}

\aistatsauthor{ 
Majid Jahani\\Lehigh University   
\And Xi He\\Lehigh University 
\And  Chenxin Ma\\Lehigh University 
\And Aryan Mokhtari\\UT Austin  
\AND Dheevatsa Mudigere\\Facebook 
\And Alejandro Ribeiro\\University of Pennsylvania 
\And Martin Tak{\'a}{\v{c}}ˇ\\ Lehigh University
}

\aistatsaddress{ } ]


\begin{abstract}
In this paper, we propose a Distributed Accumulated Newton Conjugate gradiEnt (DANCE) method in which sample size is gradually increasing to quickly obtain a solution whose empirical loss is under satisfactory statistical accuracy. Our proposed method is multistage in which the solution of a stage serves as a warm start for the next stage which contains more samples (including the samples in the previous stage). The proposed multistage algorithm reduces the number of passes over data to achieve the statistical accuracy of the full training set. Moreover, our algorithm in nature is easy to be distributed and shares the strong scaling property indicating that acceleration is always expected by using more computing nodes. Various iteration complexity results regarding descent direction computation, communication efficiency and stopping criteria are analyzed under convex setting. Our numerical results illustrate that the proposed method outperforms other comparable methods for solving learning problems including neural networks.
\end{abstract}

\section{Introduction}
In the field of machine learning, solving the expected risk minimization problem has received lots of attentions over the last decades, which is in the form of
\begin{equation}\label{eq:ExpLoss}
	\min_{w\in \R^d}L(w)=\min_{w\in \R^d}\E_z[f(w,z)],
\end{equation}
where $z$ is a $d+1$ dimensional random variable containing both feature variables and a response variable. $f(w,z)$ is a loss function with respect to $w$ and any fixed value of $z$.

In most practical problems, the distribution of $z$ is either unknown or leading great difficulties evaluating the expected loss. One general idea is to estimate the expectation with a statistical average over a large number of independent and identically distributed data samples of $z$, denoted by $\{z_1,z_2,\dots, z_N\}$ where $N$ is the total number of samples. Thus, the problem in \eqref{eq:ExpLoss} can be rewritten as the Empirical Risk Minimization (ERM) problem\vspace{-10pt}
\begin{equation}\label{eq:ERM}
	\min_{w\in \R^d}L_N(w)=\min_{w\in \R^d}\frac{1}{N} \sum_{i=1}^{N}f_i(w),\vspace{-10pt}
\end{equation}
where $f_i(w) = f(w,z_i)$.

Many studies have been done on developing optimization algorithms to find an optimal solution of the above problem under different setting. For example, the studies by \cite{Beck09fista,nesterov2013introductory,drusvyatskiy2016optimal,ma2017underestimate} are some of the gradient-based methods which require at least one pass over all data samples to evaluate the gradient $\nabla L_N(w)$. As the sample size $N$ becomes larger, these methods would be less efficient compared to stochastic gradient methods where the gradient is approximated based on a small number of samples \cite{Johnson13SVRG,roux2012stochastic,Defazio14saga,ShalevShwartz12sdca,Konecny17s2gd,nguyen2017sarah}.

Second order methods are well known to share faster convergence rate by utilizing the Hessian information. Recently, several papers by \cite{Byrd15quasi,berahassampled,jahani2019scaling, berahas2016multi,schraudolph2007stochastic,berahas2019quasi, mokhtari2015global,   Roosta-Khorasani2018} have studied how to apply second orders methods to solve ERM problem. However, evaluating the Hessian inverse or its approximation is always computationally costly, leading to a significant difficulty on applying these methods on large-scale problems.

The above difficulty can be addressed by applying the idea of adaptive sample size methods by recent works of \cite{Mokhtari_firstOredr,Eisen17,Mokhtari_AdaNewton16}, which is based on the following two facts. First, the empirical risk and the statistical loss have different minimizers, and it is not necessary to go further than the difference between the mentioned two objectives, which is called \textit{statistical accuracy}. More importantly, if we increase the size of the samples in the ERM problem the solutions should not significantly change as samples are drawn from a fixed but unknown probability distribution. The key idea of adaptive samples size methods is to solve an ERM problem with a small number of samples upto its statistical accuracy and use the obtained solution as a warm start for the next ERM problem which contains more samples. In particular, \cite{Mokhtari_AdaNewton16} reduced the complexity of Newton's method by incorporating the adaptive sample size idea; however, their approach still requires computing $\log N$ Hessian inversions which is costly when the problem dimension $d$ is large.  In order to decrease the cost of computing the Hessian inverse, \cite{Eisen17} proposed the $k$-Truncated Adaptive Newton ($k$-TAN) approach in which the inverse of Hessian is approximated by truncating the $k$ largest eigenvalues of the Hessian. The cost per iteration of this approach is $\mathcal{O}((\log k+n )d^2)$ which may not be satisfactory either when $d$ is large or  $k$ is close to $d$.

%

\begin{table}[] 
	\centering
	\caption{Comparison of computational complexity between different algorithms for convex functions}
	\label{Tab1}
	\begin{small}
		\begin{tabular}{@{}ll@{}}\toprule
			{\bf Method} & {\bf Complexity\quad}
			\\ \midrule
			{\bf AdaNewton} & $\mathcal{O}(2Nd^2 +d^3 \log_2(N))$
			\\ \hdashline  
			{\bf $k$-TAN} & $\mathcal{O}(2Nd^2 +d^2 \log_2(N)\log k \, )$
			\\ \hdashline
			{\bf DANCE } & $\tilde{\mathcal{O}}((\log_2(N))^3N^{1/4}d^2)$
			\\		
			\bottomrule 
		\end{tabular}
	\end{small}
	\vspace{-5mm}
\end{table} 

In this paper, we propose an increasing sample size second-order method which solves the Newton step in ERM problems more efficiently. Our proposed algorithm, called Distributed Accumulated Newton Conjugate gradiEnt (DANCE), starts with a small number of samples and minimizes their corresponding ERM problem. This subproblem is solved up to a specific accuracy, and the solution of this stage is used as a warm start for the next stage. We increase the number of samples in the next stage which contains all the previous samples, and use the previous solution as a warm start for solving the new ERM.
Such procedure is run iteratively until either all the samples have been included, or we find that it is unnecessary to further increase the sample size.
Our DANCE method combines the idea of increasing sample size and the inexact damped Newton method discussed in the works of \cite{Disco15} and \cite{ma2016distributed}. Instead of solving the Newton system directly, we apply preconditioned conjugate gradient (PCG) method as the solver for each Newton step.
Also, it is always a challenging problem to run first order algorithms such as SGD and Adam by \cite{kingma2014adam} in a distributed fashion. The DANCE method is designed  to be easily parallelized and shares the strong scaling property, i.e., linear speed-up property. Since it is possible to split gradient and Hessian-vector product computations across different machines, it is always expected to get extra acceleration via increasing the number of computational nodes. We formally characterize  the required number of communication rounds to reach the statistical accuracy of the full dataset. For a distributed setting, we show that DANCE is communication efficient in both theory and practice. In particular, Table \ref{Tab1} highlights the advantage of DANCE with respect to other adaptive sample size methods which will be discussed in more details in Section \ref{sec4}.

We organize this paper as following. In Section \ref{sec:2problem}, we introduce the necessary assumptions and the definition of statistical accuracy. Section \ref{sec3} describes the proposed algorithm and its distributed version. Section \ref{sec4} explores the theoretical guarantees on complexity of DANCE. In Section \ref{sec:NumExp}, we demonstrate the outstanding performance of our algorithm in practice. In Section \ref{sec6}, we close the paper by concluding remarks.


\section{Problem Formulation}\label{sec:2problem}

In this paper, we focus on finding the optimal solution $w^*$ of the problem in \eqref{eq:ExpLoss}. As described earlier, due to difficulties in the expected risk minimization, as an alternative, we aim to find a solution for the empirical loss function $L_N(w)$,  which is the empirical mean over $N$ samples. Now, consider the empirical loss $L_n(w)$ associated with $n \leq N$ samples. In \cite{estError107} and \cite{estError207}, it has been shown that the difference between the expected loss and the empirical loss $L_n$ with high probability (w.h.p.) is upper bounded by the statistical accuracy $V_n$, i.e., w.h.p.
\begin{equation}\label{eq:statAccu}
	\sup_{w \in \R^d}|L(w)-L_n(w)|\leq V_n.
\end{equation}
In other words, there exists a constant $\vartheta$ such that the inequality \eqref{eq:statAccu} holds with probability of at least $1-\vartheta$. Generally speaking, statistical accuracy $V_n$ depends on $n$ (although it depends on $\vartheta$ too, but for simplicity in notation we just consider the size of the samples), and is of order $V_n = \mathcal{O}({1}/{n^{\gamma}})$ where $\gamma \in [0.5, 1]$ \cite{Vapnik13,1444,bartlett2006convexity}.

For problem \eqref{eq:ERM}, if we find an approximate solution $w_n$ which satisfies the inequality $L_n(w_n)-L_n(\hat{w}_n) \leq V_n$, where $\hat{w}_n$ is the true minimizer of $L_n$, it is not necessary to go further and find a better solution (a solution with less optimization error). The reason comes from the fact that for a more accurate solution the summation of estimation and optimization errors does not become smaller than $V_n$. Therefore, when we say that $w_n$ is a $V_n$-suboptimal solution for the risk $L_n$, it means that $L_n(w_n)-L_n(\hat{w}_n) \leq V_n$. In other words, $w_n$ solves problem \eqref{eq:ERM} within its statistical accuracy.

It is crucial to note that if we add an additional term in the magnitude of $V_n$ to the empirical loss $L_n$, the new solution is also in the similar magnitude as $V_n$ to the expected loss $L$. Therefore, we can regularize the non-strongly convex loss function $L_n$ by ${cV_n}\tnorm{w}/{2}$ and consider it as the following problem:\vspace{-7pt}
\begin{equation}\label{eq:ERMreg}
	\min_{w\in \R^d} R_n(w):=\frac{1}{n}\sum_{i=1}^{n}f_i(w) + \frac{cV_n}{2}\tnorm{w}.
\end{equation}
The noticeable feature of the new empirical risk $R_n$ is that $R_n$ is $cV_n$-strongly convex\footnote{$cV_n$ depends on number of samples, probability, and VC dimension of the problem. For simplicity in notation, we just consider the number of samples.}, where $c$ is a positive constant depending on the VC dimension of the problem. Thus, we can utilize any practitioner-favorite algorithm. Specifically, we are willing to apply the inexact damped Newton method, which will be discussed in the next section. Due to the fact that a larger strong-convexity parameter leads to a faster convergence, we could expect that the first few steps would converge fast since the values of $cV_n$ in these steps are large (larger statistical accuracy), as will be discussed in Theorem \ref{Thm:linearIter}.
From now on, when we say $w_n$ is an $V_n$-suboptimal solution of the risk $R_n$, it means that $R_n(w_n)-R_n(w_n^*)\leq V_n$, where $w_n^*$ is the true optimal solution of the risk $R_n$. Our final aim is to find $w_N$ which is $V_N$-suboptimal solution for the risk $R_N$ which is the risk over the whole dataset.\\
In the rest of this section, first we define the self-concordant functions which have the property that its third derivative can be controlled by its second derivative. By assuming that function $f:\R^d\to \R$ has continuous third derivative, we define self-concordant function as follows.

\begin{definition}
	A convex function $f:\R^d\to \R$ is $\rho_f$-self-concordant if for any $w \in \text{dom}(f)$ and $u \in \R^d$ \vspace{-4pt}
	\begin{equation}\label{eq:defSelfConcor}
		|u^T(f^{'''}(w)[u])u| \leq\rho_f(u^T\nabla^2f(w)u)^{\tfrac{3}{2}},\vspace{-4pt}
	\end{equation}
\end{definition}

where $f^{'''}(w)[u]:=\lim_{t\to 0}\tfrac{1}{t}(\nabla^2f(w+tu)-\nabla^2 f(w))$. As it is discussed in \cite{nesterov2013introductory}, any self-concordant function $f$ with parameter $\rho_f$ can be rescaled to become standard self-concordant (with parameter 2). Some of the well-known empirical loss functions which are self-concordant are linear regression, Logistic regression and squared hinge loss. In order to prove our results the following conditions are considered in our analysis.

\begin{assumption}\label{assum:smooth}
	The loss functions $f(w,z)$ are convex w.r.t $w$ for all values of $z$. In addition, their gradients $\nabla f(w,z)$ are $M-$Lipschitz continuous
	\begin{equation}\label{eq:smoothAssum}
		\norm{\nabla f(w,z) - \nabla f(w',z)} \leq M \norm{w - w'},\,\, \forall z.
	\end{equation}
\end{assumption}
\begin{assumption}\label{assum:selfConcor}
	The loss functions $f(w,z)$ are self-concordant w.r.t $w$ for all values of $z$.
\end{assumption}
The immediate conclusion of Assumption \ref{assum:smooth} is that both $L(w)$ and $L_n(w)$ are convex and $M$-smooth. Also, we can note that $R_n(w)$ is $cV_n$-strongly convex and $(cV_n+M)$-smooth. Moreover, by Assumption \ref{assum:selfConcor}, $R_n(w)$ is also self-concordant.

\section{Distributed Accumulated Newton Conjugate Gradient Method}\label{sec3}
The goal in inexact damped Newton method, as discussed in \cite{Disco15}, is to find the next iterate based on an approximated Newton-type update. It has two important differences comparing to Newton's method. First, as it is clear from the word ``damped'', the learning rate of the inexact damped Newton type update is not $1$, since it depends on the approximation of Newton decrement. The second distinction is that there is no need to compute exact Newton direction (which is very expensive to calculate in one step). Alternatively, an approximated inexact Newton direction is calculated by applying an iterative process to obtain a direction with desirable accuracy under some measurements.

In order to utilize the important features of ERM, we combine the idea of increasing sample size and the inexact damped Newton method. In our proposed method, we start with handling a small number of samples, assume $m_0$ samples. We then solve its corresponding ERM to its statistical accuracy, i.e. $V_{m_{0}}$, using the inexact damped Newton algorithm. In the next step, we increase the number of samples geometrically with rate of $\alpha>1$, i.e., $\alpha m_0$ samples. The approximated solution of the previous ERM can be used as a warm start point to find the solution of the new ERM. The sample size increases until it equals the number of full samples.

Consider the iterate $w_m$ within the statistical accuracy of the set with $m$ samples, i.e. $\mathcal{S}_m$ for the risk $R_m$. In DANCE, we increase the size of the training set to $n=\alpha m$ and use the inexact damped Newton to find the iterate $w_n$ which is $V_n$-suboptimal solution for the sample set $\mathcal{S}_n$, i.e. $R_n(w_n)-R_n(w_n^*)\leq V_n$ after $K_n$ iterations. To do so, we initialize $\tilde{w}_0 =w_m$ and update the iterates according to the following
\begin{equation}\label{eq:updt1}
	\tilde{w}_{k+1}=\tilde{w}_k - \tfrac{1}{1+\delta_n(\tilde{w}_k)} v_k,
\end{equation}
where $v_k$ is an $\epsilon_k$-Newton direction. The outcome of applying \eqref{eq:updt1} for $k=K_n$ iterations is the approximate solution $w_n$ for the risk  $R_n$, i.e., $w_n:=\tilde{w}_{K_n}$.

To properly define the approximate Newton direction $v_k$, first consider that the gradient and Hessian of the risk $R_n$ can be evaluated as
\begin{equation}\label{eq:gradOfR_n}
	\nabla R_n(w) = \frac{1}{n}\sum_{i=1}^{n}\nabla f_i(w) +cV_nw
\end{equation}
and
\begin{equation}\label{eq:hessOfR_n}
	\nabla^2 R_n(w) = \frac{1}{n}\sum_{i=1}^{n}\nabla^2 f_i(w) +cV_nI,
\end{equation}
respectively. The favorable descent direction would be the Newton direction $-\nabla^2R_n(\tilde{w}_k)^{-1}\nabla R_n(\tilde{w}_k)$; however, the cost of computing this direction is prohibitive. Therefore, we use $v_k$ which is an $\epsilon_k$-Newton direction satisfying the condition
\begin{equation}\label{eq:apprxV_m}
	\|\nabla^2R_n(\tilde{w}_k)v_k - \nabla R_n(\tilde{w}_k)\| \leq \epsilon_k.
\end{equation}
As we use the descent direction $v_k$ which is an approximation for the Newton step, we also redefine the Newton decrement $\delta_n(\tilde{w}_k)$ based on this modification. To be more specific, we define $\delta_n(\tilde{w}_k):=(v_k^T\nabla^2R_n(\tilde{w}_k)v_k)^{1/2}$ as the approximation of (exact) Newton decrement $(\nabla R_n(\tilde{w}_k)^T \nabla^2 R_n(\tilde{w}_k)^{-1} \nabla R_n(\tilde{w}_k))^{1/2}$, and use it in the update in  \eqref{eq:updt1}.\\

In order to find $v_k$ which is an $\epsilon_k$-Newton direction, we use Preconditioned CG (PCG). As it is discussed in \cite{Disco15,nocedal2006sequential}, PCG is an efficient iterative process to solve Newton system with the required accuracy. The preconditioned matrix that we considered is in the form of $P =  \tilde{H}_n+\mu_n I$, where $\tilde{H}_n = \tfrac{1}{|\mathcal{A}_n|}\sum_{i \in \mathcal{A}_n}\nabla^2 R_n^i(w)$, $\mathcal{A}_n \subset \mathcal{S}_n$, and $\mu_n$ is a small regularization parameter. In this case, $v_k$ is an approximate solution of the system $P^{-1}\nabla^2R_n(\tilde{w}_k)v_k = P^{-1}\nabla R_n(\tilde{w}_k)$. The reason for using preconditioning is that the condition number of $P^{-1}\nabla^2R_n(\tilde{w}_k)$ may be close to 1 in the case when $\tilde{H}_n$ is close to $\nabla^2R_n(\tilde{w}_k)$; consequently, PCG can be faster than CG. The PCG steps are summarized in Algorithm~\ref{alg:alg2}. In every iteration of Algorithm \ref{alg:alg2}, a system needs to be solved in step 10. Due to the structure of matrix $P$, and as it is discussed in \cite{ma2016distributed}, this matrix can be considered as  $|\mathcal{A}_n|$ rank 1 updates on a diagonal matrix, and now, using Woodbury Formula \cite{Press03} is a very efficient way to solve the mentioned system. 
The following lemma states the required number of iterations for PCG to find an $\epsilon_k$-Newton direction $v_k$ which is used in every stage of DANCE algorithm.\\

\begin{lemma}\label{lem:lemma4Disco}
	(Lemma 4 in \cite{Disco15})
	Suppose Assumption \ref{assum:smooth} holds and $\|\tilde{H}_n - \nabla^2R_n(\tilde{w}_k) \| \leq \mu_n$. Then, Algorithm \ref{alg:alg2}, after $C_n(\epsilon_k)$ iterations calculates $v_k$ such that $\|\nabla^2R_n(\tilde{w}_k)v_k - \nabla R_n(\tilde{w}_k)\| \leq \epsilon_k$, where
	\begin{equation}
		C_n(\epsilon_k) = \left\lceil \sqrt{(1+\tfrac{2\mu_n}{cV_n})}\log\left(\tfrac{2\sqrt{\tfrac{cV_n+M}{cV_n}}\|\nabla R_n(\tilde{w}_k)\|}{\epsilon_k}\right) \right\rceil.
	\end{equation}
\end{lemma}

{Note that $\epsilon_k$ has a crucial effect on the speed of the algorithm. When $\epsilon_k=0$, then $v_k$ is the exact Newton direction, and the update in \eqref{eq:updt1} is the exact damped Newton step (which recovers the update in Ada Newton algorithm in \cite{Mokhtari_AdaNewton16} when the step-length is 1). Furthermore, 
	the number of total iterations to reach $V_N$-suboptimal solution for the risk $R_N$ is $\mathbf{K}$, i.e. $\mathbf{K}=K_{m_0}+ K_{\alpha m_0}+\dots +K_N$. Hence, if we start with the iterate $w_{m_0}$ with corresponding $m_0$ samples, after $\mathbf{K}$ iterations, we reach $w_N$ with statistical accuracy of $V_N$ for the whole dataset. In Theorem \ref{Thm:linearIter}, the required rounds of communication to reach the mentioned statistical accuracy will be discussed.
	
	\begin{algorithm}[tb]
		\caption{DANCE}
		\label{alg:alg1}
		\begin{algorithmic}[1]
			\STATE Initialization: Sample size increase constant $\alpha$, initial sample size $n=m_0$ and $w_n=w_{m_0}$ with $\|\nabla R_n(w_n)\| < (\sqrt{2c})V_n$
			\WHILE {$n \leq N$}
			\STATE Update $w_m=w_n$ and $m=n$
			\STATE Increase sample size: $n=\min\{\alpha m,N\}$
			\STATE Set $\tilde{w}_0={w}_m$ and set $k = 0$
			\REPEAT  
			\STATE Calculate $v_k$ and $\delta_n(\tilde{w}_k)$ by \textbf{Algorithm \ref{alg:alg2} PCG}
			\STATE Set $ \tilde{w}_{k+1}=\tilde{w}_{k} - \tfrac{1}{1+\delta_n(\tilde{w}_{k})} v_k$
			\STATE $k=k+1$
			\UNTIL satisfy stop criteria leading to $R_n(\tilde{w}_{k})-R_n(w_n^*)\leq V_n$
			\STATE Set $w_n=\tilde{w}_{k}$
			\ENDWHILE
		\end{algorithmic}
	\end{algorithm}
	
	\begin{algorithm*}[tb]
		\caption{PCG} 
		\label{alg:alg2}
		
		\begin{algorithmic}[1]
			\STATE \textbf{Master Node:} \hfill \textbf{Worker Nodes (\pmb{$i=1,2,\dots,\mathcal{K}$}):} 
			\STATE {\bfseries Input:} $\tilde{w}_k \in \R^d$, $\epsilon_k$, and $\mathcal{A}_n$
			\STATE Let $H=\nabla^2 R_n(\tilde{w}_k)$, $\displaystyle{P=\tfrac{1}{|\mathcal{A}_n|}\sum_{i \in \mathcal{A}_n}\nabla^2 R_n^i(\tilde{w}_k)+\mu_n I}$
			\STATE {\color{green!40!black}{\it \textbf{Broadcast:}}} $\tilde{w}_k$ \hfill \textcolor{green!40!black}{$\pmb{\longrightarrow}$}
			\hfill Compute  $\nabla R_n^i (\tilde{w}_k)$
			\STATE {\color{red!50!black}{\it \textbf{Reduce:}}} $\nabla R_n^i (\tilde{w}_k)$ to $\nabla R_n (\tilde{w}_k)$ \hfill \textcolor{red!50!black}{$\pmb{\longleftarrow}$} \hfill $\ \ \ $
			\STATE Set $r^{(0)}=\nabla R_n(\tilde{w}_k)$, $u^{(0)}=s^{(0)}=P^{-1}r^{(0)}$
			\STATE Set $v^{(0)}=0$, $t=0$
			\REPEAT
			\STATE {\color{green!40!black}{\it \textbf{Broadcast:}}} $u^{(t)}$ and $v^{(t)}$ \hfill \textcolor{green!40!black}{$\pmb{\longrightarrow}$}
			\hfill Compute  $\nabla^2 R_n^i(\tilde{w}_k)u^{(t)}$ and $\nabla^2 R_n^i(\tilde{w}_k)v^{(t)}$
			\STATE {\color{red!50!black}{\it \textbf{Reduce:}}} $\nabla^2 R_n^i(\tilde{w}_k)u^{(t)}$ and $\nabla^2 R_n^i(\tilde{w}_k)v^{(t)}$ to $Hu^{(t)}$ and $Hv^{(t)}$ \hfill \textcolor{red!50!black}{$\pmb{\longleftarrow}$} \hfill $\ \ \ $
			\STATE Compute $\gamma_t = \tfrac{\langle r^{(t)},s^{(t)}\rangle}{\langle u^{(t)},Hu^{(t)}\rangle}$
			\STATE Set $v^{(t+1)} = v^{(t)}+\gamma_t u^{(t)}$, $r^{(t+1)} = r^{(t)}-\gamma_t H u^{(t)}$
			\STATE Compute $\zeta_t = \tfrac{\langle r^{(t+1)},s^{(t+1)}\rangle}{\langle r^{(t)},s^{(t)}\rangle}$
			\STATE Set $ Ps^{(t+1)} = r^{(t+1)}$, $u^{(t+1)} = s^{(t+1)}+\zeta_t u^{(t)}$
			\STATE Set $t=t+1$
			\UNTIL{$\|r^{(t+1)}\| \leq \epsilon_k$ }
			\STATE {\bfseries Output:} $v_k=v^{(t+1)}$, $\delta_n(\tilde{w}_k) = \sqrt{v_k^THv^{(t)}+\gamma_tv_k^TH u^{(t)}}$
		\end{algorithmic}
	\end{algorithm*}
	Our proposed method is summarized in Algorithm \ref{alg:alg1}. We start with $m_0$ samples, and an initial point $w_{m_0}$ which is an $V_{m_0}-$ suboptimal solution for the risk $R_{m_0}$.
	In every iteration of outer loop of Algorithm \ref{alg:alg1}, we increase the sample size geometrically with rate of $\alpha$ in step 4. In the inner loop of Algorithm \ref{alg:alg1}, i.e. steps 6-10, in order to calculate the approximate Newton direction and approximate Newton decrement, we use PCG algorithm which is shown in Algorithm \ref{alg:alg2}. This process repeats till we get the point $w_N$ with statistical accuracy of $V_N$. The practical stopping criteria for Algorithm \ref{alg:alg1} is discussed in Section \ref{pracStopCriterion}.\vspace{-4pt}

	\paragraph{Distributed Implementation} Similar to the algorithm in \cite{Disco15}, Algorithms \ref{alg:alg1} and \ref{alg:alg2} can also be implemented in a distributed environment. Suppose the entire dataset is stored across $\mathcal{K}$ machines, i.e., each machine stores $N_i$ data samples such that $\sum_{i=1}^{\mathcal{K}} N_i = N$. Under this setting, each iteration in Algorithm~\ref{alg:alg1} can be executed on different machines in parallel with $\sum_{i=1}^\mathcal{K} n_i = n$, where $n_i$ is the batchsize on $i^{th}$ machine. To implement Algorithm~\ref{alg:alg2} in a distributed manner, a broadcast operation is needed at each iteration to guarantee that each machine will share the same $\tilde w_k$ value. Moreover, the gradient and Hessian-vector product can be computed locally and later reduce to the master machine. With the increasing of batch size, computation work on each machine will increase while we still have the same amount of communication need. As a consequence, the computation expense will gradually dominate the communication expense before the algorithm terminates. Therefore the proposed algorithm could take advantage of utilizing more machines to shorten the running time of Algorithm~\ref{alg:alg2}.

	\section{Complexity Analysis}\label{sec4}
	In this section, we study the convergence properties of our algorithm. To do so, we analyze the required number of communication rounds and total computational complexity of DANCE to solve every subproblem up to its statistical accuracy. 
	
	We analyze the case when we have $w_m$ which is a $V_m$-suboptimal solution of the risk $R_m$, and we are interested in deriving a bound for the number of required communication rounds to ensure that $w_n$ is a $V_n$-suboptimal solution for the risk $R_n$. 
	\begin{theorem}\label{Thm:linearIter}
		Suppose that Assumptions \ref{assum:smooth} and \ref{assum:selfConcor} hold. Consider $w_m$ which satisfies $R_m(w_m) - R_m(w_m^*) \leq V_m$ and also the risk $R_n$ corresponding to sample set $\mathcal{S}_n \supset \mathcal{S}_m$ where $n = \alpha m,\,\,\alpha >1$. Set the parameter $\epsilon_k$ (the error in \eqref{eq:apprxV_m}) as following
		\begin{equation}\label{eq:epsthm1}
			\epsilon_k = \beta(\tfrac{cV_n}{M+cV_n})^{1/2} \| \nabla R_n(\tilde{w}_k)\|,
		\end{equation}
		where $\beta \leq \tfrac{1}{20}$. Then, in order to find the variable $w_n$ which is an $V_n$-suboptimal solution for the risk $R_n$, i.e $R_n(w_n) - R_n(w_n^*) \leq V_n$, the number of communication rounds $T_n$ satisfies in the following:
		\begin{align}\label{eq:bndIter1}
			T_n \leq &K_n\left(1+C_n(\epsilon_k)\right),\,\,\,\, \text{w.h.p}.
		\end{align}
		where $K_n=\Big\lceil \tfrac{R_n(w_m) - R_n(w_n^*) }{\frac{1}{2}\omega (1/6)} \Big\rceil  + \Big\lceil \log_2(\tfrac{2\omega(1/6)}{V_n})\Big\rceil $. Here $\lceil t\rceil$ shows the smallest nonnegative integer larger than or equal to $t$.
	\end{theorem}
	As a result, the update in \eqref{eq:updt1} needs to be done for $K_n = \mathcal{O}(\log_2{n})$ times in order to attain the solution $w_n$ which is $V_n$-suboptimal solution for the risk $R_n$. 
	Also, based on the result in \eqref{eq:bndIter1}, by considering the risk $R_n$, we can note that when the strong-convexity parameter for the mentioned risk ($cV_n$) is large, less number of iterations (communication rounds) are needed (or equally faster convergence is achieved) to reach the iterate with $V_n$-suboptimal solution; and this happens in the first steps. More importantly, DANCE is a multi-stage algorithm which any stage of DANCE converges linearly (the reason is that in every stage of our DANCE method, we use ineaxt damped-Newton method, and as it is shown in \cite{Disco15} that with $\epsilon_k $ in \eqref{eq:epsthm1}, the ineaxt damped-Newton method converges linearly). Furthermore, if we consider DANCE as 1-stage algorithm with full samples in each iteration, it covers DiSCO which converges linearly.
	\begin{corollary}\label{Cor:linearIter}
		Suppose that Assumptions \ref{assum:smooth} and \ref{assum:selfConcor} hold. Further, assume that $w_m$ is a $V_m$-suboptimal solution for the risk $R_m$ and consider $R_n$ as  the risk corresponding to sample set $\mathcal{S}_n \supset \mathcal{S}_m$ where $n = 2 m$. If we set parameter $\epsilon_k$ (the error in \eqref{eq:apprxV_m}) as \eqref{eq:epsthm1}, then  with high probability $\tilde{T}_n$ communication rounds
		\begin{align}\label{eq:bndIter11}
			\tilde{T}_n \leq &\Big(\Big\lceil \tfrac{\Big(3+\big(1-\tfrac{1}{2^{\gamma}}\big)\big(2+\tfrac{c}{2}\tnorm{w^{*}}
				\big) \Big)V_m}{\frac{1}{2}\omega (1/6)} \Big\rceil  + \Big\lceil \log_2(\tfrac{2\omega(1/6)}{V_n})\Big\rceil \Big)\nonumber\\
			&\left(1+\Big\lceil \sqrt{1+\tfrac{2\mu}{cV_n})}\log_2\Big(\tfrac{2(cV_n+ M)}{\beta cV_n}\Big)\Big\rceil\right),
		\end{align}
		are needed to reach the point $w_n$ with statistical accuracy of $V_n$ for the risk $R_n$.
	\end{corollary}\vspace{2mm}
	\begin{corollary}\label{Cor:totalNumOfCommunication}
		By assuming $\alpha =2$, the total number of communication rounds to reach a point with the statistical accuracy of $V_N$ of the full training set is w.h.p.  
		\begin{align}\label{eq:bndIter1Total}
            \tilde{\mathcal{T}} = \tilde{\mathcal{O}}(\gamma(\log_2{N})^2\sqrt{N^{\gamma}}\log_2{N^{\gamma}})
		\end{align}
			 where $\gamma \in [0.5, 1]$.
	\end{corollary}

	\begin{figure*}[ht]
		\centering
		\includegraphics[width=0.245\textwidth]{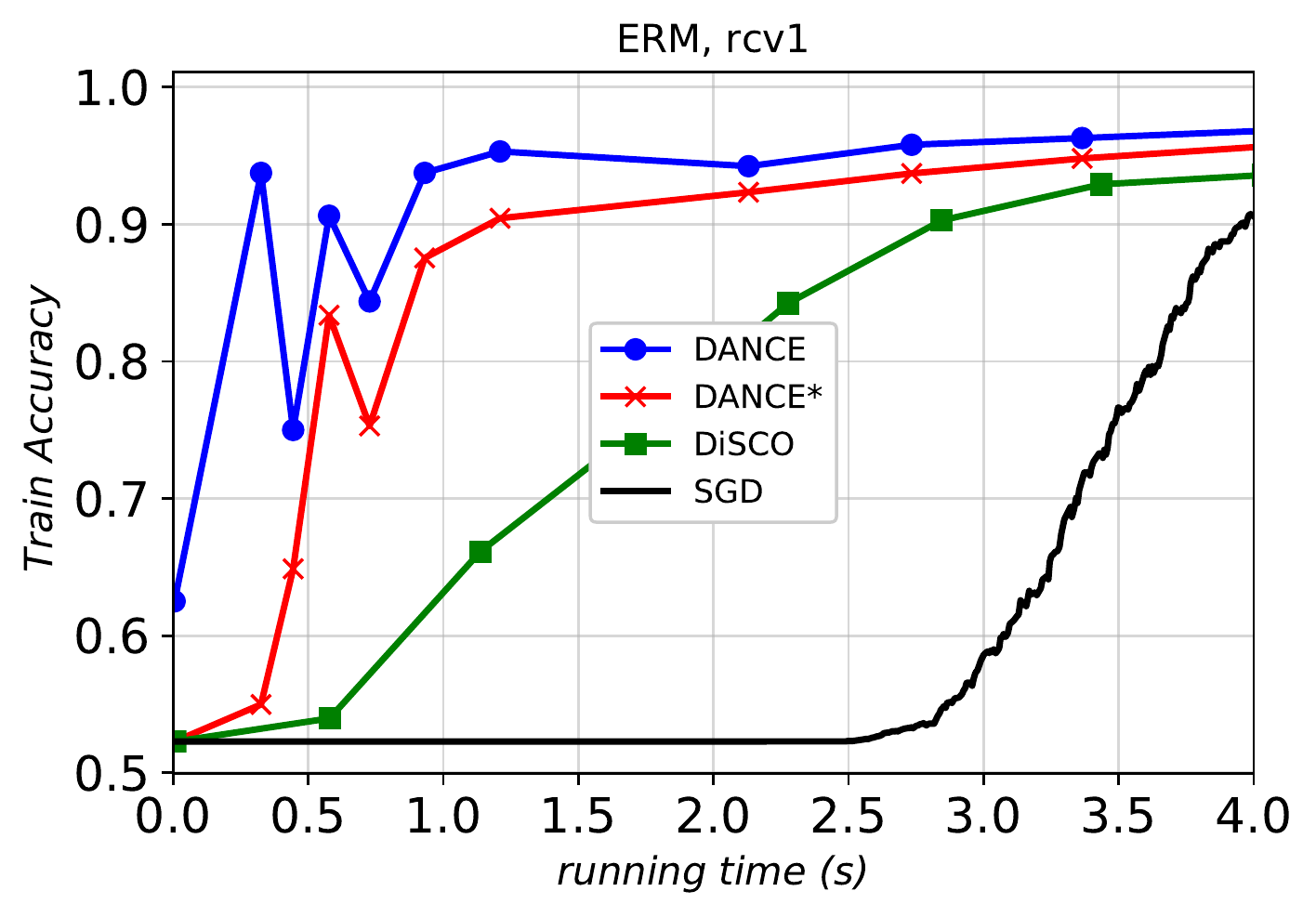}
		\includegraphics[width=0.245\textwidth]{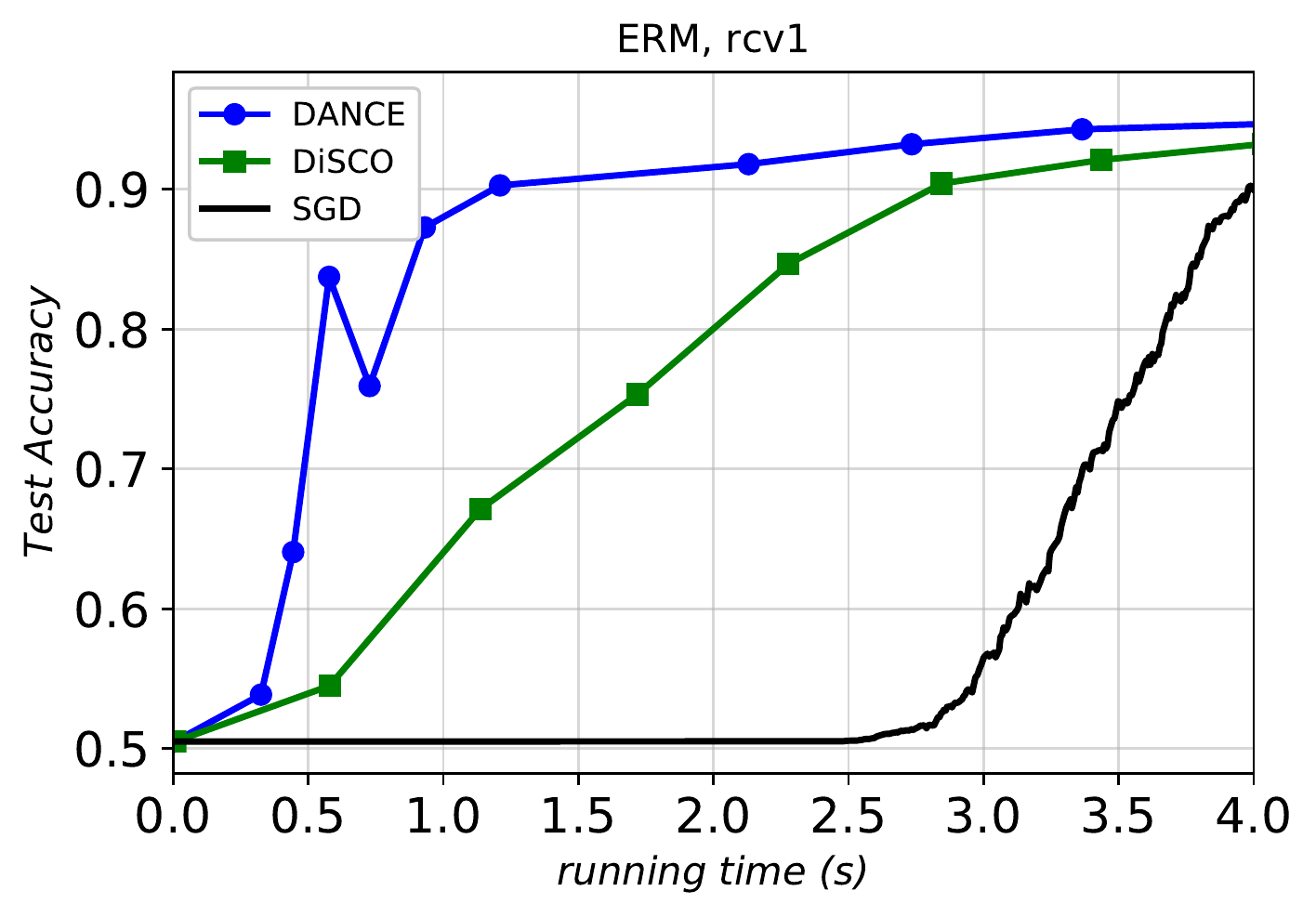}
		\includegraphics[width=0.245\textwidth]{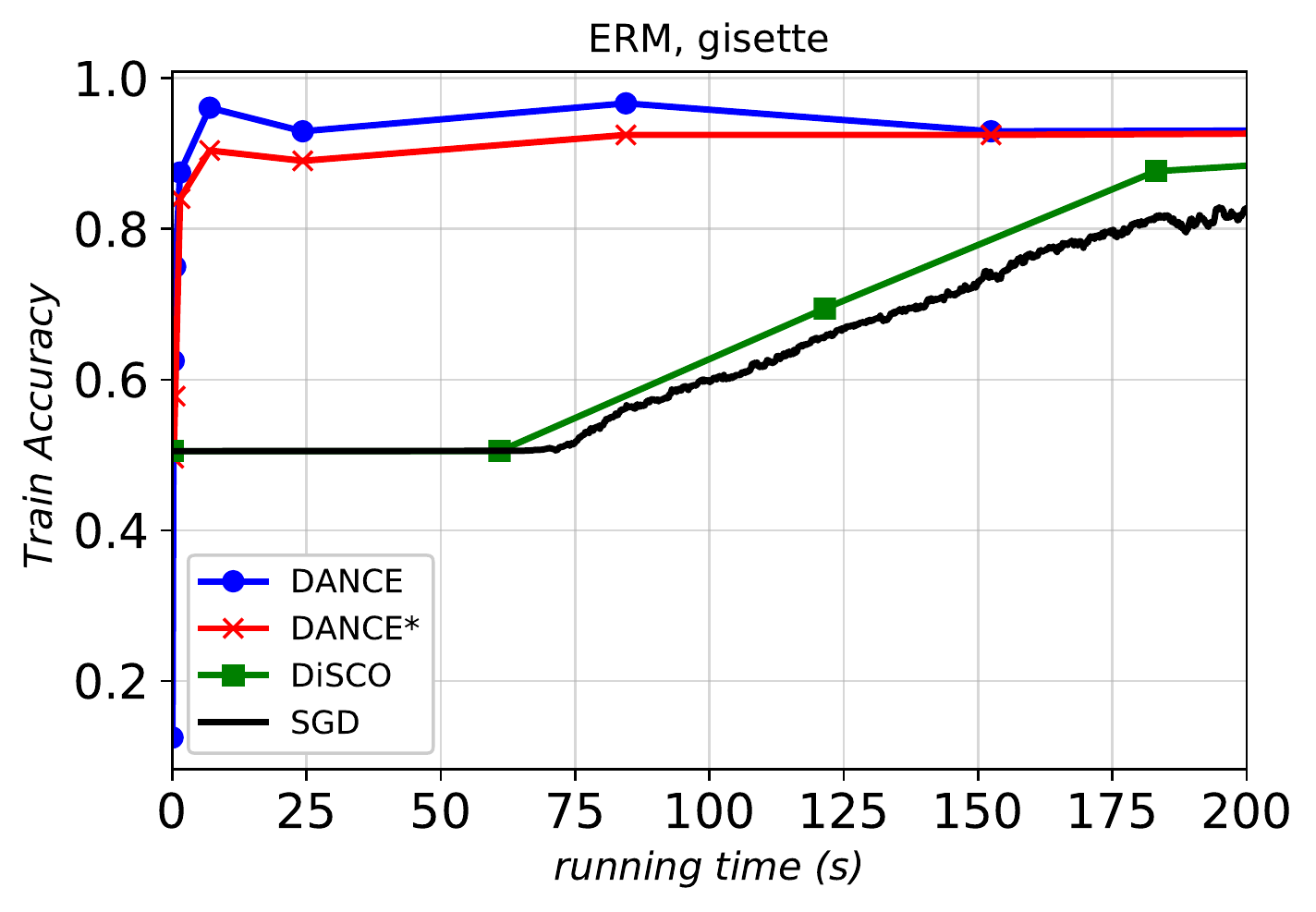}
		\includegraphics[width=0.245\textwidth]{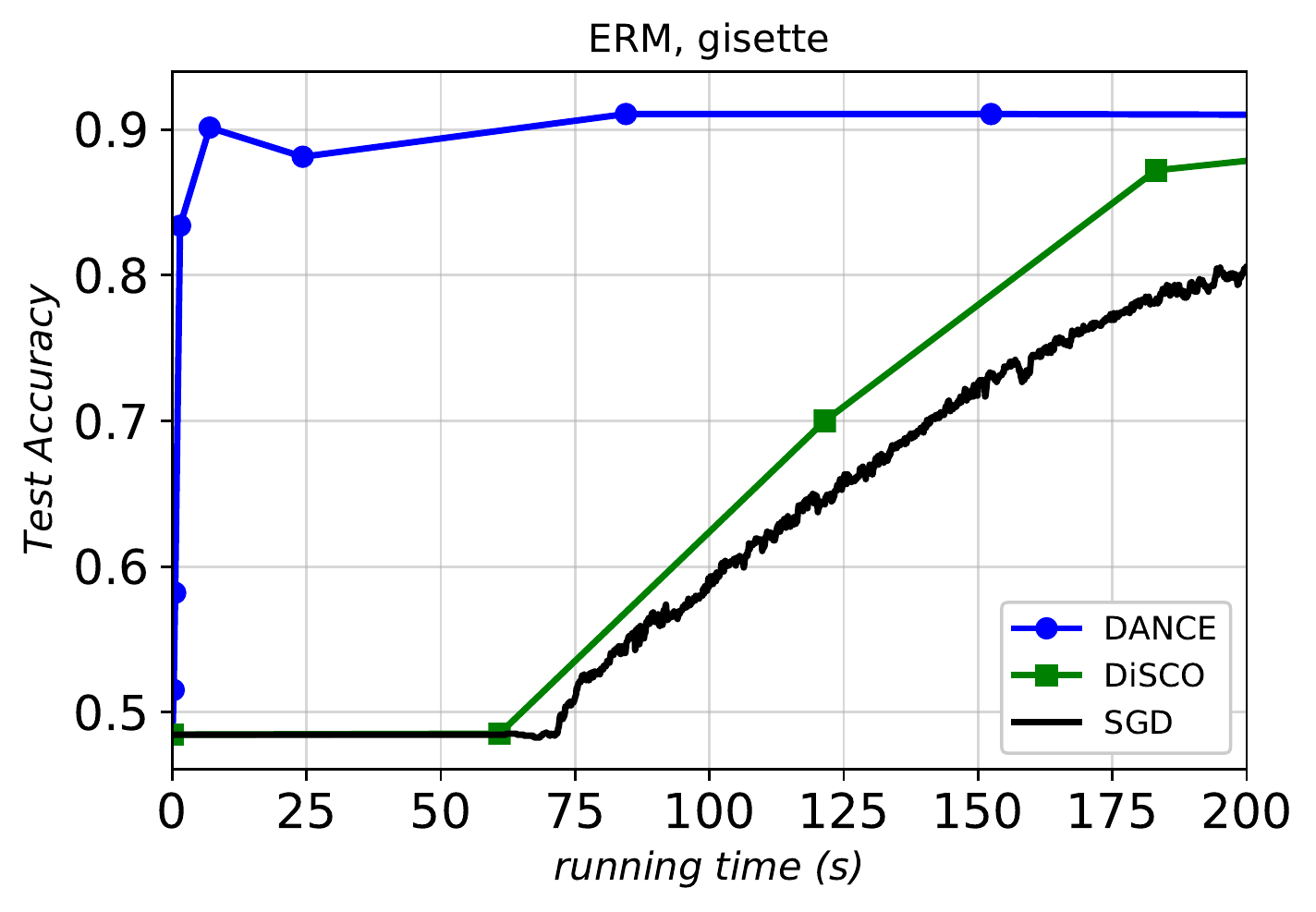}
		\vspace{-15pt}
		\caption{Performance of different algorithms on a logistic regression problem with rcv1 and gissete datasets. 
			In the left two figures, the plot \textit{DANCE*} is the training accuracy based on the entire training set, while the plot \textit{DANCE} represents the training accuracy based on the current sample size. }
		\label{fig:convex, rcv}
	\end{figure*}
	
	The rounds of communication for DiSCO in \cite{Disco15}\footnote{In order to have fair comparison, we put $f=R_N$, $\epsilon = V_N$, and $\lambda = cV_N$ in their analysis, and also the constants are ignored for the communication complexity.} is $\tilde{\mathcal{T}}_{DiSCO} = \tilde{\mathcal{O}}((R_N(w_0)-R_N(w_N^*)+\gamma(\log_2{N}))\sqrt{N^{\gamma}}\log_2{N^{\gamma}})$ where $\gamma \in [0.5, 1]$. Comparing these bounds shows that the communication complexity of DANCE is independent of the choice of initial variable $w_0$ and the suboptimality $R_N(w_0)-R_N(w_N^*)$, while the overall communication complexity of DiSCO depends on the initial suboptimality. In addition, implementation of each iteration of DiSCO requires processing all the samples in the dataset, while DANCE only operates on an increasing subset of samples at each phase. Therefore, the computation complexity of DANCE is also lower than DiSCO for achieving the statistical accuracy of the training set.\vspace{-2mm}
	
	\begin{theorem}\label{Thm:totalComplexity}
		The total complexity of DANCE Algorithm in order to reach a point with the statistical accuracy of $V_N$ of the full training set is w.h.p.  
		\begin{align}\label{eq:totalComplexity}
			\tilde{\mathcal{O}}((\log_2(N))^3N^{1/4}d^2)
		\end{align}

	\end{theorem}	
	 Table \ref{Tab1} shows that the total complexity of the $k-$Truncated method \cite{Eisen17} is lower than the one for AdaNewton \cite{Mokhtari_AdaNewton16}. Further, as $(\log_2(N))^3 \ll N^{3/4} $, the total complexity of DANCE is lower than both AdaNewton and $k-$TAN methods. All in all, the theoretical results highlight that DANCE is more efficient than DiSCO in terms of communication, and has lower complexity than previous adaptive sample size methods including AdaNewton and $k-$TAN.
	\vspace{-5pt}
	\section{Numerical Experiments}\label{sec:NumExp}
	In this section, we present numerical experiments on several large real-world datasets to show that our restarting DANCE algorithm can outperform other existed methods on solving both convex and non-convex problems. Also, we compare the results obtained from utilizing different number of machines to demonstrate the strong scaling property for DANCE. All the algorithms are implemented in Python with PyTorch \cite{paszke2017automatic} library and we use MPI for Python \cite{dalcin2011parallel} distributed environment\footnote{All codes to reproduce these experimental results are available at \href{https://github.com/OptMLGroup/DANCE}{https://github.com/OptMLGroup/DANCE.}}. For all plots in this section, vertical pink dashed lines represent restarts in our DANCE method. 
		
		

\begin{figure*}[t]
		\centering
		\includegraphics[width=0.22\textwidth]{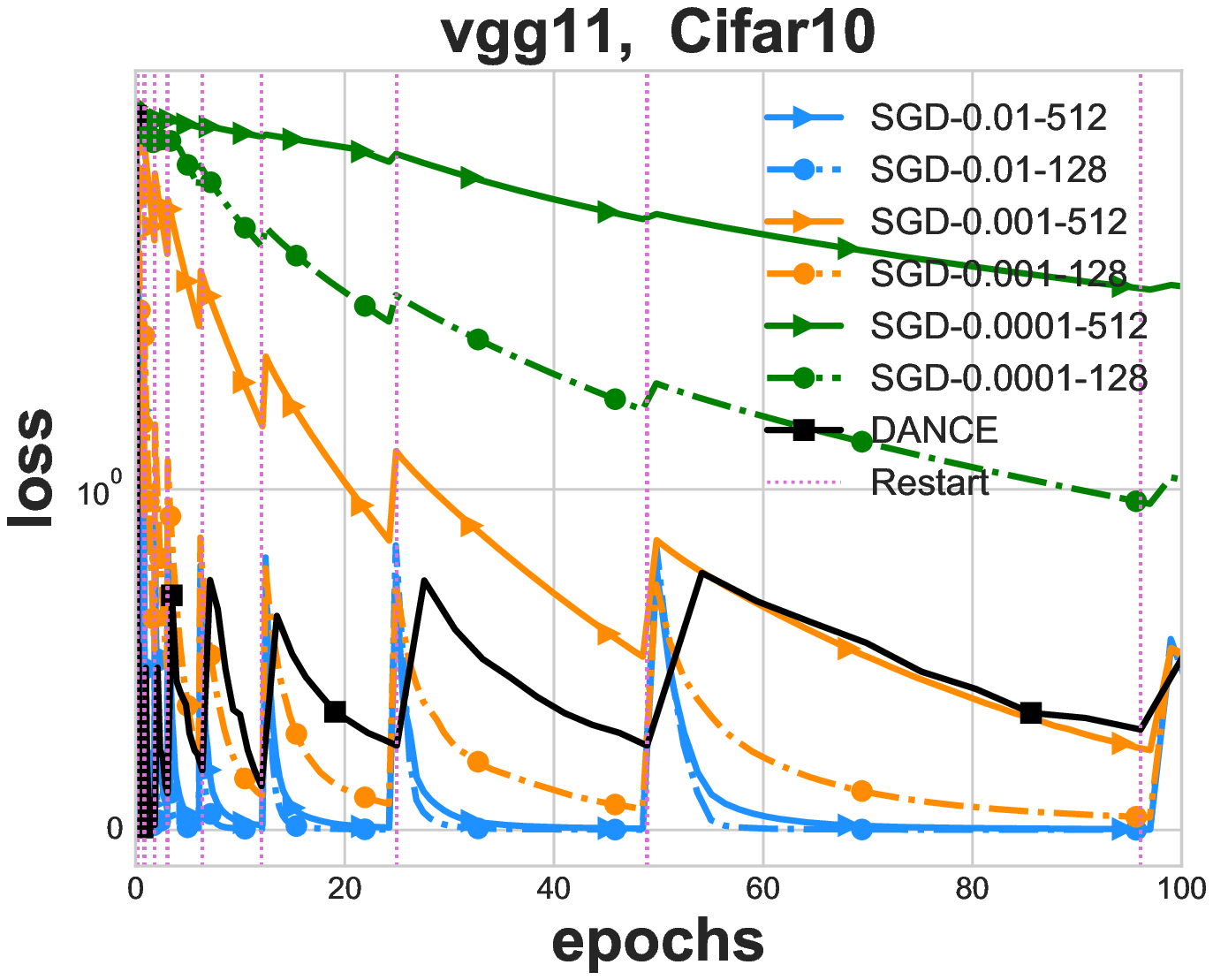}
		\includegraphics[width=0.22\textwidth]{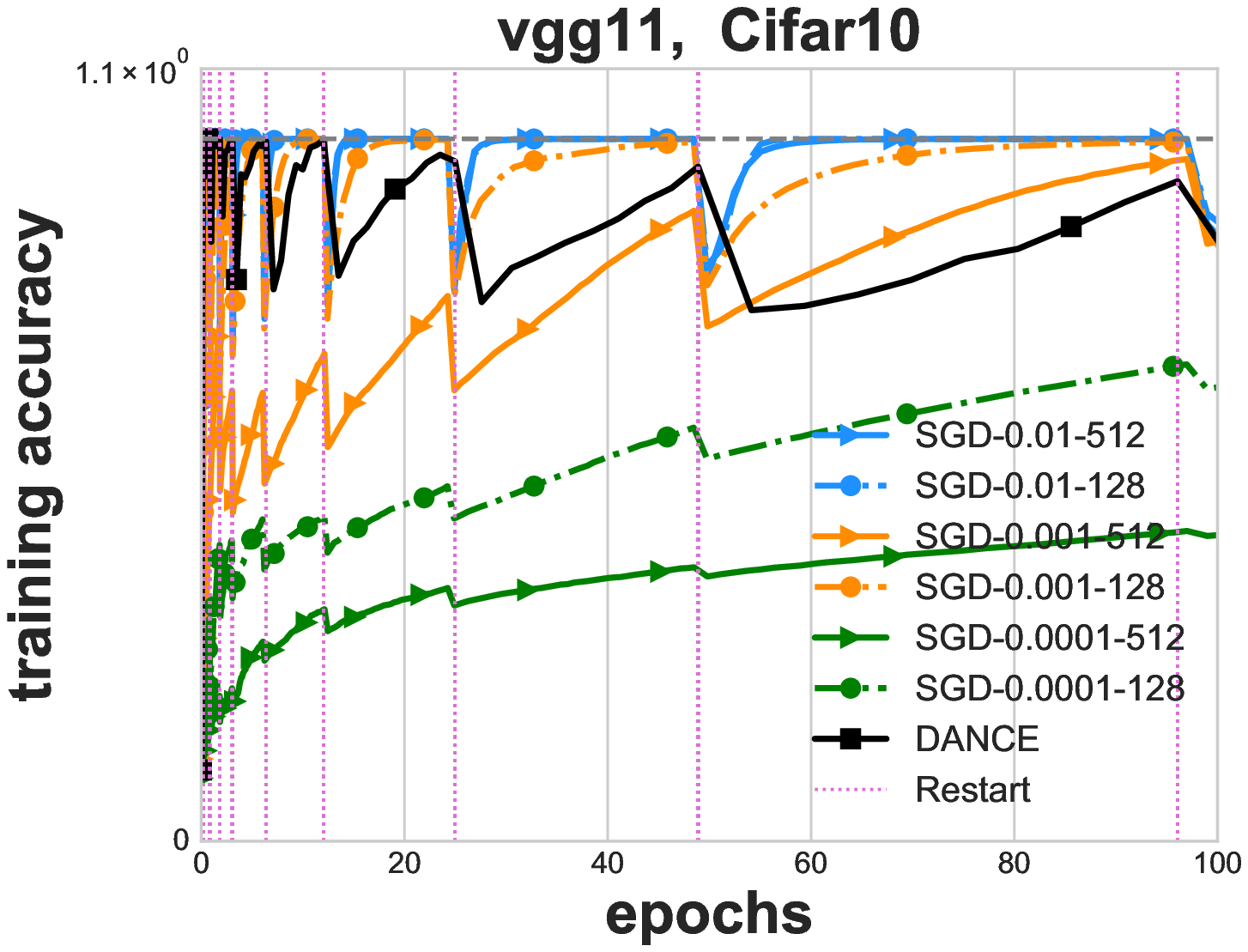}
		\includegraphics[width=0.22\textwidth]{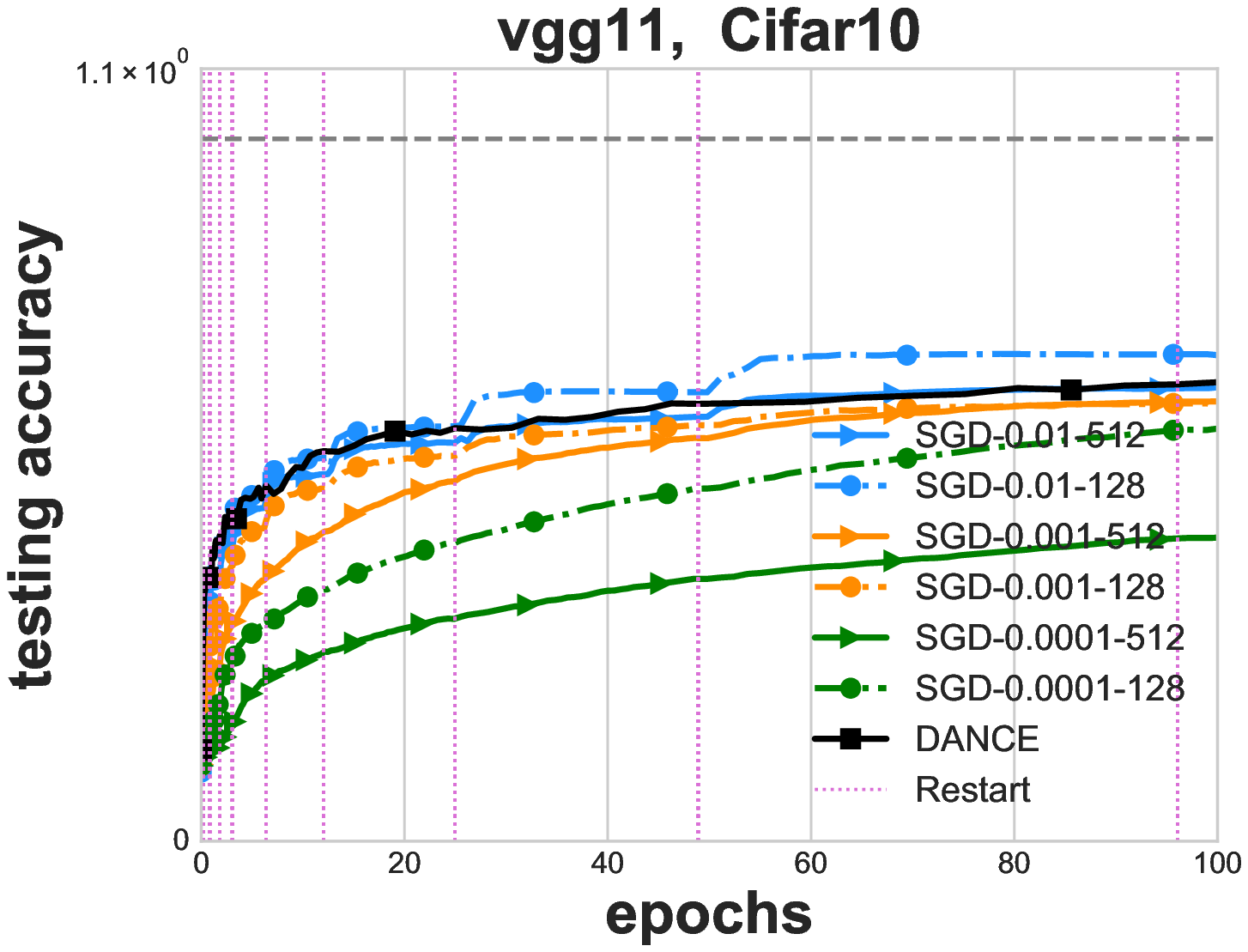}
		
		\includegraphics[width=0.22\textwidth]{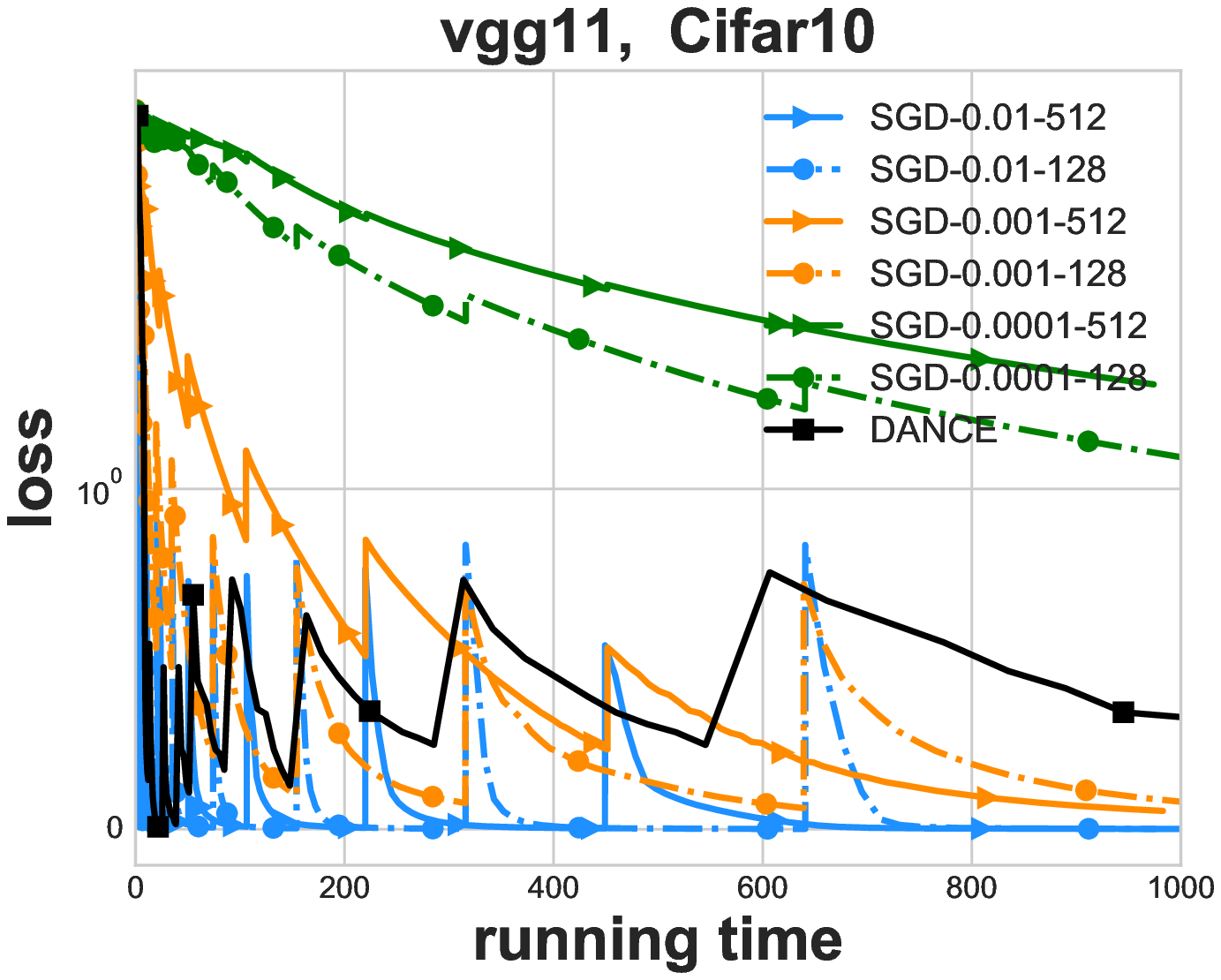}
		\includegraphics[width=0.22\textwidth]{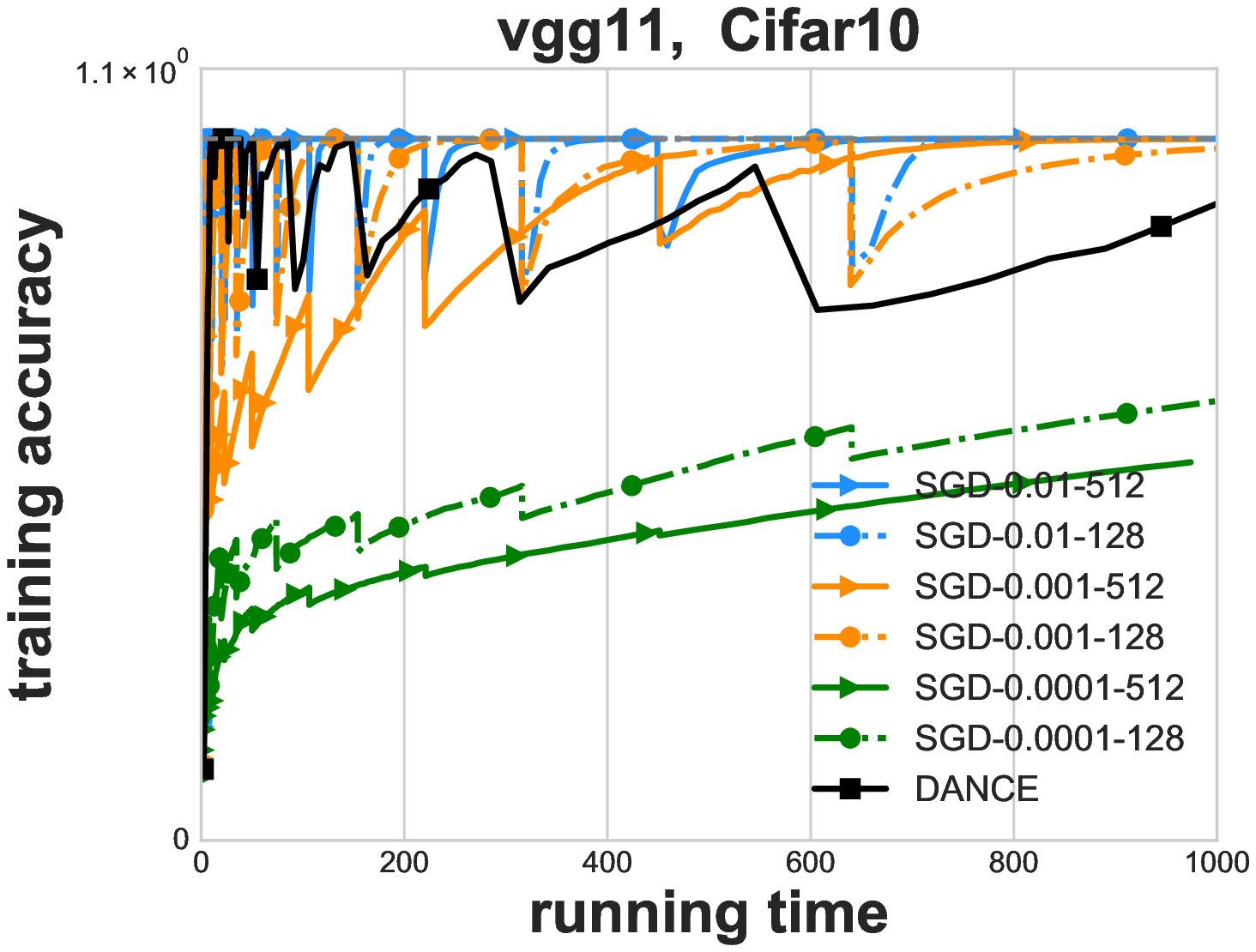}
		\includegraphics[width=0.22\textwidth]{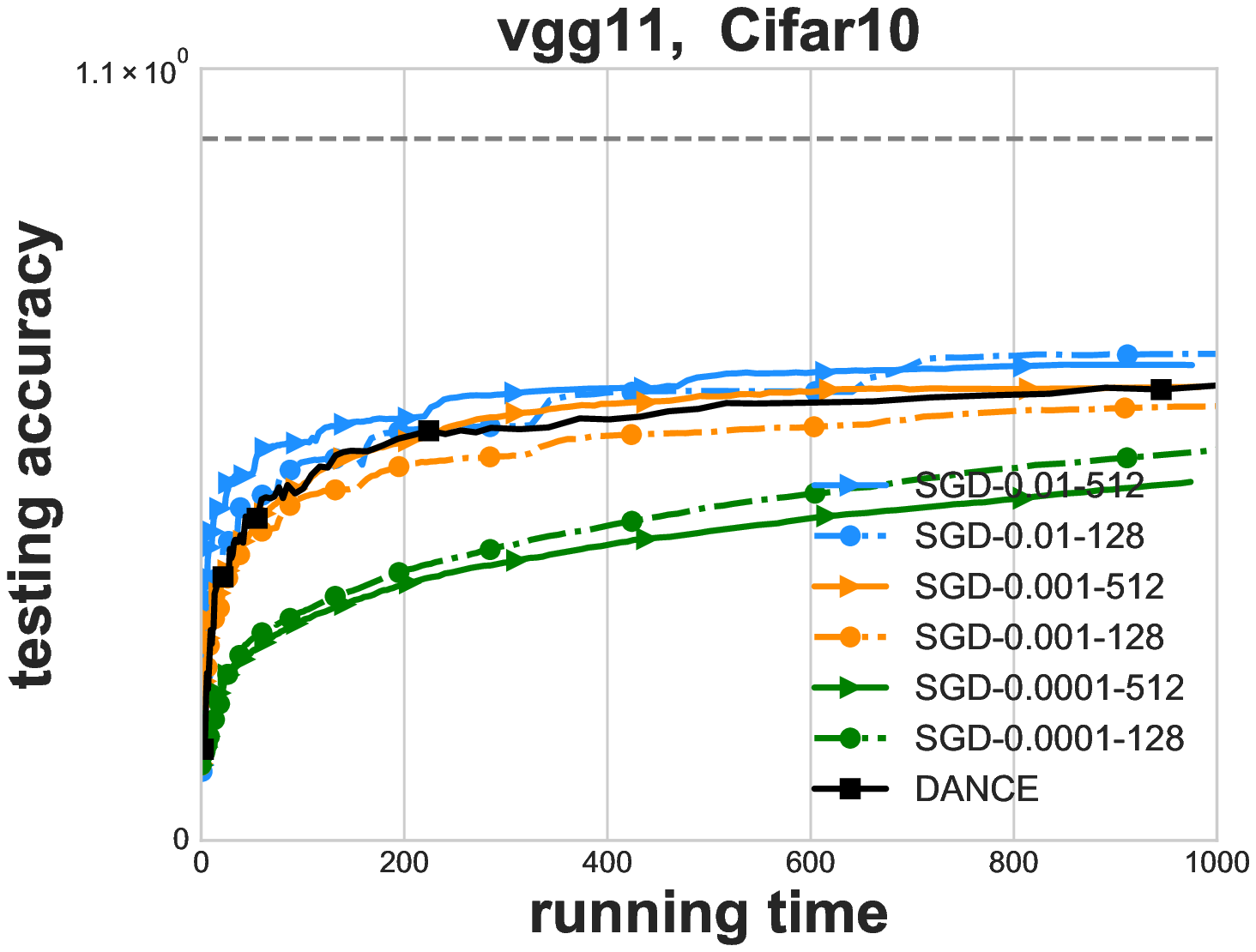}
		\vspace{-3mm}
		\caption{Comparison between DANCE and SGD with various hyper-parameters setting on Cifar10 dataset and vgg11 network. vgg11 represents {simonyan2014very} a $28$ layers convolutional neural network (see details at Appendix~\ref{sec: details concering experimental}).  Figures on the top and bottom show how loss values, training  accuracy and test accuracy are changing with respect to epochs and running time. Note that we force both algorithms to restart (double training sample size) after achieving the following number of epochs: $0.2, 0.8, 1.6. 3.2, 6.4, 12, 24, 48, 96$.  For SGD, we varies learning rate from $0.01, 0.001, 0.0001$ and batchsize from $128, 512$. 
		}		\vspace{-10pt}
		\label{fig: cifar10, vgg11, sgd}
\end{figure*}
		
\vspace{-6pt}

	\paragraph{Convex problems.}
	First, we compare DANCE with two algorithms SGD (mini-batch)\footnote{The  batch size is 10 in our experiments} and DiSCO \cite{Disco15}, for solving convex problems.	The experiments in this section are performed on a cluster with \textit{16 Xeon E5-2620 CPUs (2.40GHz)}.
		\begin{figure}[ht]
		\centering
		\includegraphics[width=0.22\textwidth]{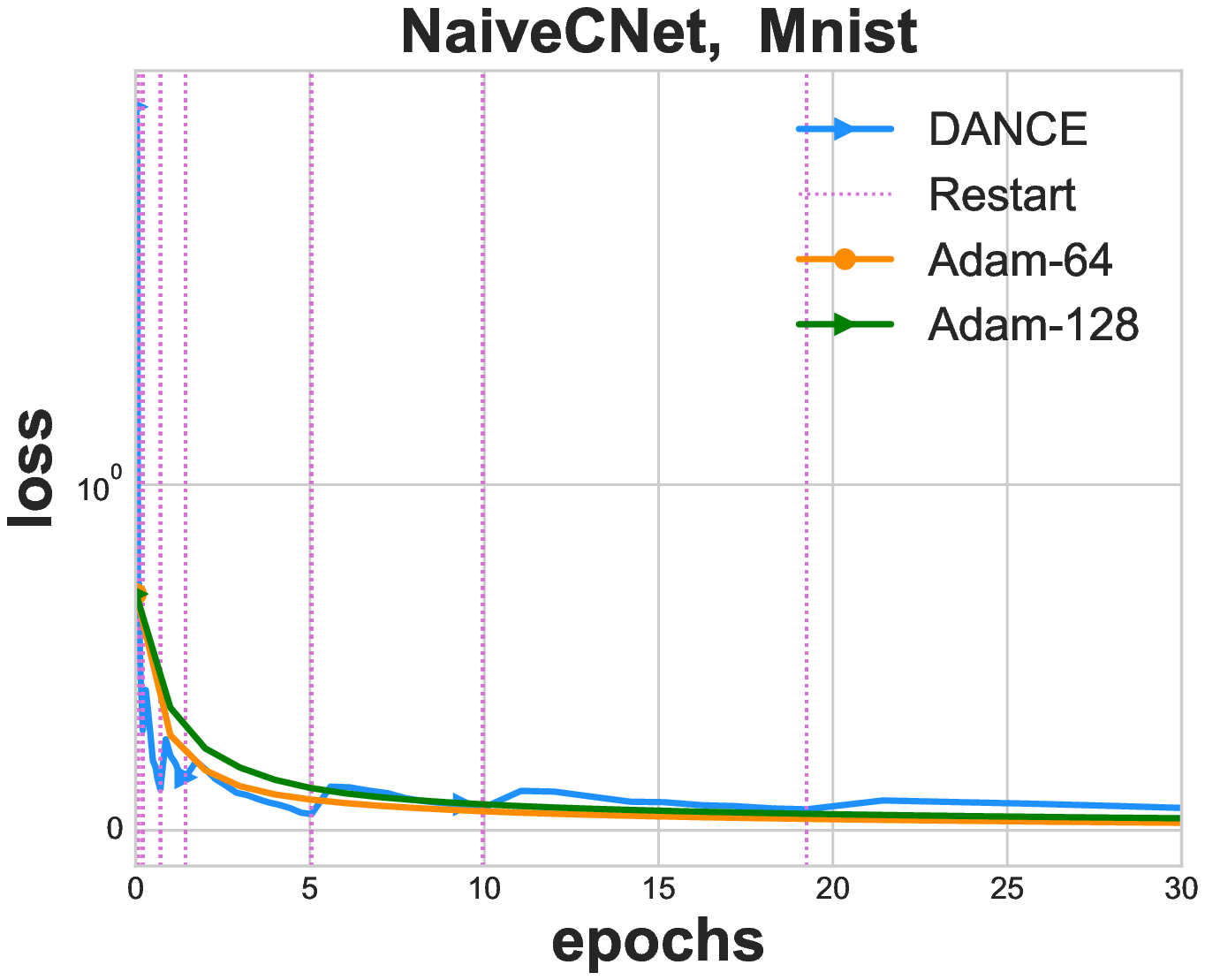}
		\includegraphics[width=0.22\textwidth]{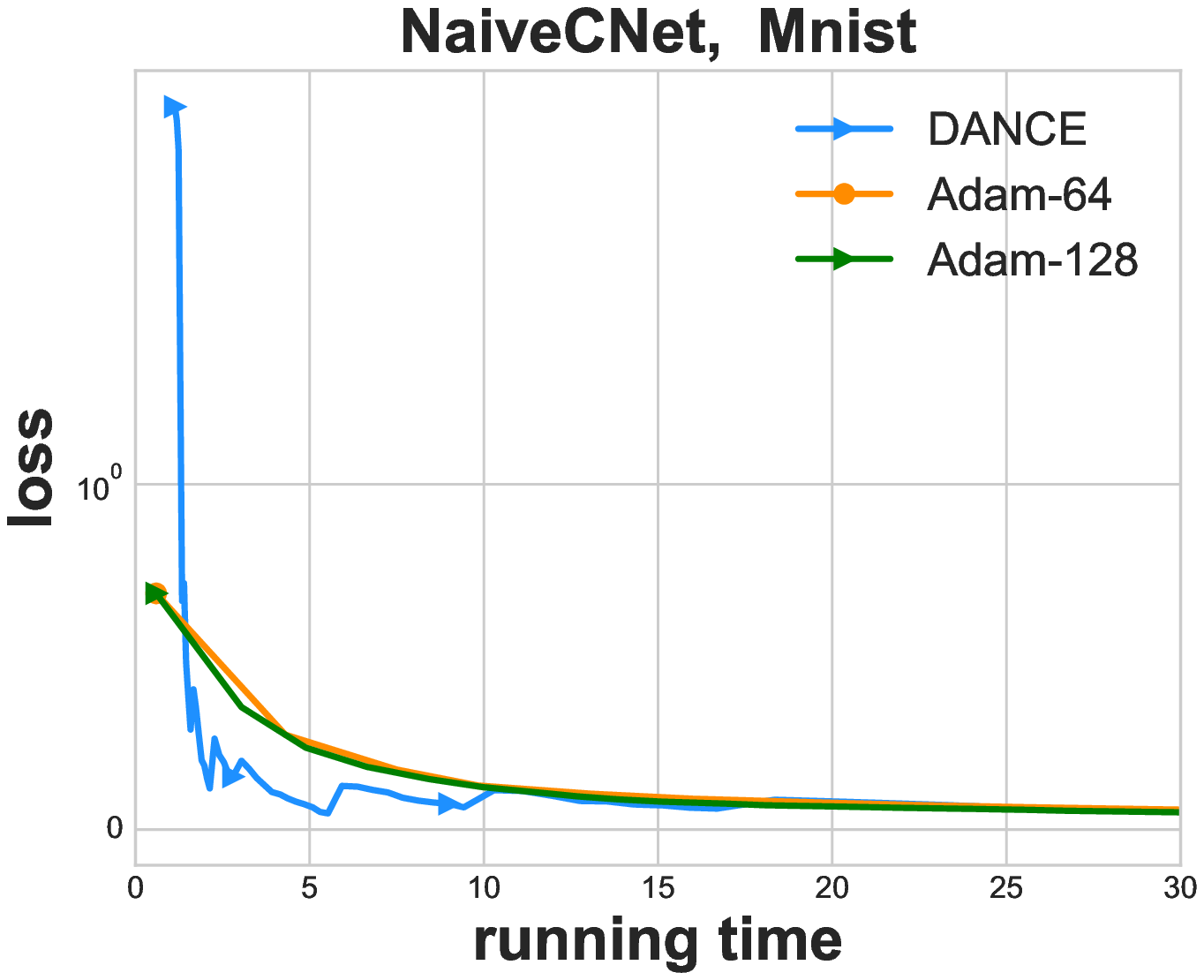}
				
		\includegraphics[width=0.22\textwidth]{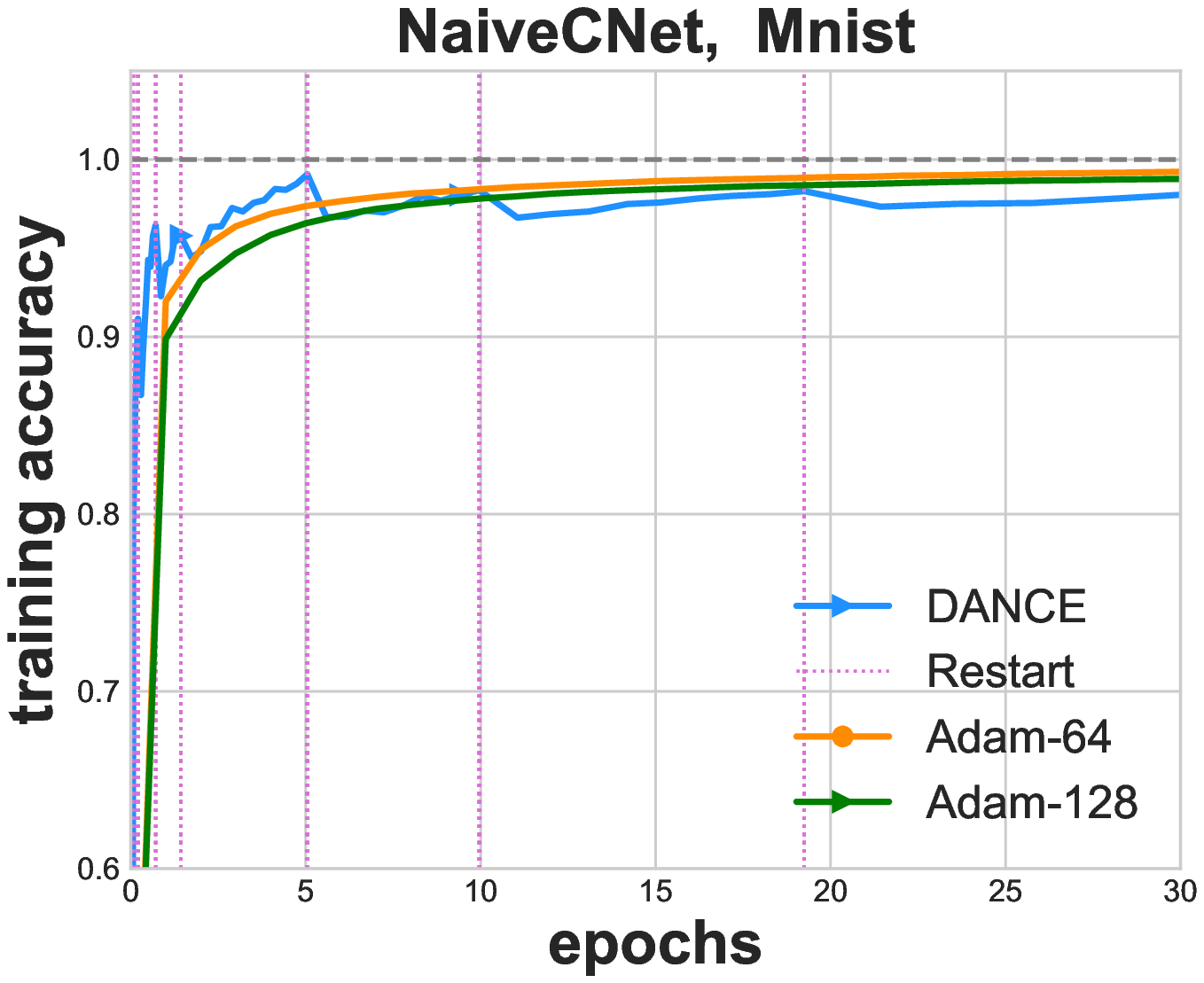}
		\includegraphics[width=0.22\textwidth]{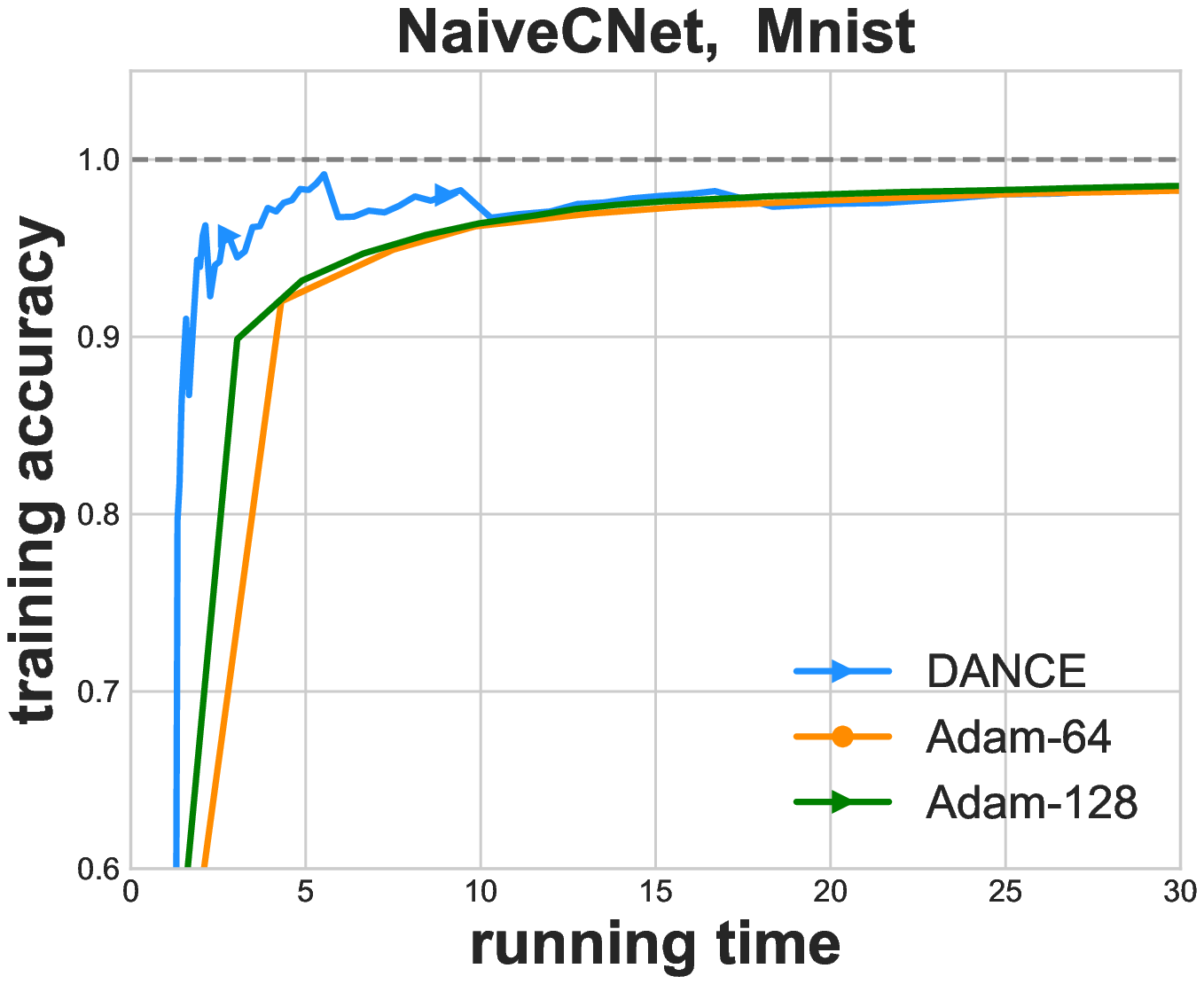}
		
		\includegraphics[width=0.22\textwidth]{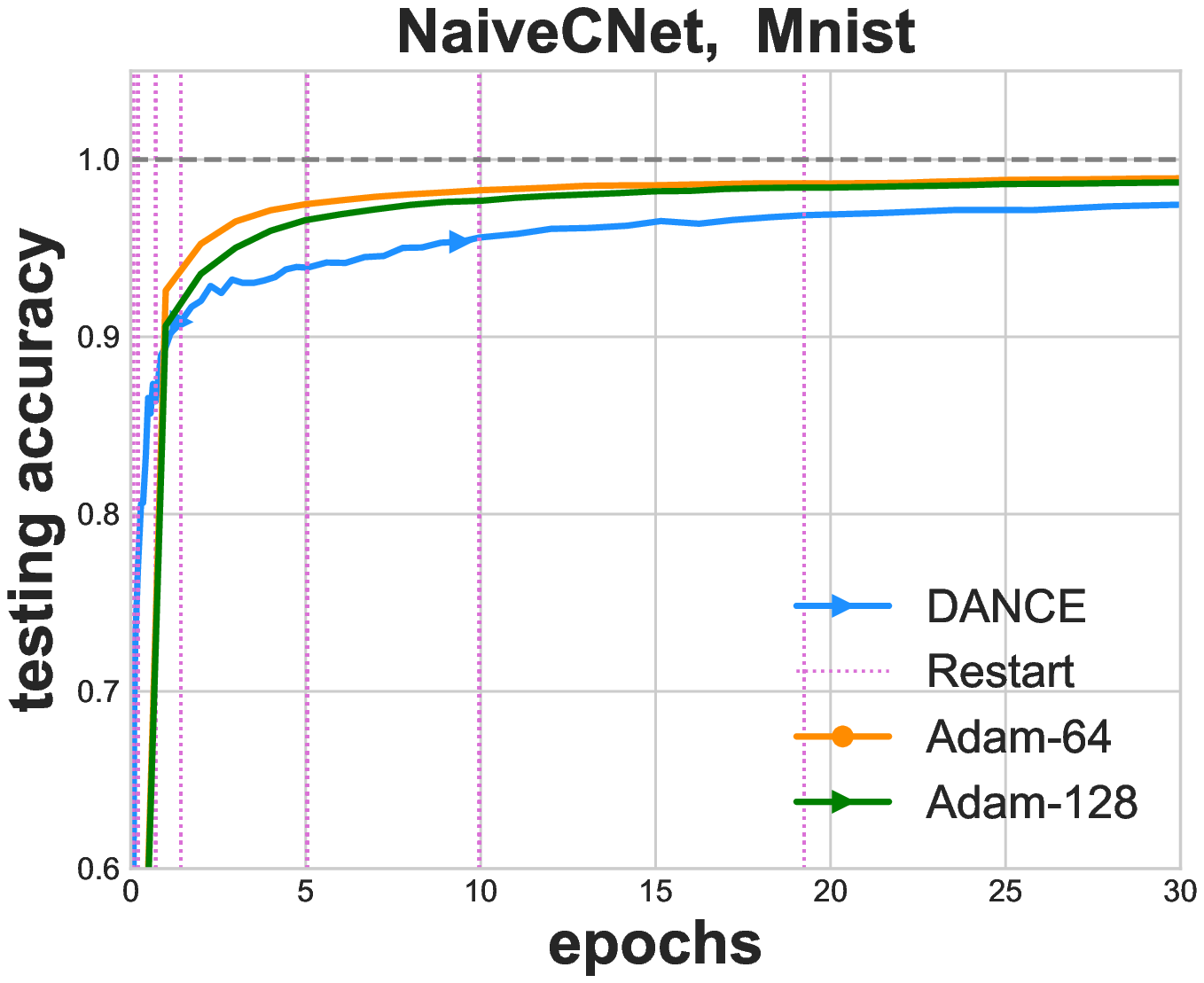}
		\includegraphics[width=0.22\textwidth]{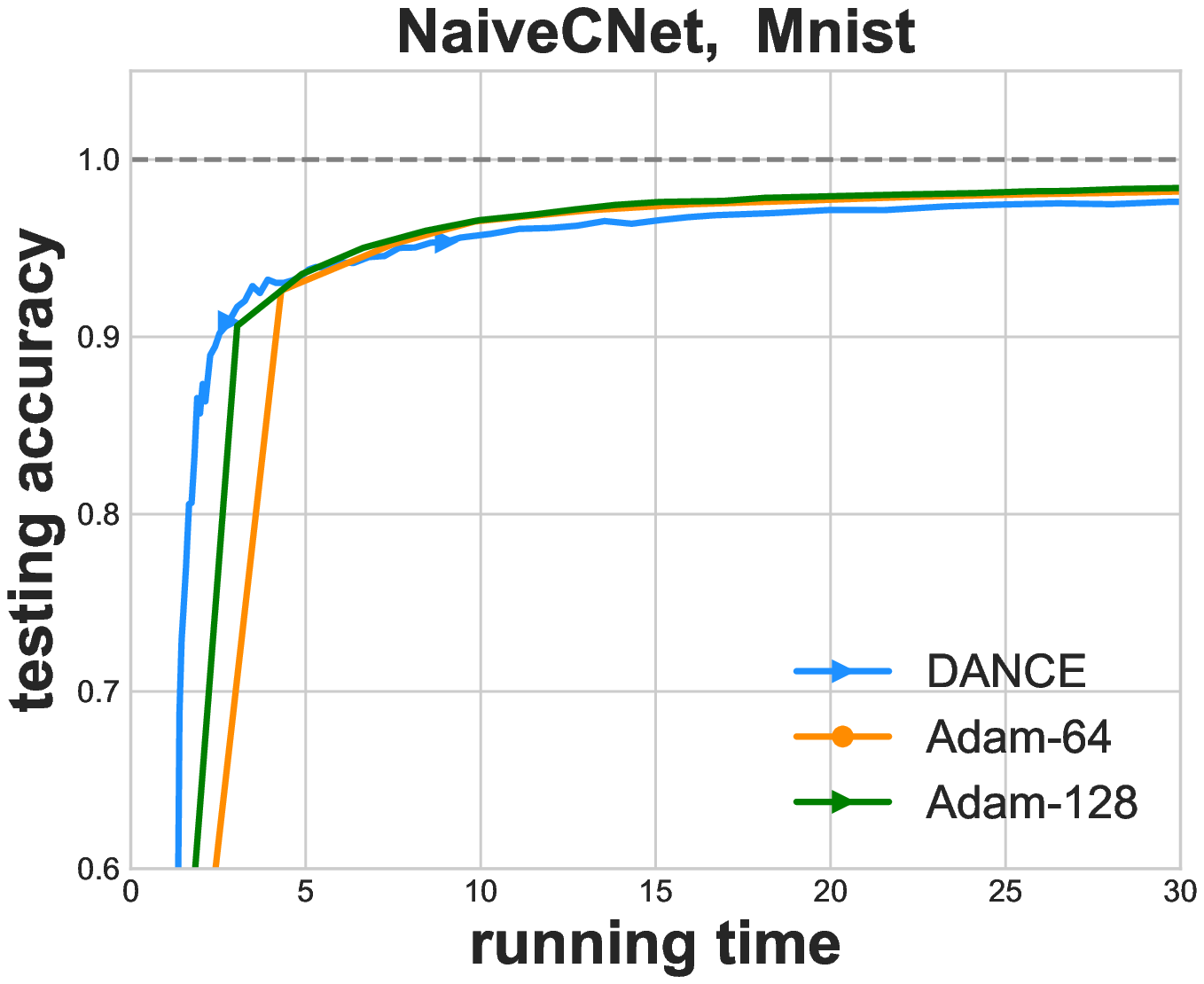}
		\vspace{-3mm}
		\caption{Comparison between DANCE and Adam on Mnist dataset and NaiveCNet. For DANCE, the initial batchsize is $1024$. For Adam, the learning rate is $10^{-4}$ and the batchsize is either $64$ or $128$. }\vspace{-10pt}
		\label{fig:mnist adam}
	\end{figure}
				
		

	We use logistic regression model for two binary classification tasks based on \textit{rcv1} and \textit{gisette} \cite{CC01a} datasets for our convex test problem. 
	We use logistic loss function defined as $f_i(w):=\log(1+\exp(-y_iw^Tx_i))$, where $x_i\in \R^d$ is data sample and $y_i\in\{-1,1\}$ is binary label corresponding to $x_i, i\in[m]$. Then we minimize the empirical loss function as \eqref{eq:ERMreg}. Note that there is a fixed $\ell_2$-regularization parameter in DiSCO and SGD and we set $c=0.1$ in \eqref{eq:ERMreg} to form the $\ell_2$-regularization parameter for our DANCE method. In Section \ref{sec:senRobust}, we numerically show that DANCE is robust w.r.t. different setting of hyper-parameters\footnote{Please see Figures \ref{fig:11}, \ref{fig:12}, \ref{fig:13} and \ref{fig:14}}. 
	
	We run our algorithm and compare algorithms with different datasets using $8$ nodes. The starting batchsize on each node for our DANCE algorithm is set to $16$ while other two algorithms go over the whole dataset at each iteration. For DANCE implementation, number of samples used to form the new ERM loss are doubled from previous iteration after each restarting. 

	In Figure~\ref{fig:convex, rcv}, we observe consistently that DANCE has a better performance over the other two methods from the beginning stages (please see Figure \ref{fig:convexComplete} for additional results). Both training and test accuracy for DANCE converges to optimality after processing a small number of samples. This observation suggests that DANCE finds a good initial solution and updates it over time. Compared with DiSCO, our restarting approach helps to reduce computational cost for the first iterations, where the second order methods usually performs less efficiently comparing to first order methods. The key difference comes from utilizing the idea of increasing sample size where DANCE goes over small number of samples and finds a suboptimal solution, and use it as a warm-start for the next stage. In this way, less passes over data is needed in the beginning but with satisfactory accuracy. On the other hand, DiSCO uses total number of samples from the beginning which some passes over data is needed in order to reach the neighborhood of global solution. Therefore, DANCE behaves efficiently and reaches the optimal solution with less passes over data. 
	\vspace{-10pt}
	\paragraph{Non-convex problems.}
	Even though the complexity analysis in Section \ref{sec4} only covers the convex case, the DANCE algorithm is also able to handle nonconvex problems efficiently. In this section, we compare our method with several stochastic first order algorithms, stochastic gradient descent (SGD), SGD with momentum (SGDMom), and Adam \cite{kingma2014adam}, on training convolution neural networks (CNNs) on two image classification datasets \textit{Mnist} and \textit{Cifar10}. The details of the datasets and the CNNs architecture applied on each dataset are presented in Appendix~\ref{sec: details concering experimental}. To perform a fair comparison with respect to first order variants, we assume data comes in an online streaming manner, e.g., only a few data samples can be accessed at the beginning, and new data samples arrives at a fixed rate. Such setting is common in industrial production, where business data is collected in a streaming fashion. We feed new data points to all algorithms only if the amount of new samples is equal to the number of existing  samples. The experiments in this section are run on an \textit{AWS p2.xlarge instance with an NVIDIA K80 GPU}.
	
	
	In Figure~\ref{fig: cifar10, vgg11, sgd}, we compare DANCE with the build-in SGD optimizer in pyTorch on Cifar dataset to train a $28$ layers CNN (Vgg11) architecture. Note that there are several hyper-parameters we need to tune for SGD to reach the best performance, such as batch size and learning rate, which are not required for DANCE. Since we have the online streaming data setting, we don't need to determine a restarting criterion. The results show that SGD is sensitive to hyper-parameters tuning, i.e., different combination of hyper-parameters affect the performance of SGD a lot and tune them well to achieve the best performance could be painful. However, our DANCE algorithm does not have such weakness and its performance is comparable to SGD with the best parameters setting. We also show that the DANCE algorithm leads to a faster decreasing on the loss value, which is similar to our convex experiments. Again, this is due to fast convergence rate of the second order methods. One could also found the additional experiments regarding the comparison with SGD with momentum and Adam in terms of Mnist with NaiveCNet at Appendix~\ref{sec: additional plots}.
	
	Regarding Figure~\ref{fig:mnist adam}, the performance of build-in Adam optimizer and our DANCE algorithm are compared regarding Mnist dataset and a $4$ layer NaiveCNet (see the details in Appendix~\ref{sec: details concering experimental}). In this experiment, we do not assume that the data samples follow an online streaming manner for Adam, i.e., the Adam algorithm does not have a restarting setting and therefore it runs on whole dataset directly. Also, this experiment is performed only on CPUs. We set the learning-rate for Adam as $10^{-4}$ and varies the running batch-size from $64$ and $128$. The evolution of loss, training accuracy, testing accuracy with respect to epochs and running time regarding the whole dataset are reported in Figure~\ref{fig:mnist adam} for different algorithms. One could observe that under the same epochs, Adam eventually achieves the better testing accuracy, while if we look at running time, our DANCE algorithm would be faster due to the distributed implementation. 
	
\vspace{-5pt}
\paragraph{Strong scaling}
Moreover, we demonstrate that our DANCE method shares a strong scaling property. As shown in Figure~\ref{fig:scaling}, whenever we increase the number of nodes, we obtain acceleration towards optimality. 
\begin{wrapfigure}{r}{0.25\textwidth}
\vspace{-14pt}
	\centering
	\includegraphics[width=0.25\textwidth]{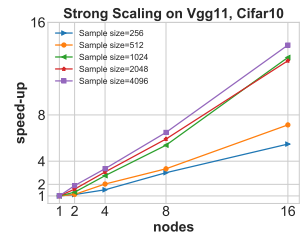}
	\caption{Performance of DANCE w.r.t. different number of nodes.} 
    \vspace{-6pt}
	\label{fig:scaling}
\end{wrapfigure}
 We use the starting batchsize from $256$ upto $4096$, and the speed-up compared to the serial run ($1$ node) is reported. It indicates that as we increase the batchsize, the speed-up becomes closer to ideal linear speed-up. 
The advantage of the setting is to utilize the large batch over multiple nodes efficiently but not sacrifice the convergence performance.
Regarding first order methods like SGD, it is hard to achieve nice linear scaling since the small batch is often required, which makes the computation time to be comparable with communication cost.\vspace{-5pt}

	\section{Conclusion}\label{sec6}	\vspace{-5pt}
	We proposed DANCE an efficient distributed Hessian free algorithm with an increasing sample size strategy to solve the empirical risk minimization problem. Our algorithm obtains a solution within the statistical accuracy of the ERM problem in very few epochs and also can be implemented in a distributed environment. 
	We analyzed the communication-efficiency of DANCE and highlighted its efficiency with respect to DiSCO \cite{Disco15} in term of communication and relative to AdaNewton and $k-$TAN methods in terms of total computational complexity. The presented numerical experiments demonstrated the fast convergence of DANCE for both convex and non-convex problems.
	\vspace{-6pt}
    \section{Acknowledgement}
    This work was supported by U.S. National Science Foundation, under award number NSF:CCF:1618717, NSF:CMMI:1663256 and NSF:CCF:1740796.
\clearpage
\newpage
\bibliographystyle{apalike}
\bibliography{reference_dance}

\clearpage

\onecolumn
\appendix
\section{Technical Proofs}
Before talking about the main results, the following lemma is used in our analysis.
\begin{lemma}\label{lem:Prop5inMokhtari_AdaNewton16}
	(Proposition 5 in \cite{Mokhtari_AdaNewton16})
	Consider the sample sets $\mathcal{S}_m$ with size $m$ and $\mathcal{S}_n$ with size $n$ such that $\mathcal{S}_m \subset \mathcal{S}_n$. Let $w_m$ is $V_m$-suboptimal solution of the risk $R_m$. If assumptions \ref{assum:smooth} and \ref{assum:selfConcor} hold, then the following is true:
	\begin{align}\label{eq:UpBndMR_nW_m}
		R_n(w_m) - R_n(w_n^*) &\leq  V_m + \tfrac{2(n-m)}{n}(V_{n-m}+V_m)+\nonumber\\
		& 2(V_m - V_n)+\tfrac{c(V_m-V_n)}{2}\tnorm{w^*},\,\,\, w.h.p.
	\end{align}
\end{lemma}
If we consider $V_n = \mathcal{O}(\tfrac{1}{n^{\gamma}})$  where $\gamma \in [0.5,1]$, and assume that $n=2m$ (or $\alpha =2$), then \eqref{eq:UpBndMR_nW_m} can be written as (w.h.p):
\begin{equation}\label{eq:UpBndMR_nW_mNew}
	R_n(w_m) - R_n(w_n^*) \leq \Big[3+\Big(1-\tfrac{1}{2^{\gamma}}\Big)\Big(2+\tfrac{c}{2}\tnorm{w^{*}}
	\Big) \Big]V_m.
\end{equation}
\subsection{Practical stopping criterion}\label{pracStopCriterion}
 Here we discuss two stopping criteria to fulfill the $10^{th}$ line of Algorithm~\ref{alg:alg1}. At first, considering $w_n^*$ is unknown in practice, we can use strong convexity inequality as $R_n(\tilde{w}_{k})-R_n(w_n^*) \leq \tfrac{1}{2cV_n} \tnorm{\nabla R_n(\tilde{w}_{k})}$ to find a stopping criterion for the inner loop, which satisfies $\|\nabla R_n(\tilde{w}_{k})\|< (\sqrt{2c})V_n$. 
	Another stopping criterion is discussed by \cite{Disco15}, using the fact that the risk $R_n$ is self-concordant. This criterion can be written as $\delta_n(\tilde{w}_k) \leq (1-\beta)\sqrt{V_n}$
	, where $\beta \leq \tfrac{1}{20}$. The later stopping criterion implies that $R_n(\tilde{w}_k) -R_n(w_n^*) \leq V_n $ whenever $V_n \leq 0.68^2$. 
For the risk $R_n$, the same as \cite{Disco15} we can define the following auxiliary function and vectors:
\begin{align}
	\omega_* (t) = -t - \log(1-t),\,\,\,\,\,\,\,\,\ 0 \leq t< 1.\label{eq:OmegaStar}\\
	\tilde{u}_n(\tilde{w}_k) = [\nabla^2 R_n(\tilde{w}_k)]^{-1/2}\nabla R_n(\tilde{w}_k), \\ 
	\tilde{v}_n(\tilde{w}_k) = [\nabla^2 R_n(\tilde{w}_k)]^{1/2}v_n.\label{eq:uTilde}
\end{align}

We can note that $\|\tilde{u}_n(\tilde{w}_k)\| = \sqrt{\nabla R_n(\tilde{w}_k)[\nabla^2 R_n(\tilde{w}_k)]^{-1}\nabla R_n(\tilde{w}_k)}$, which is the exact Newton decrement, and, the norm $\|\tilde{v}_n(\tilde{w}_k)\| = \delta_n(\tilde{w}_k)$ which is the approximation of Newton decrement (and $\tilde{u}_n(\tilde{w}_k)= \tilde{v}_n(\tilde{w}_k)$ in the case when $\epsilon_k =0$).
As a result of Theorem 1 in the study  of \cite{Disco15}, we have:
\begin{equation}\label{u_nv_n11}
	(1-\beta)\|\tilde{u}_n(\tilde{w}_k)\| \leq \|\tilde{v}_n(\tilde{w}_k)\| \leq (1+\beta)\|\tilde{u}_n(\tilde{w}_k)\|,
\end{equation}
where $\beta \leq \tfrac{1}{20}$.
Also, by the equation in \eqref{eq:uTilde}, we know that $\|\tilde{v}_n(\tilde{w}_k)\|= \delta_n(\tilde{w}_k)$.\\

As it is discussed in the section 9.6.3. of the study of \cite{boyd2004convex}, we have $\omega_*(t) \leq t^2$ for $0\leq t\leq 0.68$.

According to Theorem 4.1.13 in the study of \cite{nesterov2013introductory}, if $\|\tilde{u}_n(\tilde{w}_k)\| <1$ we have:
\begin{equation}
	\omega(\|\tilde{u}_n(\tilde{w}_k)\|) \leq R_n(\tilde{w}_k) -R_n(w_n^*) \leq \omega_*(\|\tilde{u}_n(\tilde{w}_k)\|).
\end{equation}
Therefore, if $\|\tilde{u}_n(\tilde{w}_k)\| \leq 0.68$, we have:
\begin{align}\label{stpCirPrac}
	R_n(\tilde{w}_k) -R_n(w_n^*) &\leq \omega_*(\|\tilde{u}_n(\tilde{w}_k)\|) \leq \tnorm{\tilde{u}_n(\tilde{w}_k)}\nonumber\\
	&\stackrel{\text{\eqref{u_nv_n11}}}{\leq} \tfrac{1}{(1-\beta)^2}\tnorm{\tilde{v}_n(\tilde{w}_k)}= \tfrac{1}{(1-\beta)^2} \delta^2_n(\tilde{w}_k)
\end{align}
Thus, we can note that
$\delta_n(\tilde{w}_k) \leq (1-\beta)\sqrt{V_n}$ concludes that $R_n(\tilde{w}_k) -R_n(w_n^*) \leq V_n $ when $V_n \leq 0.68^2$.
\subsection{Proof of Theorem \ref{Thm:linearIter}}\label{proofOfThm1}
According to the Theorem 1 in \cite{Disco15}1, we can derive the iteration complexity by starting from $w_m$ as a good warm start, to reach $w_n$ which is $V_n$-suboptimal solution for the risk $R_n$. By 
Corollary 1 in \cite{Disco15}, we can note that if we set $\epsilon_k$ the same as \eqref{eq:epsthm1}, after $K_n$ iterations we reach the solution $w_n$ such that $R_n(w_n)-R_n(w_n^*)\leq V_n$ where
\begin{align}\label{eq:Kiter}
	K_n&= \Big\lceil \tfrac{R_n(w_m) - R_n(w_n^*)}{\frac{1}{2}\omega (1/6)} \Big\rceil+\Big\lceil \log_2(\tfrac{2\omega(1/6)}{V_n})\Big\rceil.
\end{align}
Also, in Algorithm \ref{alg:alg2}, before the main loop, 1 communication round is needed, and in every iteration of the main loop in this algorithm, 1 round of communication happens.
According to Lemma \ref{lem:lemma4Disco}, we can note that the number of PCG steps needed to reach the approximation of Newton direction with precision $\epsilon_k$ is as following:
\begin{align}\label{eq:KiterMu}
	C_n(\epsilon_k)&= \Big \lceil \sqrt{1+\tfrac{2\mu_n}{cV_n})}\log_2\Big(\tfrac{2\sqrt{\tfrac{cV_n+M}{cV_n}}\|\nabla R_n(\tilde{w}_k)\|}{\epsilon_k}\Big)\Big\rceil\nonumber\\
	&\stackrel{\text{\eqref{eq:epsthm1}}}{=}\Big\lceil\sqrt{1+\tfrac{2\mu_n}{cV_n})}\log_2\Big(\tfrac{2(cV_n+ M)}{\beta cV_n}\Big)\Big\rceil.
\end{align}
Therefore, in every call of Algorithm \ref{alg:alg2}, the number of communication rounds is not larger than $1+C_n(\epsilon_k)$. Thus, we can note that when we start from $w_m$, which is $V_m$-suboptimal solution for the risk $R_m$, $T_n$ communication rounds are needed, where $T_n \leq K_n (1+C_n(\epsilon_k))$, to reach the point $w_n$ which is $V_n$-suboptimal solution of the risk $R_n$, which follows \eqref{eq:bndIter1}.\\
Suppose the initial sample set contains $m_0$ samples, and consider the set $\mathcal{P}=\{m_0,\alpha m_0, \alpha^2 m_0, \dots,N\}$, then with high probability with $\mathcal{T}$ rounds of communication, we reach $V_N$-suboptimal solution for the whole data set:
\begin{align}\label{eq:KiterTotal}
	\mathcal{T} &\leq \sum_{i =2}^{|\mathcal{P}|}\Big(\Big\lceil \tfrac{R_{\mathcal{P}[i]}(w_{\mathcal{P}[i-1]}) - R_{\mathcal{P}[i]}(w_{\mathcal{P}[i]}^*)}{\frac{1}{2}\omega (1/6)} \Big\rceil+ \Big\lceil \log_2(\tfrac{2\omega(1/6)}{V_{\mathcal{P}[i]}})\Big\rceil \Big)\Big(1+\Big\lceil \sqrt{1+\tfrac{2\mu_{\mathcal{P}[i]}}{cV_{\mathcal{P}[i]}})}\log_2\Big(\tfrac{2(cV_{\mathcal{P}[i]}+ M)}{\beta cV_{\mathcal{P}[i]}}\Big)\Big\rceil\Big).
\end{align}
\subsection{Proof of Corollary \ref{Cor:linearIter}} \label{proofOfCo1}
The proof of the first part is trivial. According to Lemma \ref{lem:Prop5inMokhtari_AdaNewton16}, we can find the upper bound for $R_n(w_m) - R_n(w_n^*)$, and when $\alpha =2$, by utilizing the bound \eqref{eq:UpBndMR_nW_mNew} we have:
\begin{align}\label{eq:Kiter11}
	K_n&= \Big\lceil \tfrac{R_n(w_m) - R_n(w_n^*)}{\frac{1}{2}\omega (1/6)} \Big\rceil+\Big\lceil \log_2(\tfrac{2\omega(1/6)}{V_n})\Big\rceil \nonumber\\
	& \stackrel{\text{\eqref{eq:UpBndMR_nW_mNew}}}{\leq} \underbrace{\Big\lceil\tfrac{\Big(3+\big(1-\tfrac{1}{2^{\gamma}}\big)\big(2+\tfrac{c}{2}\tnorm{w^{*}}\big) \Big)V_m}{\frac{1}{2}\omega (1/6)} \Big\rceil+\Big\lceil \log_2(\tfrac{2\omega(1/6)}{V_n})\Big\rceil}_{:= \tilde{K}_n}.
\end{align}
Therefore, we can notice that when we start from $w_m$, which is $V_m$-suboptimal solution for the risk $R_m$, with high probability with $\tilde{T}_n$ communication rounds, where $\tilde{T}_n \leq \tilde{K}(1+C_n(\epsilon_k))$, and $C_n(\epsilon_k)$ is defined in \eqref{eq:KiterMu}, we reach the point $w_n$ which is $V_n$-suboptimal solution of the risk $R_n$, which follows \eqref{eq:bndIter11}.\\
Suppose the initial sample set contains $m_0$ samples, and consider the set $\mathcal{P}=\{m_0,2 m_0, 4 m_0, \dots,N\}$, then the total rounds of communication, $\tilde{\mathcal{T}}$, to reach $V_N$-suboptimal solution for the whole data set is bounded as following:
\begin{align}
	\tilde{\mathcal{T}} \leq &\sum_{i =2}^{|\mathcal{P}|}\Big(\Big\lceil \tfrac{\Big(3+\big(1-\tfrac{1}{2^{\gamma}}\big)\big(2+\tfrac{c}{2}\tnorm{w^{*}}
		\big) \Big)V_{\mathcal{P}[i-1]}}{\frac{1}{2}\omega (1/6)} \Big\rceil  + \Big\lceil \log_2(\tfrac{2\omega(1/6)}{V_{\mathcal{P}[i]}})\Big\rceil \Big)\Big\rceil \Big)\nonumber\\&\Big(\Big\lceil \sqrt{1+\tfrac{2\mu}{cV_{\mathcal{P}[i]}})}
	\log_2\Big(\tfrac{2(cV_{\mathcal{P}[i]}+ M)}{\beta cV_{\mathcal{P}[i]}}\Big)\Big\rceil\Big)\nonumber \\
	\leq &\Bigg(\log_2{\tfrac{N}{m_0}} +\Big( \tfrac{\Big(3+\big(1-\tfrac{1}{2^{\gamma}}\big)\big(2+\tfrac{c}{2}\tnorm{w^{*}}
		\big) \Big)}{\frac{1}{2}\omega (1/6)} \tfrac{1-(\tfrac{1}{2^{\gamma}})^{\log_2{\tfrac{N}{m_0}}}}{1-\tfrac{1}{2^{\gamma}}}V_{m_0}\Big) \nonumber \\
	&+ \sum_{i =2}^{|\mathcal{P}|} \Big\lceil \log_2(\tfrac{2\omega(1/6)}{V_{\mathcal{P}[i]}})\Big\rceil \Bigg ) \left(\Big\lceil \sqrt{1+\tfrac{2\mu}{cV_N})}\log_2\Big(\tfrac{2}{\beta} +\tfrac{2M}{\beta c}.\tfrac{1}{V_N}\Big)\Big\rceil\right)\nonumber \\
	\leq &\Bigg(2\log_2{\tfrac{N}{m_0}} +\Big( \tfrac{\Big(3+\big(1-\tfrac{1}{2^{\gamma}}\big)\big(2+\tfrac{c}{2}\tnorm{w^{*}}
		\big) \Big)}{\frac{1}{2}\omega (1/6)} \tfrac{1-(\tfrac{1}{2^{\gamma}})^{\log_2{\tfrac{N}{m_0}}}}{1-\tfrac{1}{2^{\gamma}}}V_{m_0}\Big) \nonumber \\
	&+  \log_2{\tfrac{N}{m_0}}\log_2(\tfrac{2\omega(1/6)}{V_N}) \Bigg ) \left(\Big\lceil \sqrt{1+\tfrac{2\mu}{cV_N})}\log_2\Big(\tfrac{2}{\beta} +\tfrac{2M}{\beta c}.\tfrac{1}{V_N}\Big)\Big\rceil\right)\nonumber,\,\,\,\text{w.h.p}.
\end{align}
where $\mu = \max\{\mu_{m_0},\mu_{\alpha m_0},\dots,\mu_N\}$. 

\subsection{Proof of Corollary \ref{Cor:totalNumOfCommunication}} \label{proofOfCo2}
By Corollary \ref{Cor:linearIter}, it is shown that.} after $\tilde{\mathcal{T}}$ rounds of communication we reach a point with the statistical accuracy of $V_N$ of the full training set, where with high probability $\tilde{\mathcal{T}}$ is bounded above by
	\begin{align}\label{eq:totBndIteTot1}
		\tilde{\mathcal{T}} \leq  &\Bigg(2\log_2{\tfrac{N}{m_0}}+  \log_2{\tfrac{N}{m_0}}\log_2(\tfrac{2\omega(1/6)}{V_N}) \nonumber\\
		&+\Big( \tfrac{\Big(3+\big(1-\tfrac{1}{2^{\gamma}}\big)\big(2+\tfrac{c}{2}\tnorm{w^{*}}
			\big) \Big)}{\frac{1}{2}\omega (1/6)} \tfrac{1-(\tfrac{1}{2^{\gamma}})^{\log_2{\tfrac{N}{m_0}}}}{1-\tfrac{1}{2^{\gamma}}}V_{m_0}\Big)
		\Bigg ) \nonumber \\
		&\left(1+\Big\lceil \sqrt{(1+\tfrac{2\mu}{cV_N})}\log_2\Big(\tfrac{2}{\beta} +\tfrac{2M}{\beta c}.\tfrac{1}{V_N}\Big)\Big\rceil\right),
	\end{align}
	where $m_0$ is the size of the initial training set. Note that the result in \eqref{eq:totBndIteTot1} implies that the overall rounds of communication to obtain the statistical accuracy of the full training set is $\tilde{\mathcal{T}} = \tilde{\mathcal{O}}(\gamma(\log_2{N})^2\sqrt{N^{\gamma}}\log_2{N^{\gamma}})$. Hence, when $\gamma =1$, we have $\tilde{\mathcal{T}} = \tilde{\mathcal{O}}((\log_2{N})^3\sqrt{N})$, and for $\gamma = 0.5$, the result is $\tilde{\mathcal{O}} = \mathcal{O}((\log_2{N})^3 N^{\frac{1}{4}})$.

\subsection{Proof of Theorem \ref{Thm:totalComplexity}} \label{proofOfThm2}


By using Woodbury Formula \cite{ma2016distributed,Press03}, every PCG iteration has the cost of $\mathcal{O}(d^2 )$. The reason comes from the fact that the total computations needed for applying Woodbury Formula are $\mathcal{O}(\Lambda^3)$, where $\Lambda = \max\{|\mathcal{A}_{m_0}|,|\mathcal{A}_{\alpha m_0}|,\dots,|\mathcal{A}_N|\}$, and  $\Lambda \leq 100$. The complexity of every iteration of PCG is $\mathcal{O}(d^2 +\Lambda^3 )$ or equivalently $\mathcal{O}(d^2)$, and by using the results in the proof of Corrolary \ref{Cor:totalNumOfCommunication}, the total complexity of DANCE is $\tilde{\mathcal{O}}((\log_2(N))^3N^{1/4}d^2)$.

\section{Details Concerning Experimental Section}
\label{sec: details concering experimental}

In this section, we describe our datasets and implementation details. Along this work, we select four datasets to demonstrate the efficiency of our Algorithm~\ref{alg:alg1}. Two of them are for convex loss case for a binary classification task using logistic model and the other two are non-convex loss for a multi-labels classification task using convolutional neural networks. The details of the dataset are summarized in Table~\ref{tab:datasets}.
\begin{table}[htb]
	\centering
	\caption{Summary of two binary classification datasets and two multi-labels classification datasets}
	\begin{tabular}{l|ccc}
		\hline
		
		\hline
		\textbf{Dataset} & \textbf{\# of samples } & \textbf{\# of features}& \textbf{\# of categories} \\
		\hline
		rcv1& 20,242& 47,326 & 2\\
		gisette& 7,242& 5,000 & 2\\
		Mnist& 60,000& 28*28 & 10\\
		Cifar10& 60,000& 28*28*3 & 10\\
		\hline
		
		\hline
	\end{tabular}
	
	\label{tab:datasets}
\end{table}

In terms of non-convex cases, we select two convolutional structure for the demonstration. NaiveCNet is a simple two convolutional layer network for Mnist dataset, and Vgg11 is a relative larger model with $8$ convolutional layers. The details of the network architecture is summarized in Table~\ref{tab:cnn architecture nonconvex}. Note that for vgg11, a batch normalization layer is applied right after each convolutional layer.

\begin{table}[htb]
	\centering
	\caption{Summary of two convolutional neural network architecture}
	\begin{tabular}{r|ll}
		\hline
		
		\hline
		\textbf{Architecture}&\textbf{NaiveCNet} & \textbf{Vgg11} \\
		\hline
		conv-1 &  $(5\times5\times16)$,  stride=1& $(3\times3\times64)$, stride=1\\
		max-pool-1 &  $(2\times2)$, stride=2& $(2\times2)$,stride=2\\
		conv- 2&  $(5\times5\times32)$,  stride=1& $(3\times3\times128)$, stride=1\\
		max-pool-2 &  $(2\times2)$, stride=2& $(2\times2)$, stride=2\\
		conv- 3&  & $(3\times3\times256)$, stride=1\\
		max-pool-3 &  & $(2\times2)$, stride=2\\
		conv- 4&  & $(3\times3\times256)$\\
		max-pool-4 &  & $(2\times2)$, stride=2\\
		conv- 5&  & $(3\times3\times512)$, stride = 1\\
		max-pool-5 &  & $(2\times2)$, stride=2\\
		conv- 6&  & $(3\times3\times512)$, stride = 1\\
		max-pool-6 &  & $(2\times2)$, stride=2\\
		conv- 7&  & $(3\times3\times512)$, stride = 1\\
		max-pool-7 &  & $(2\times2)$, stride=2\\
		conv- 8&  & $(3\times3\times512)$, stride = 1\\
		max-pool-8 &  & $(2\times2)$, stride=2\\
		fc & & $512$\\
		output & 10 & 10\\
		\hline
		
		\hline
	\end{tabular}
	\label{tab:cnn architecture nonconvex}
\end{table}
	\begin{figure*}[h]
		\centering
		\includegraphics[width=0.245\textwidth]{eps/data_rcv1_train_yLabelP_TrainAccuracy.pdf}
		\includegraphics[width=0.245\textwidth]{eps/data_rcv1_train_yLabelP_TestAccuracy.pdf}
		\includegraphics[width=0.245\textwidth]{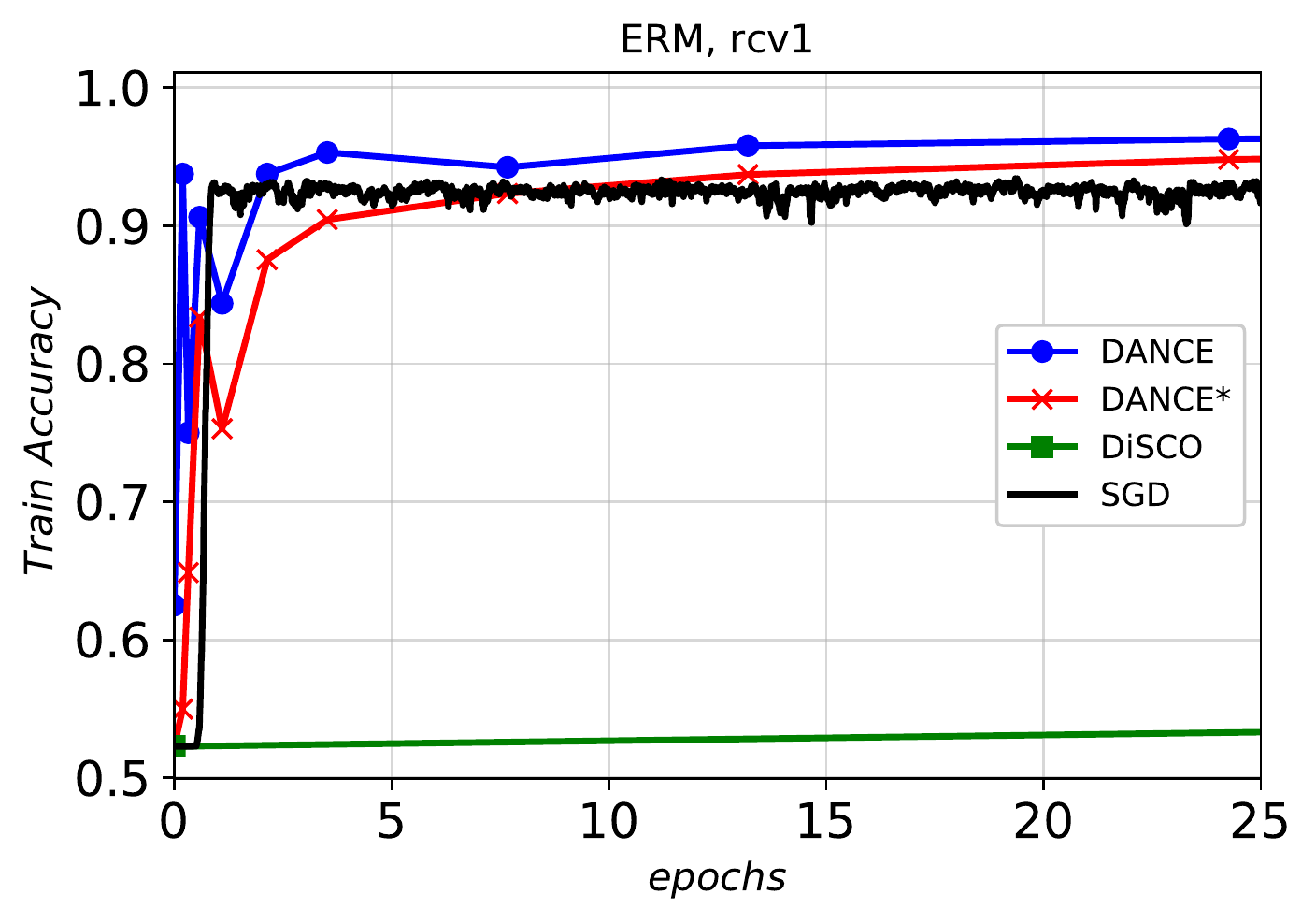}
		\includegraphics[width=0.245\textwidth]{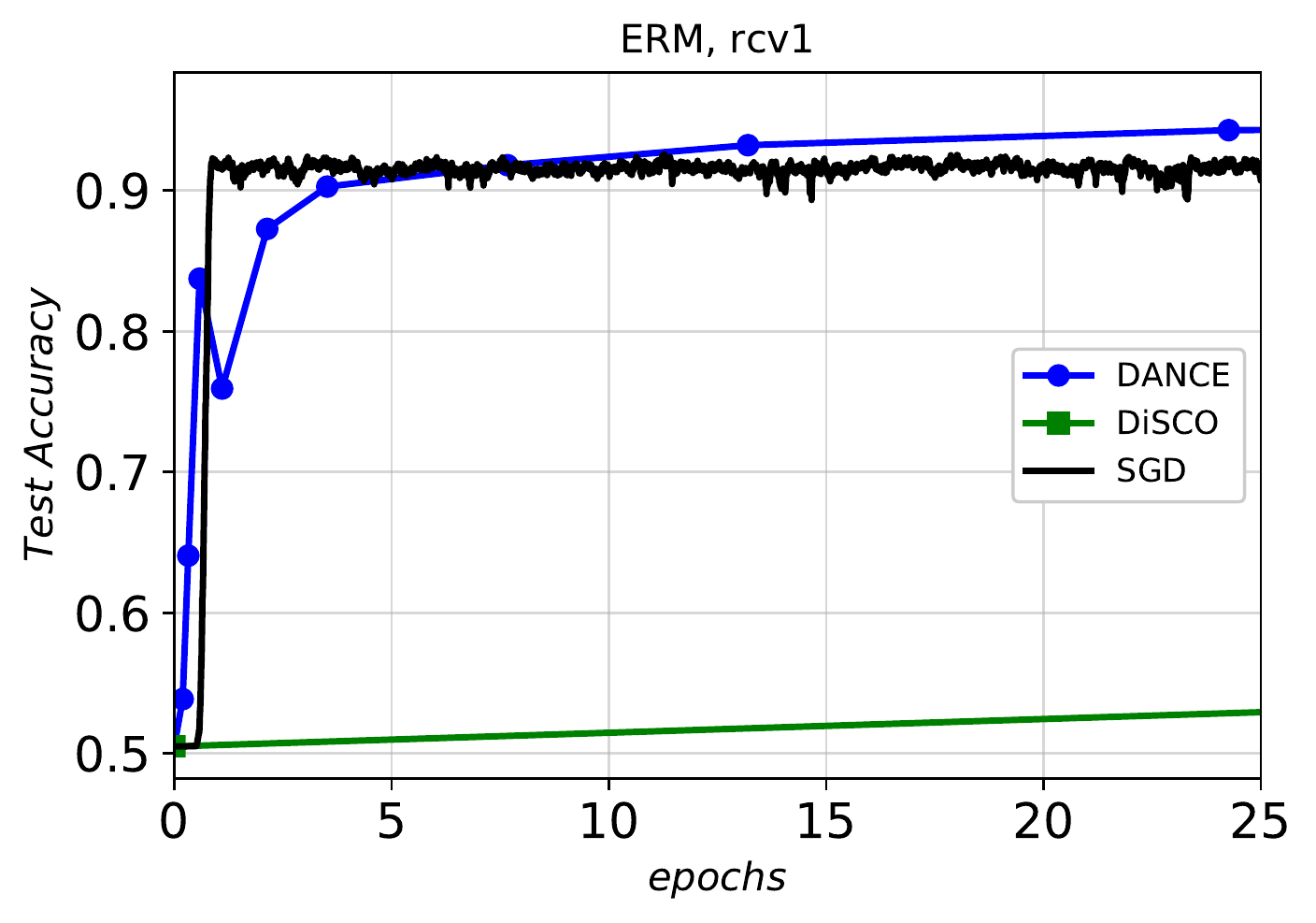}
		
		\includegraphics[width=0.245\textwidth]{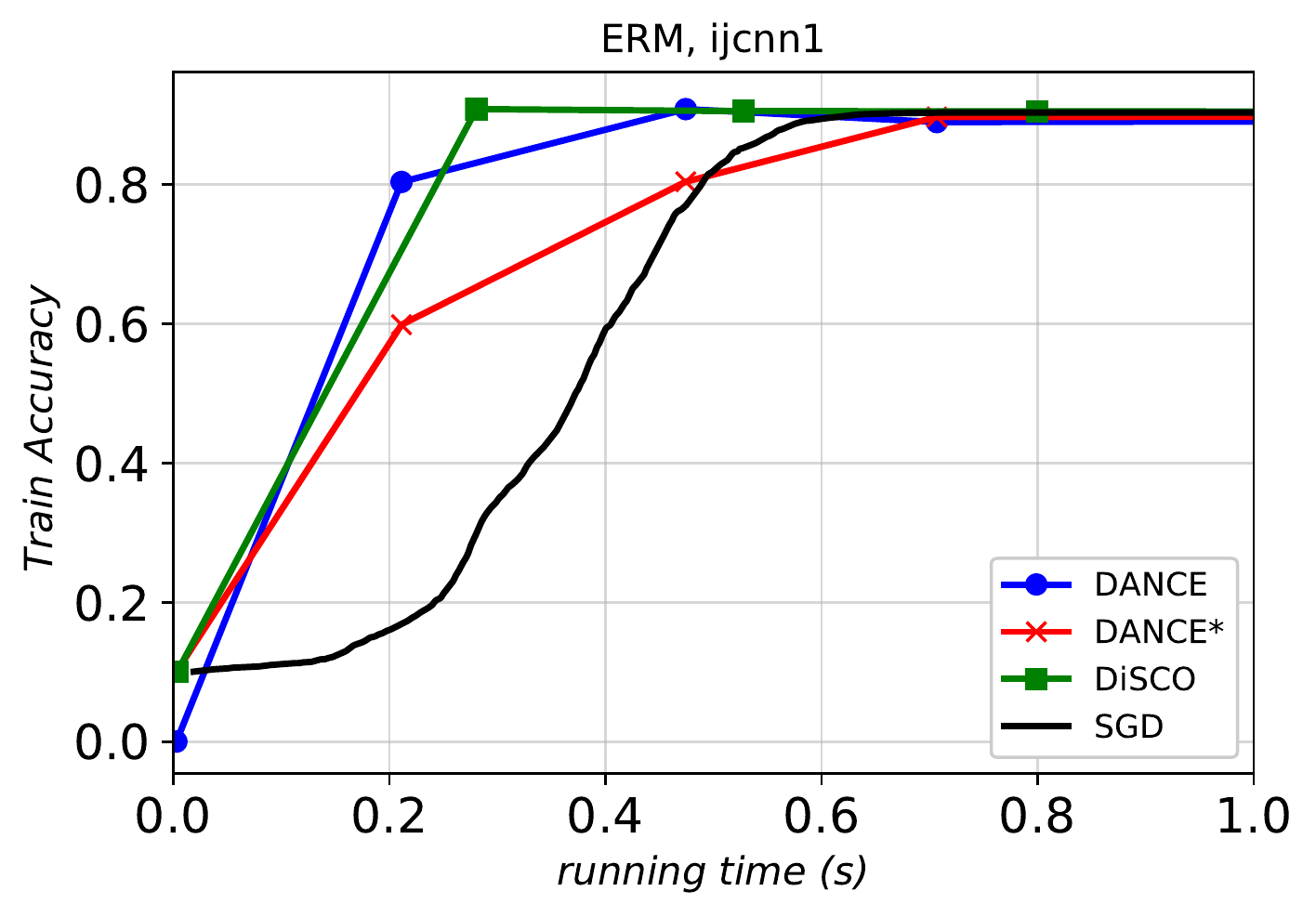}
		\includegraphics[width=0.245\textwidth]{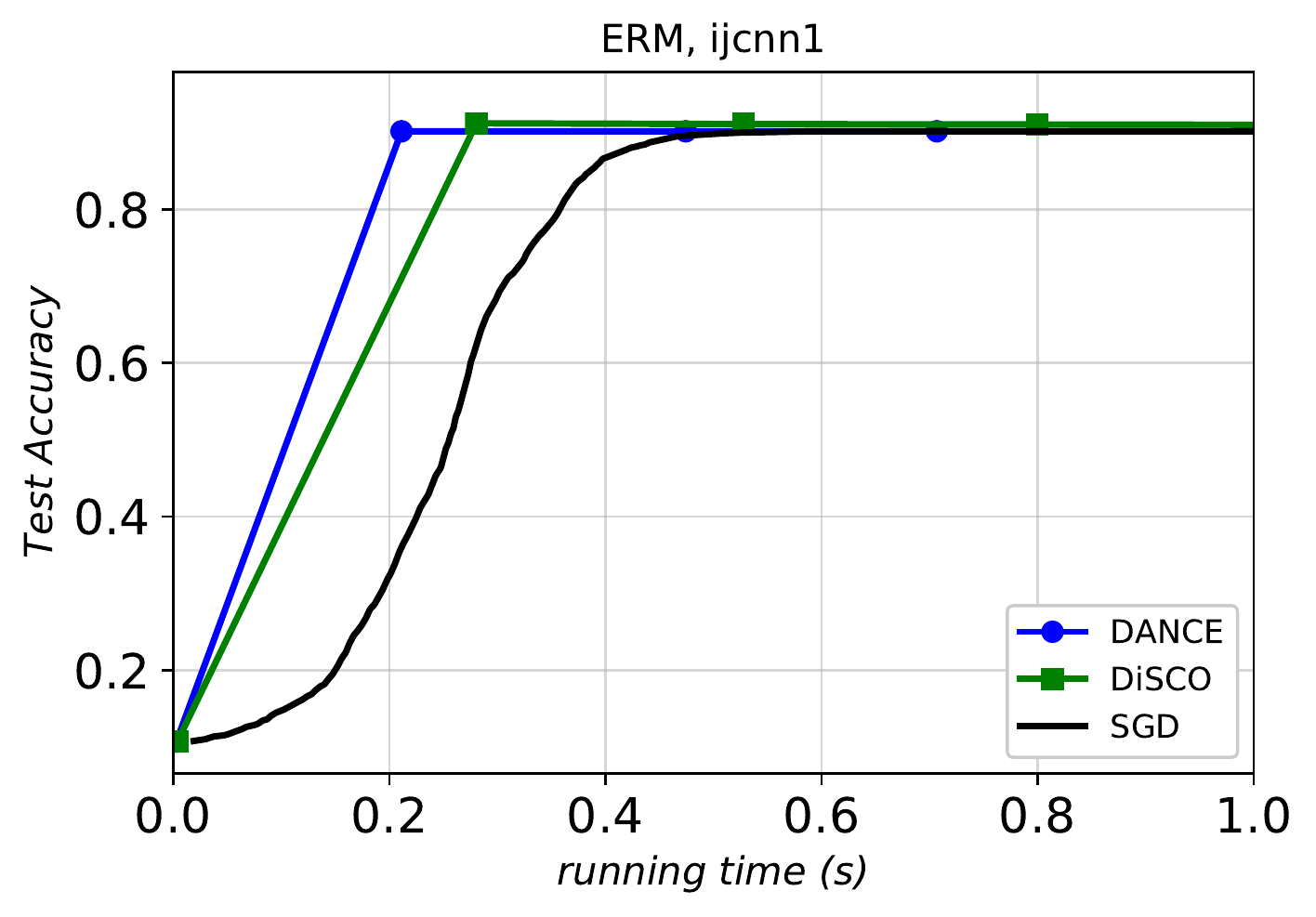}		
		\includegraphics[width=0.245\textwidth]{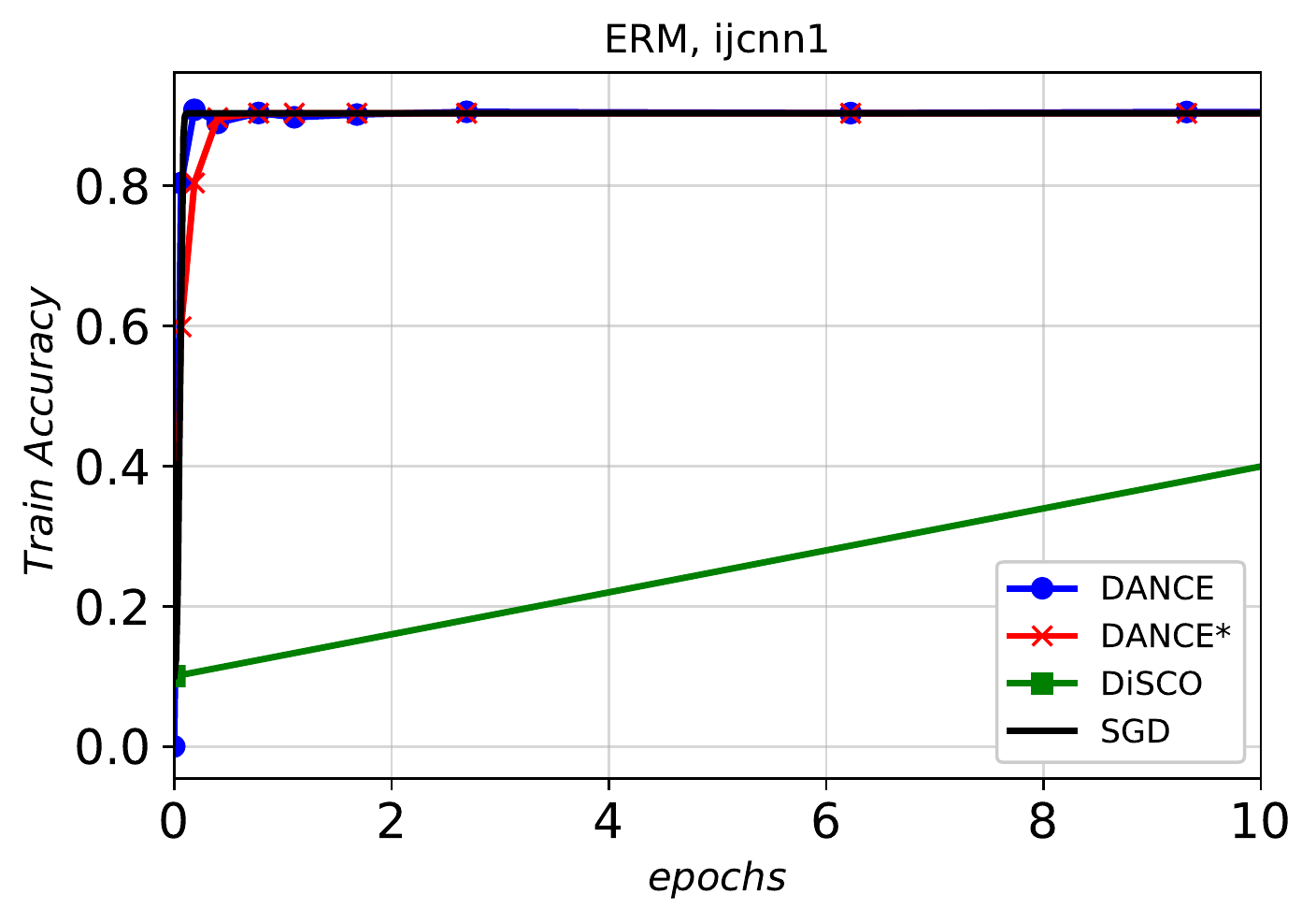}
		\includegraphics[width=0.245\textwidth]{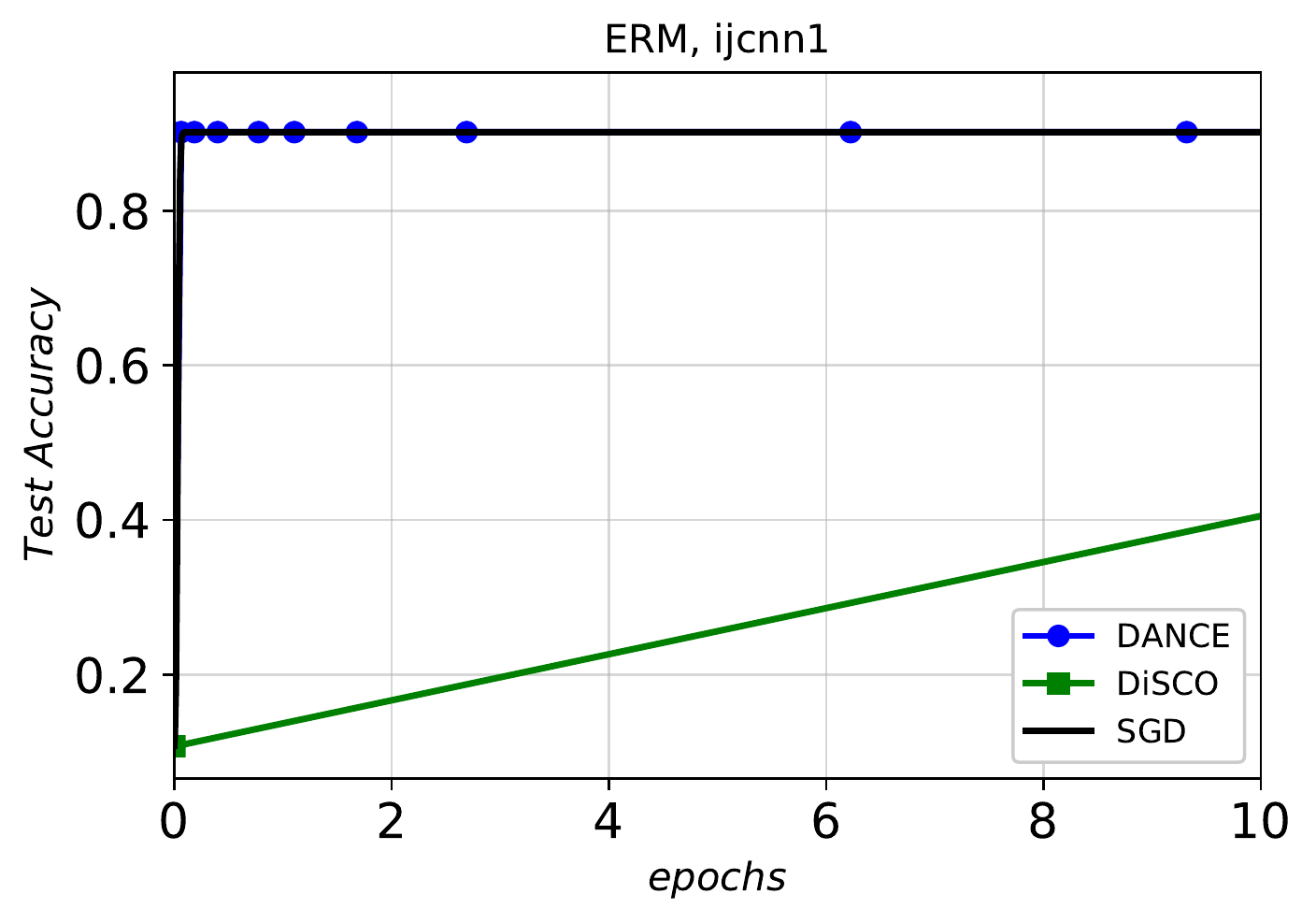}
		
		\includegraphics[width=0.245\textwidth]{eps/data_gisette_scale_yLabelP_TrainAccuracy.pdf}
		\includegraphics[width=0.245\textwidth]{eps/data_gisette_scale_yLabelP_TestAccuracy.pdf}
		\includegraphics[width=0.245\textwidth]{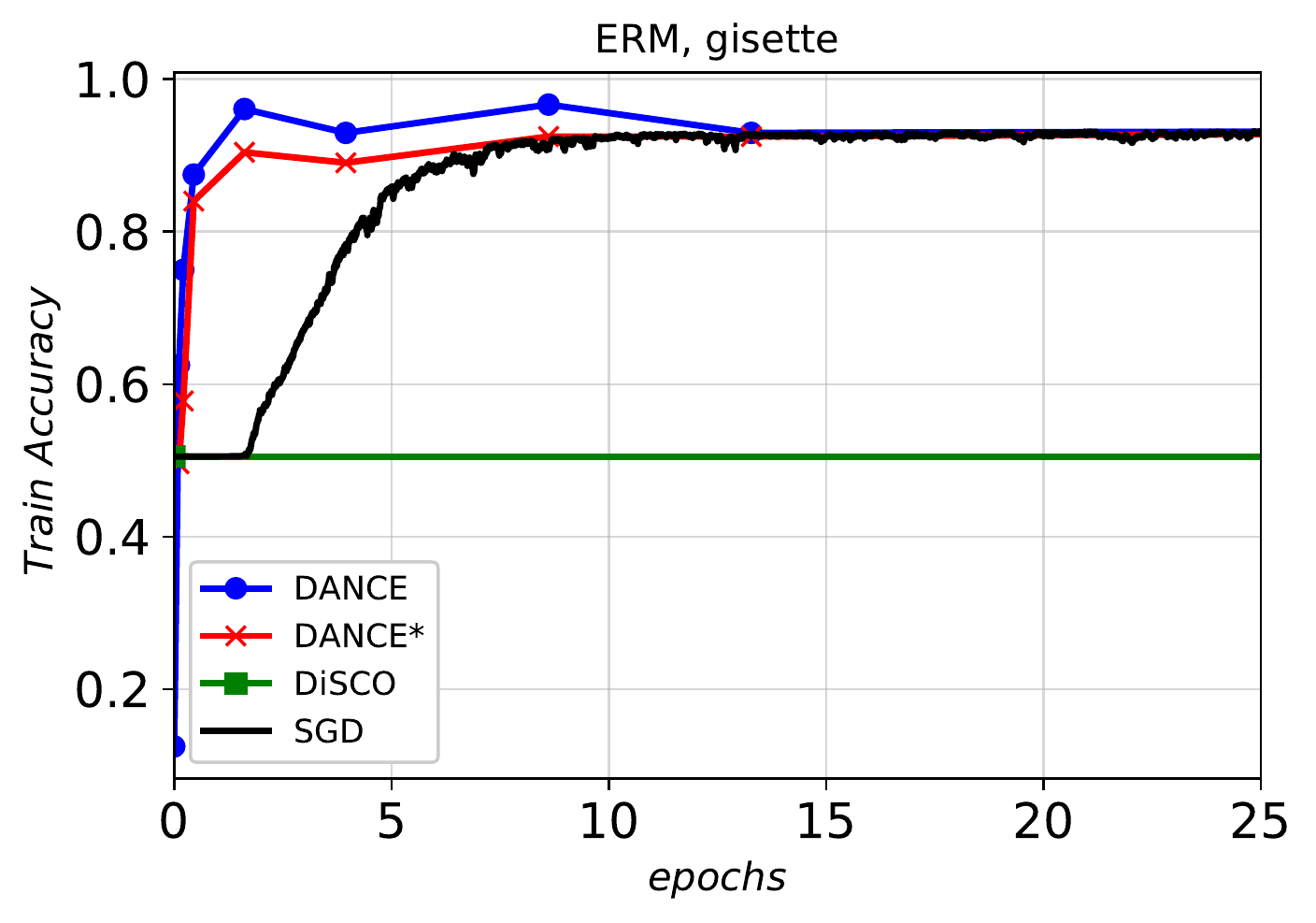}
		\includegraphics[width=0.245\textwidth]{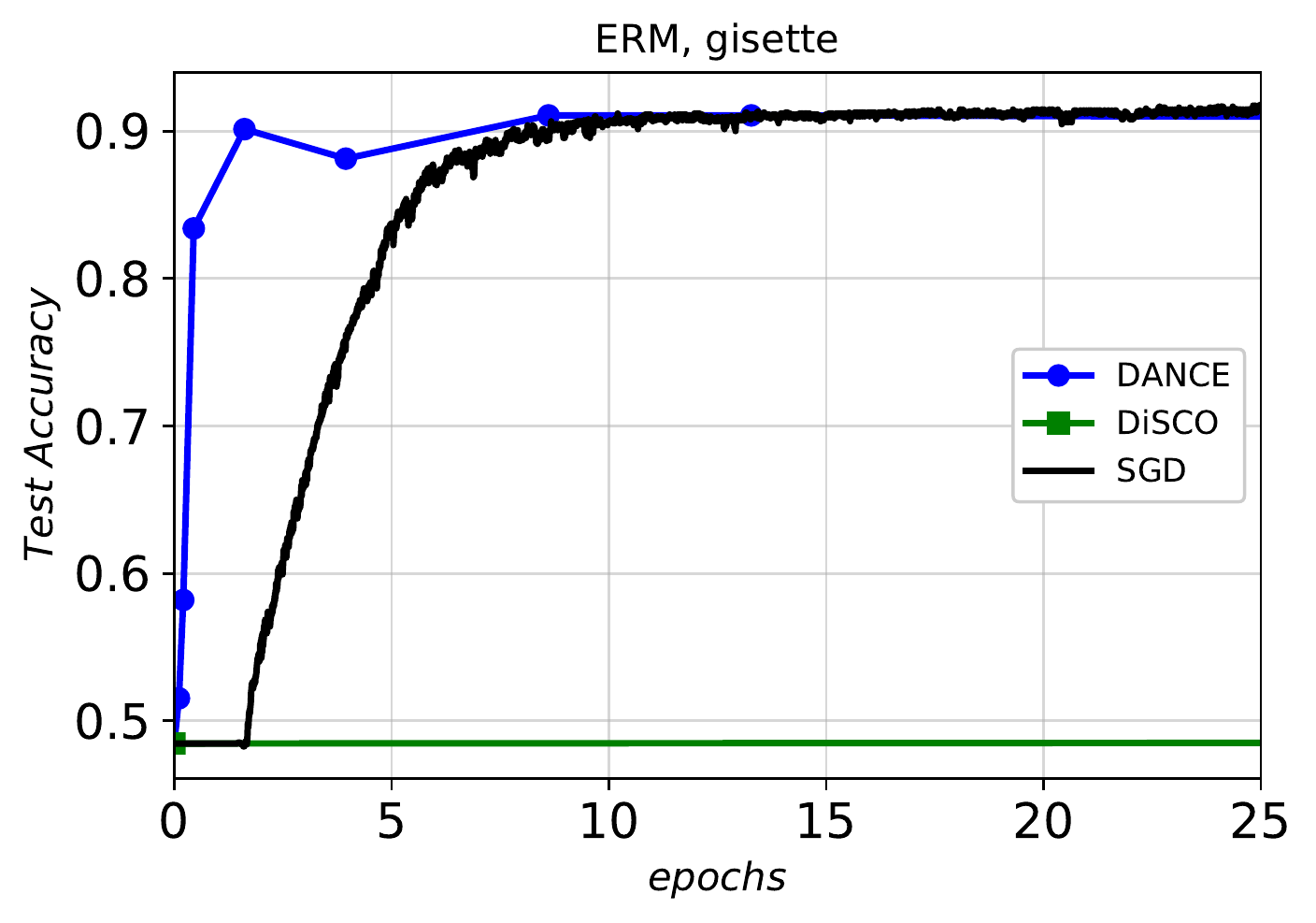}		
		\vspace{-15pt}
		\caption{Performance of different algorithms on a logistic regression problem with rcv1, ijcnn1 and gissete datasets. 
			In the left two figures, the plot \textit{DANCE*} is the training accuracy based on the entire training set, while the plot \textit{DANCE} represents the training accuracy based on the current sample size. }
		\label{fig:convexComplete}
	\end{figure*}

\section{Additional Plots}
\label{sec: additional plots}
Besides the plots in Section \ref{sec:NumExp}, we also experimented different data sets, and the other corresponding settings are described in the main body.


\subsection{Sensitivity Analysis of DANCE's Parameters}\label{sec:senRobust}
\begin{figure*}[ht]
	\centering
	\includegraphics[width=0.30\textwidth]{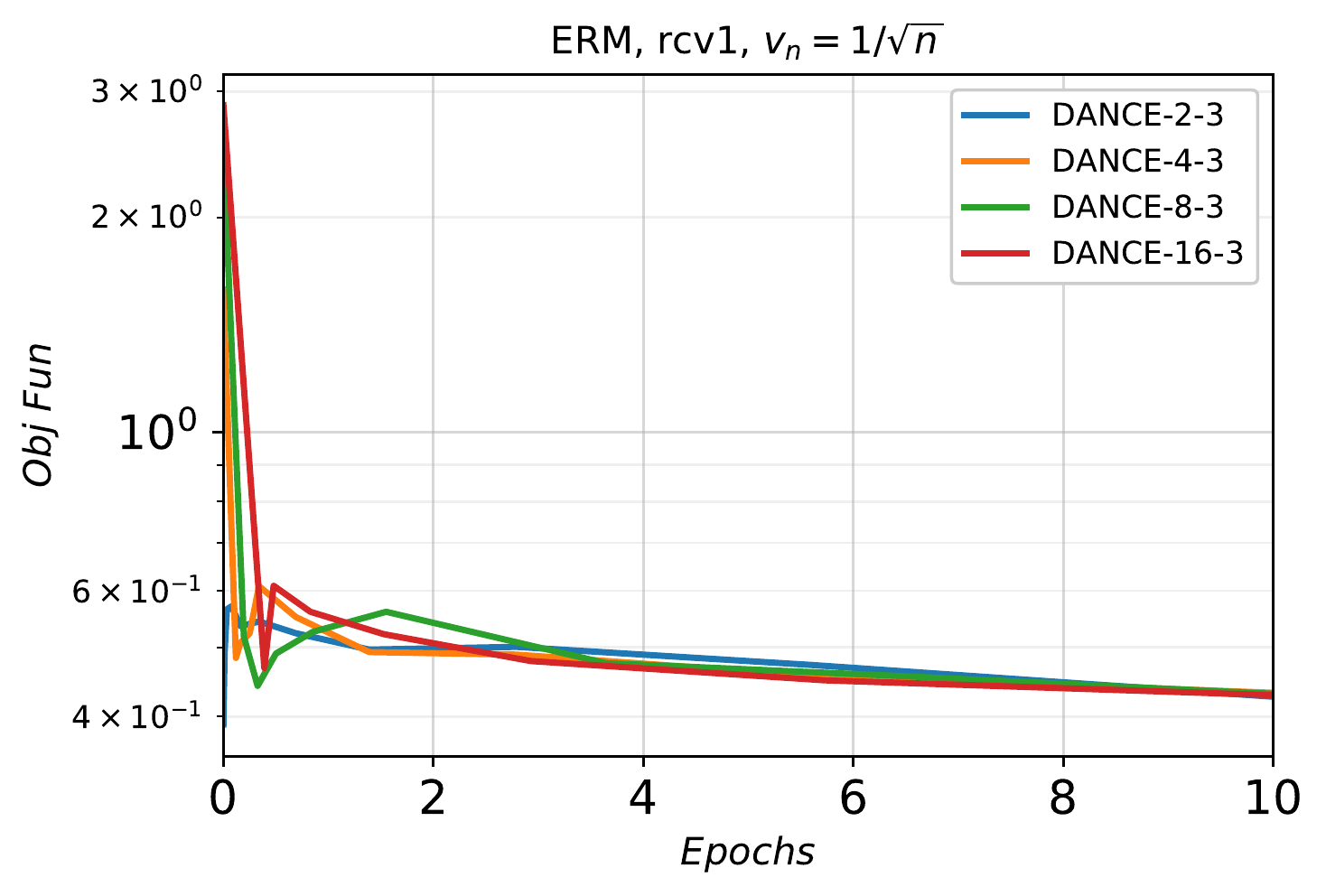}
	\includegraphics[width=0.30\textwidth]{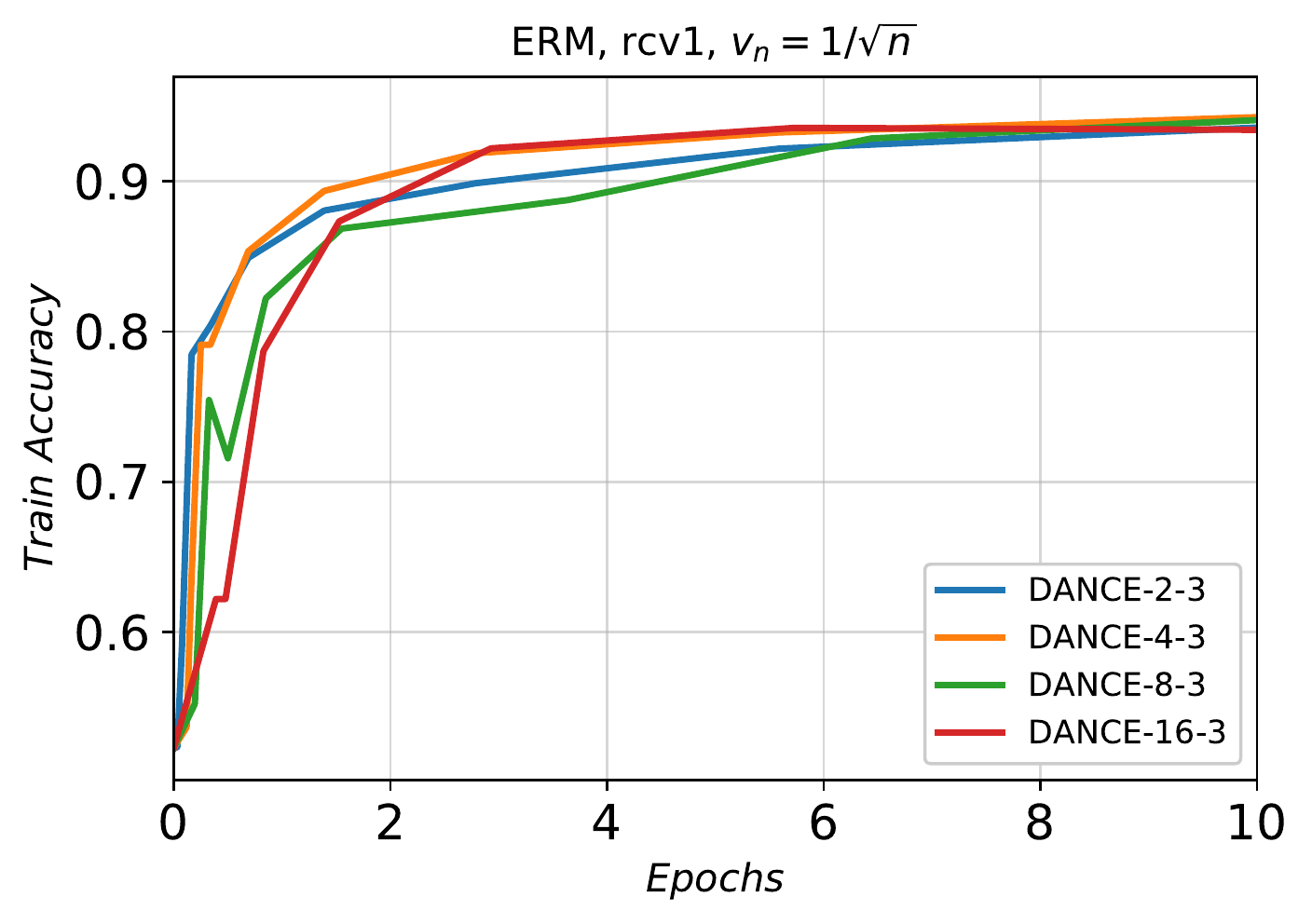}
	\includegraphics[width=0.30\textwidth]{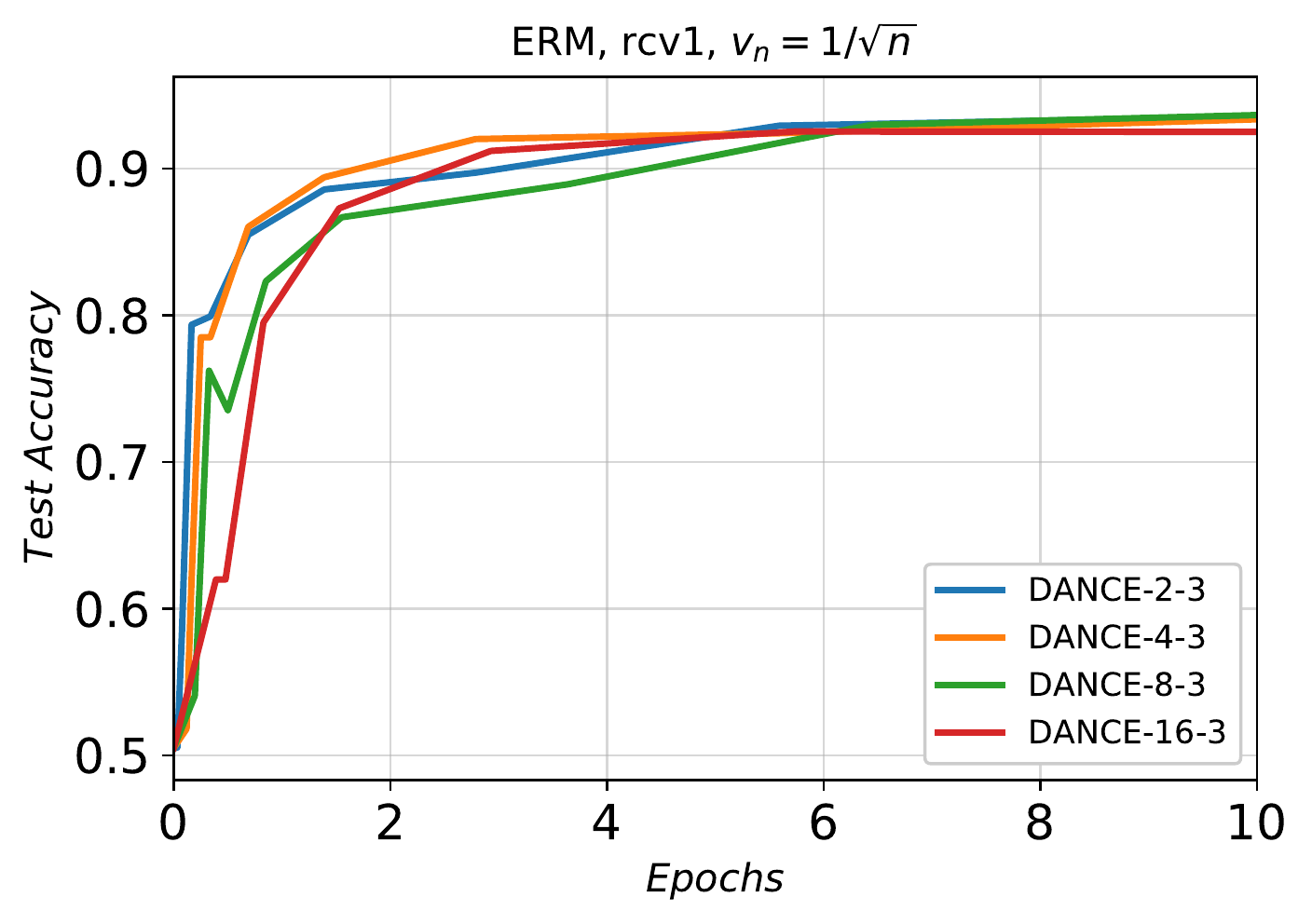}
	
	\includegraphics[width=0.30\textwidth]{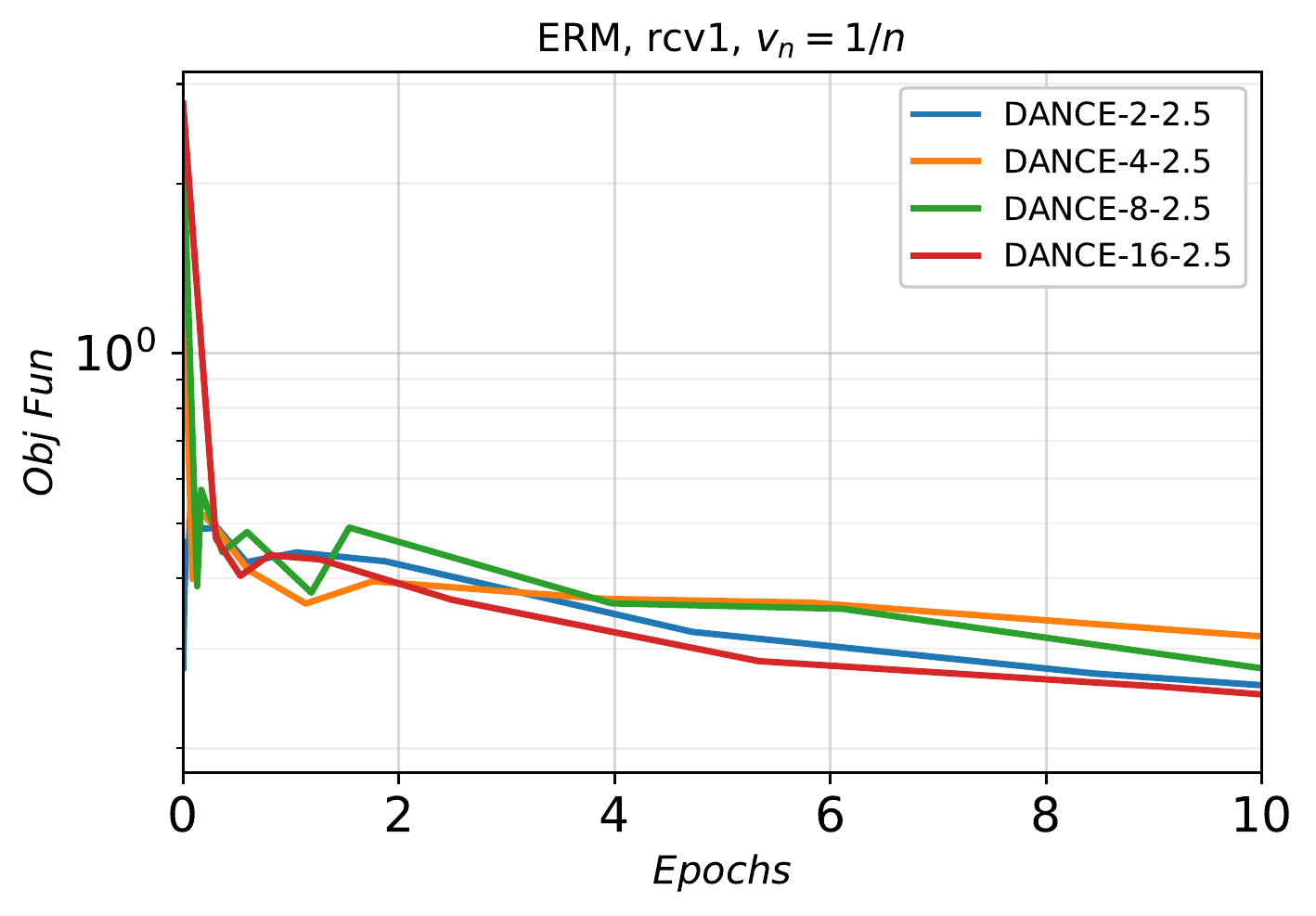}
	\includegraphics[width=0.30\textwidth]{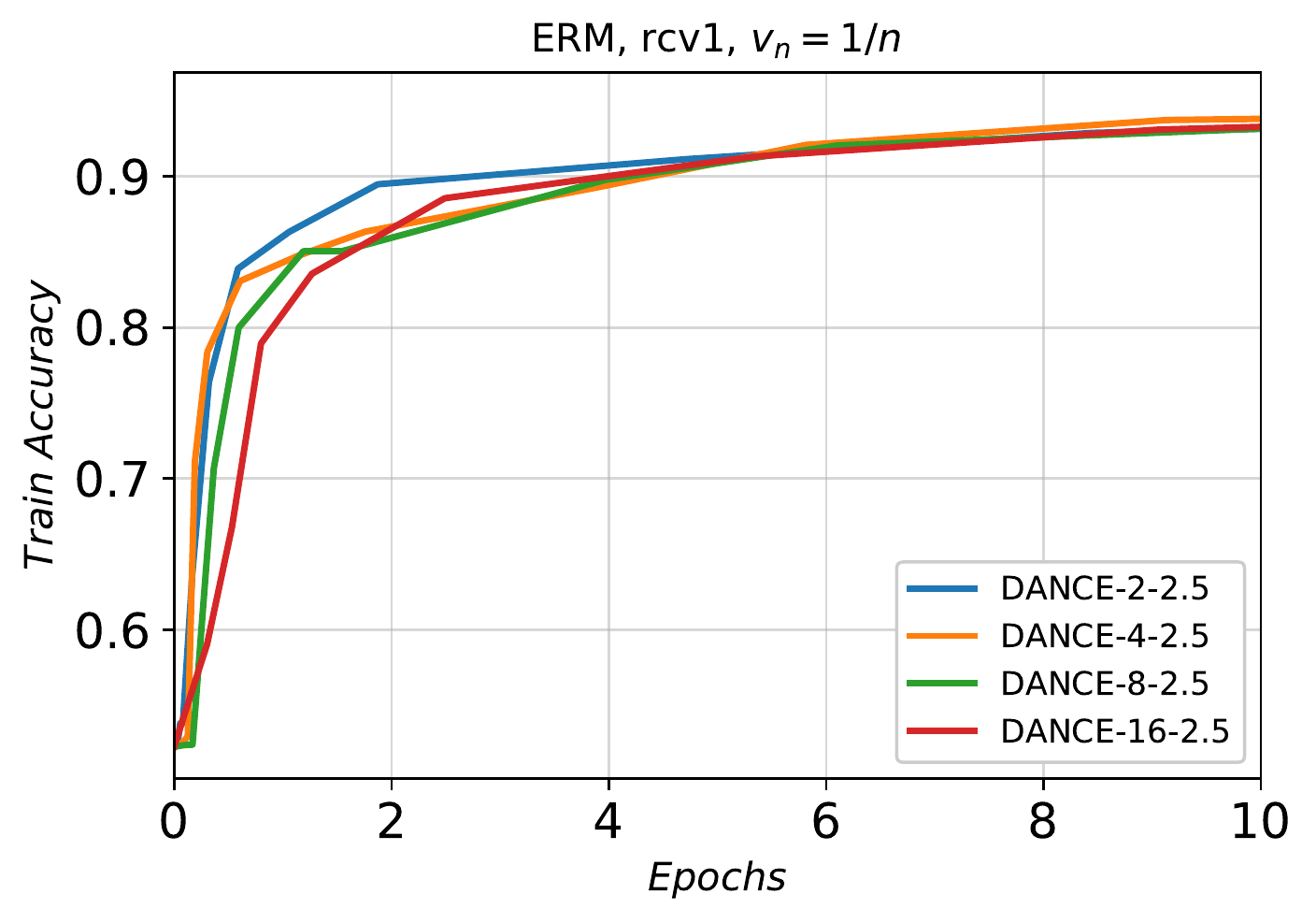}
	\includegraphics[width=0.30\textwidth]{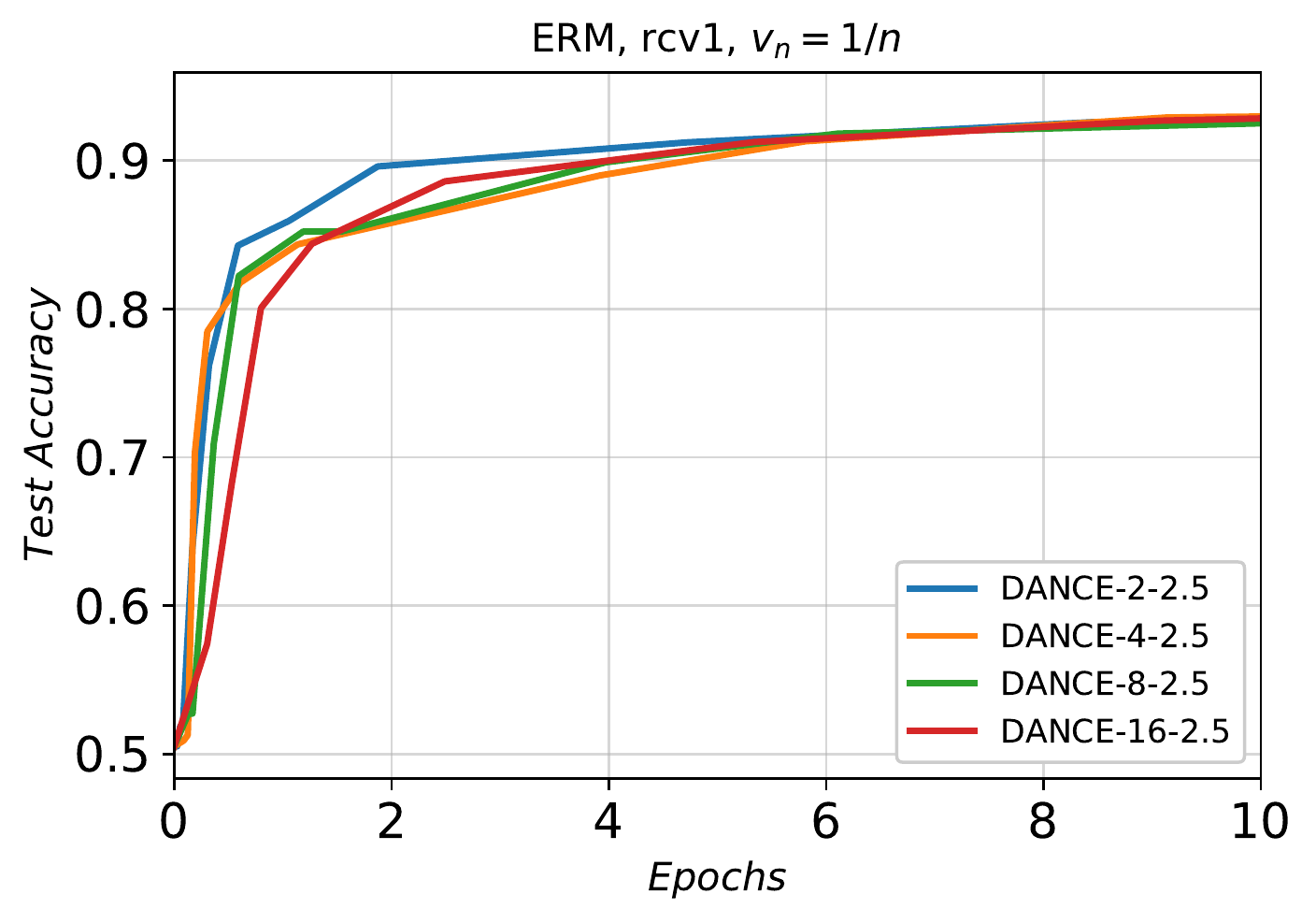}
	\caption{DANCE performance with respect to  fixed $\alpha$ and different values of initial samples.  The first row shows the results of DANCE for ``rcv1" dataset when $V_n = \dfrac{1}{\sqrt{n}}$. The second row shows the results of DANCE for ``rcv1" dataset when when $V_n = \dfrac{1}{n}$.  }
		\label{fig:11}
\end{figure*}

\begin{figure*}[h]
	\centering
	\includegraphics[width=0.30\textwidth]{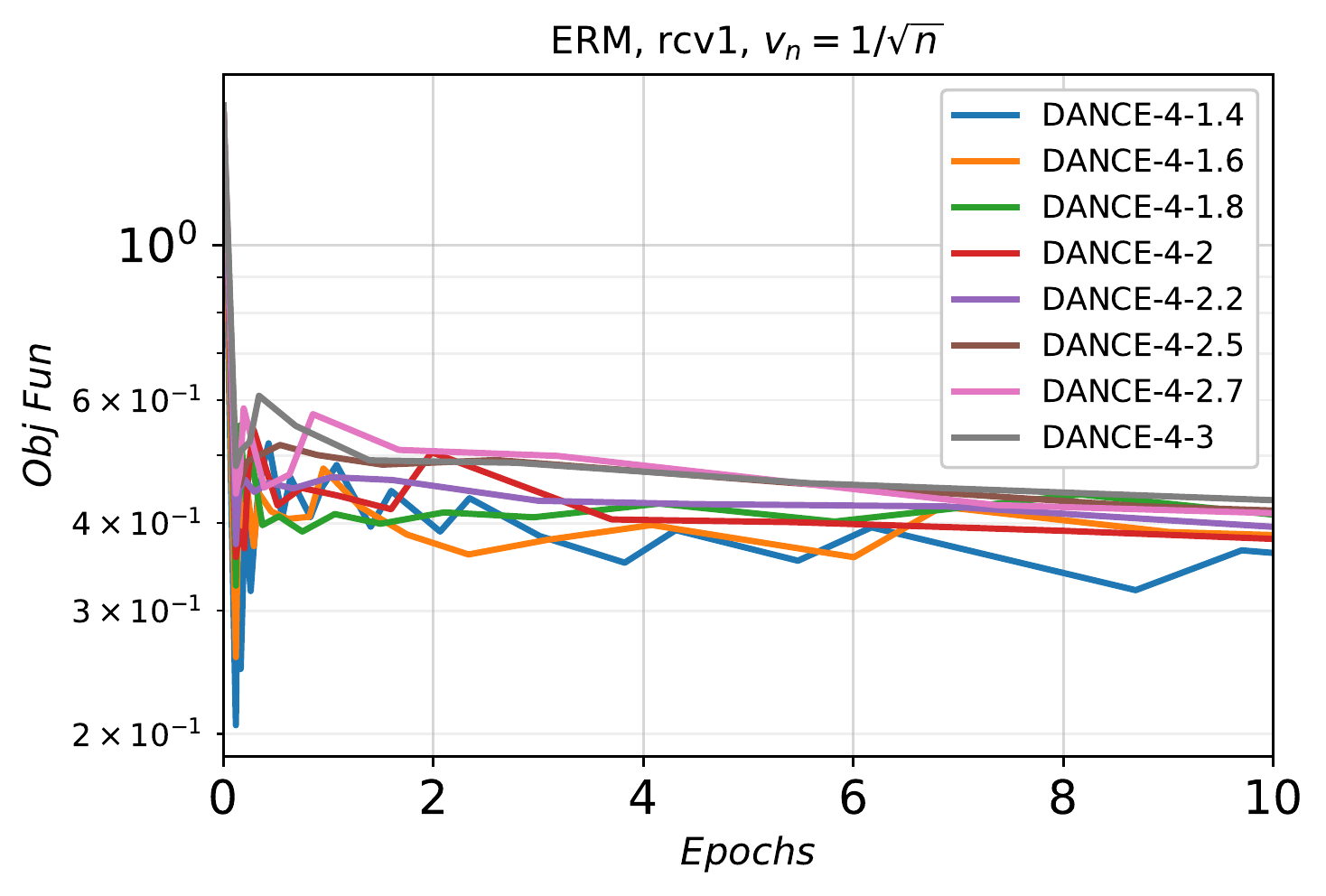}
	\includegraphics[width=0.30\textwidth]{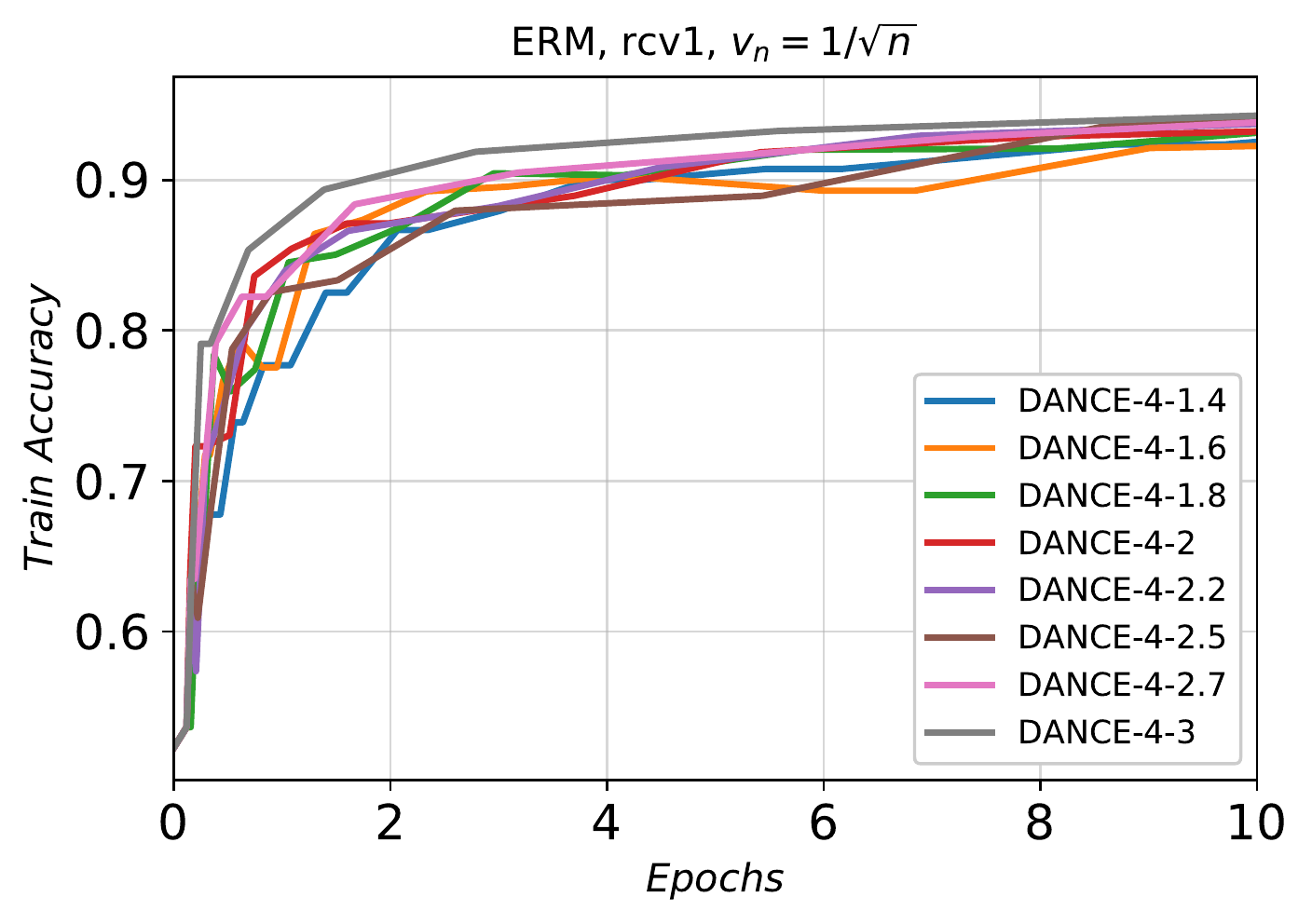}
	\includegraphics[width=0.30\textwidth]{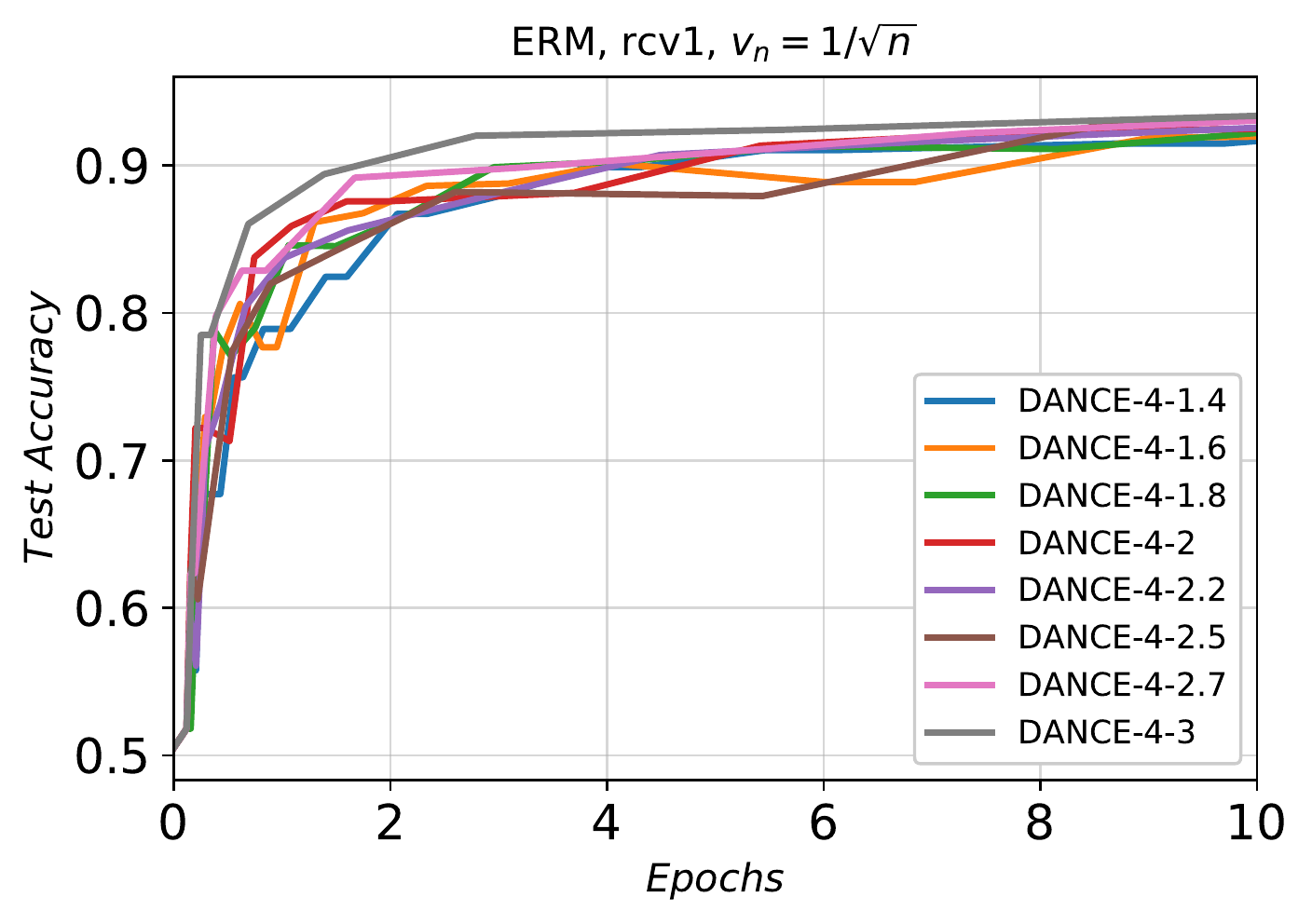}
	
	\includegraphics[width=0.30\textwidth]{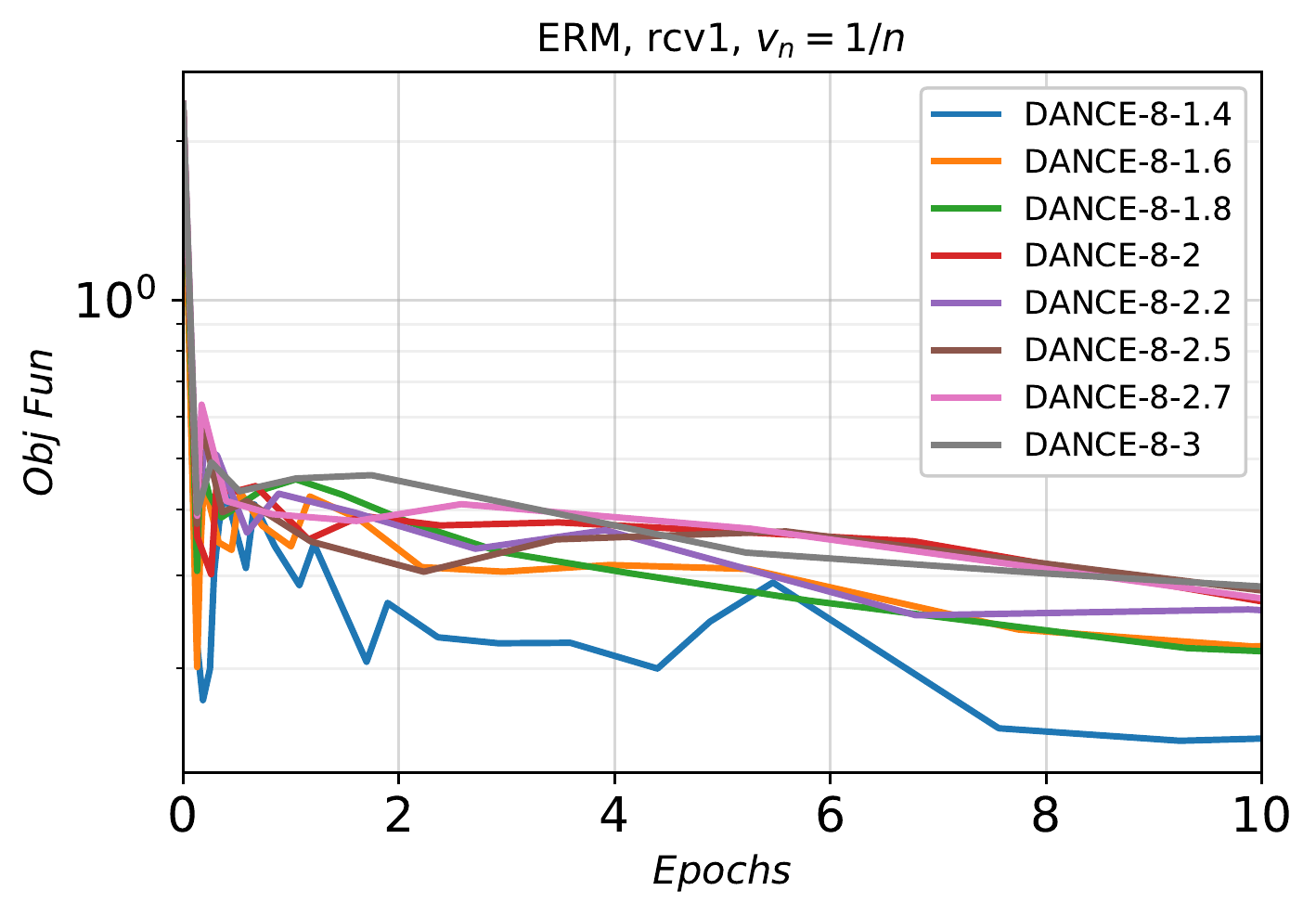}
	\includegraphics[width=0.30\textwidth]{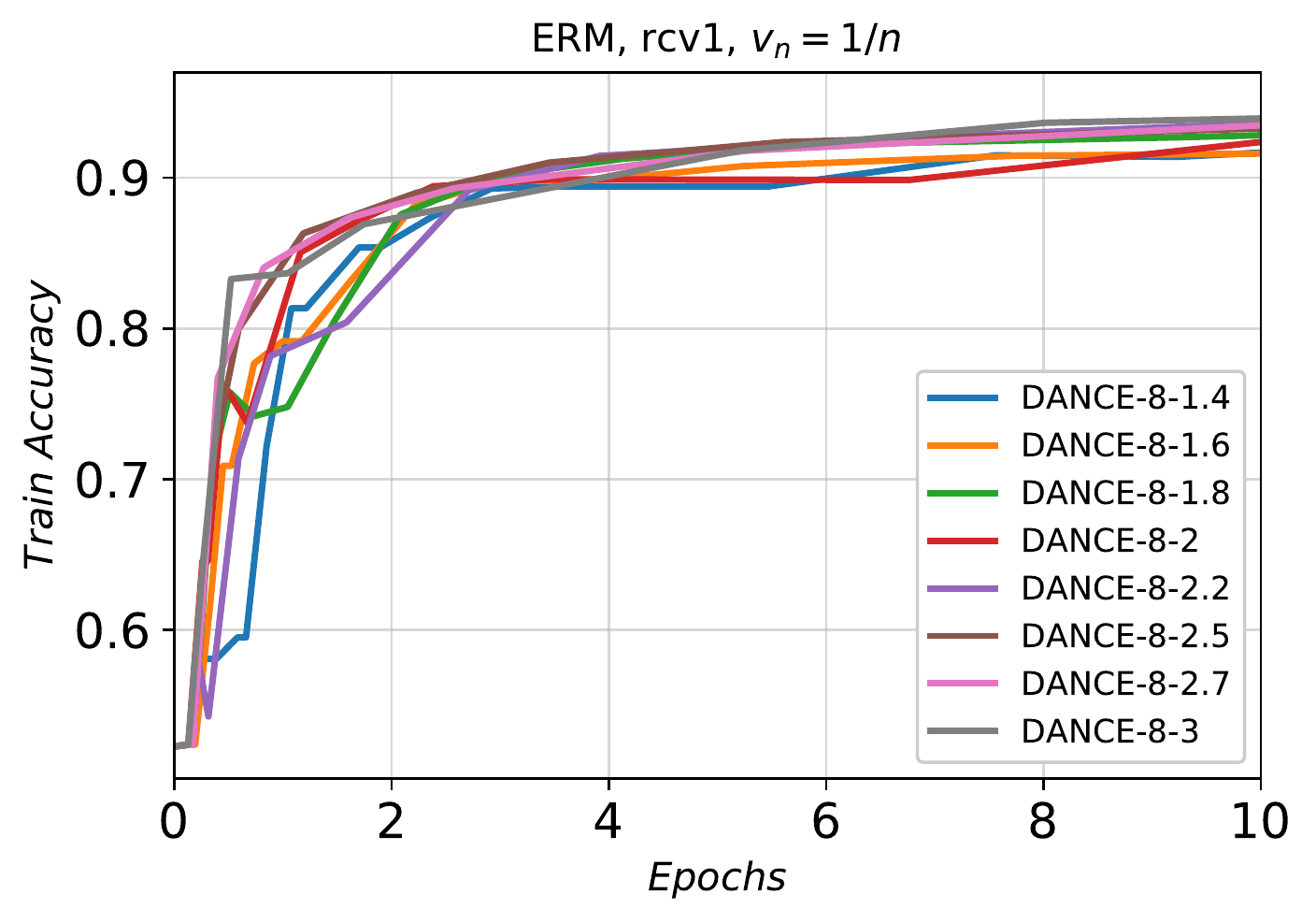}
	\includegraphics[width=0.30\textwidth]{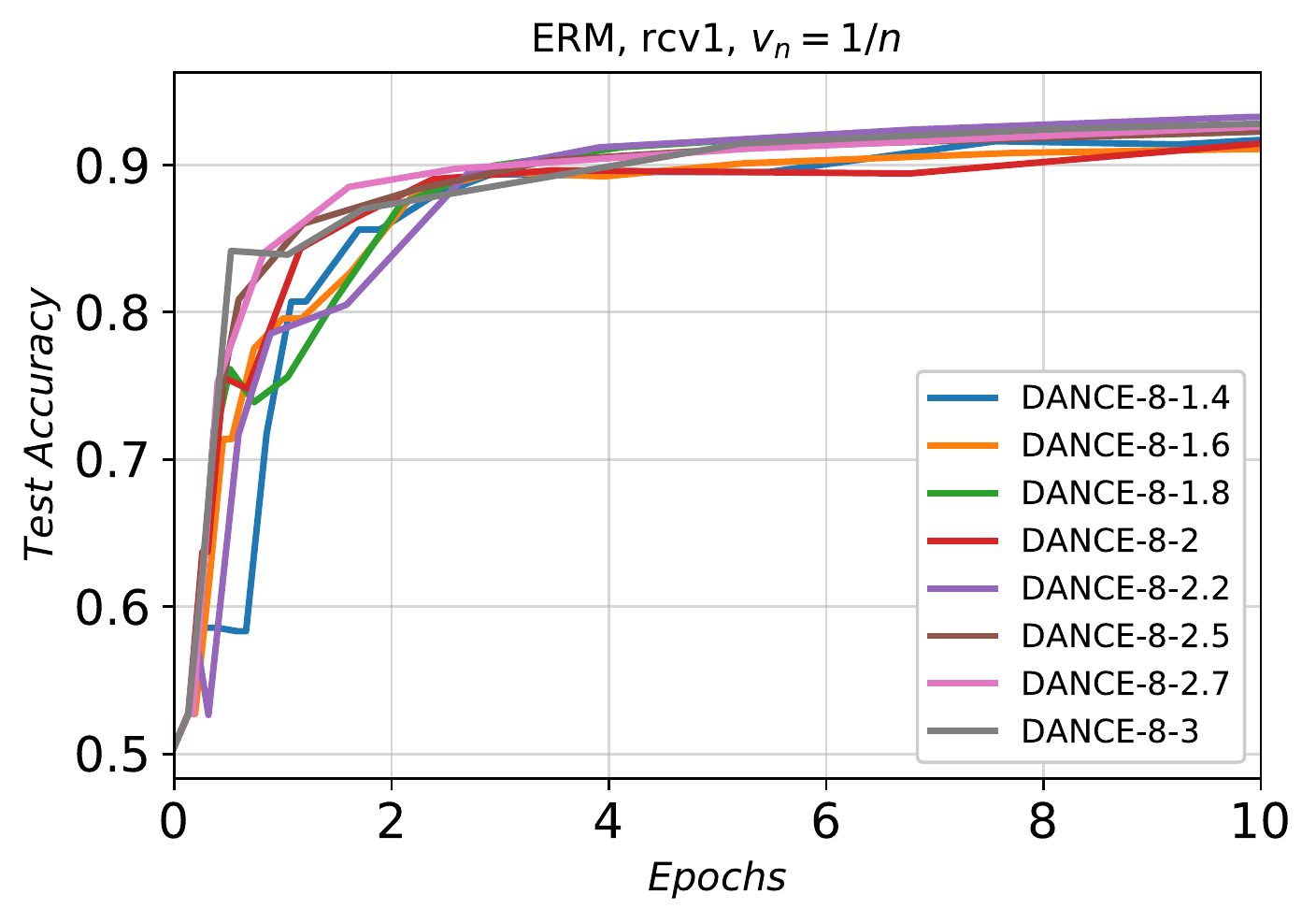}
	\caption{DANCE performance with respect to different values of $\alpha$ and fixed number of initial sample.  The first row shows the results of DANCE for ``rcv1" dataset when $V_n = \dfrac{1}{\sqrt{n}}$. The second row shows the results of DANCE for ``rcv1" dataset when when $V_n = \dfrac{1}{n}$. }
	\label{fig:12}
\end{figure*}

\begin{figure*}[h]
	\centering
	\includegraphics[width=0.30\textwidth]{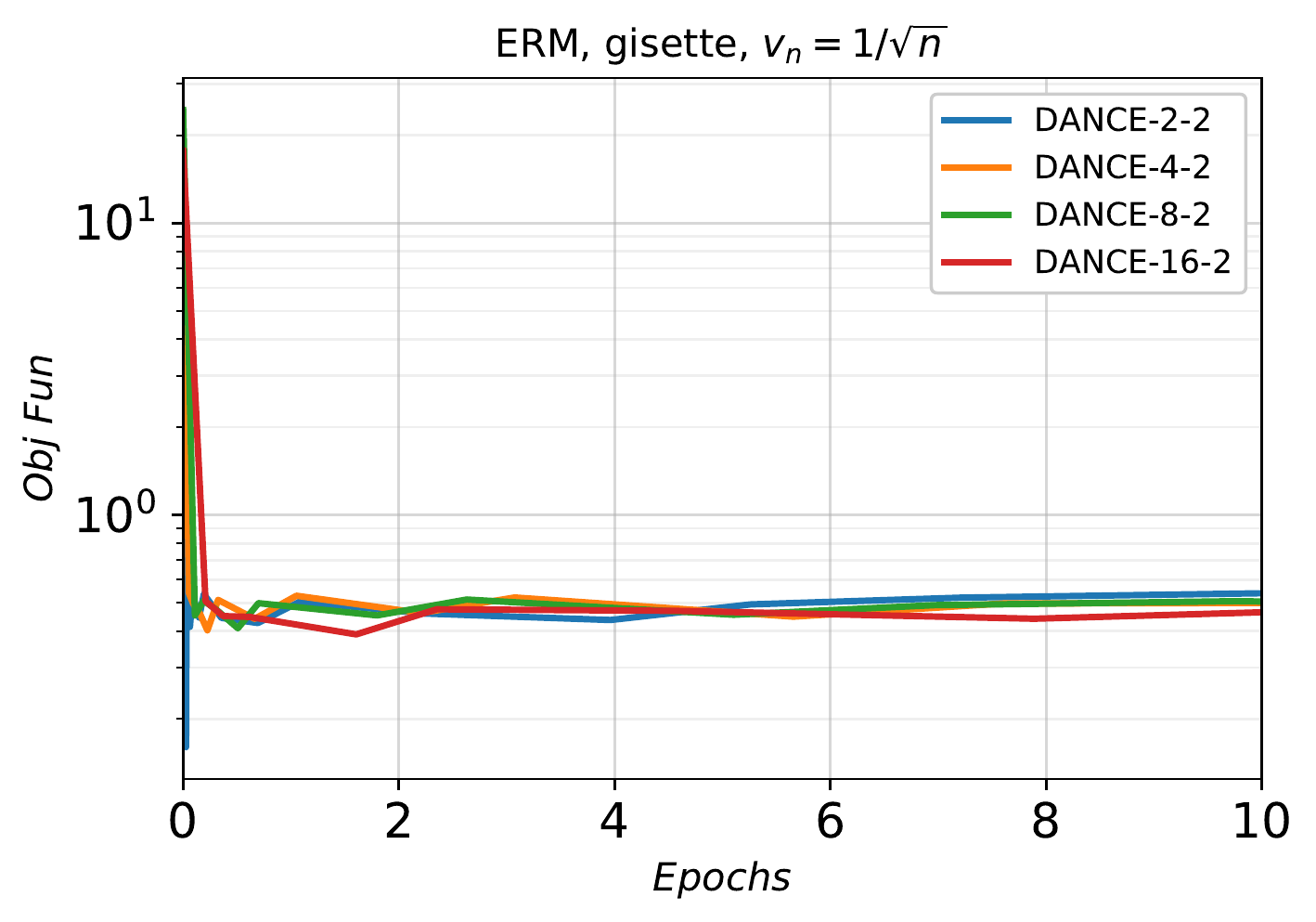}
	\includegraphics[width=0.30\textwidth]{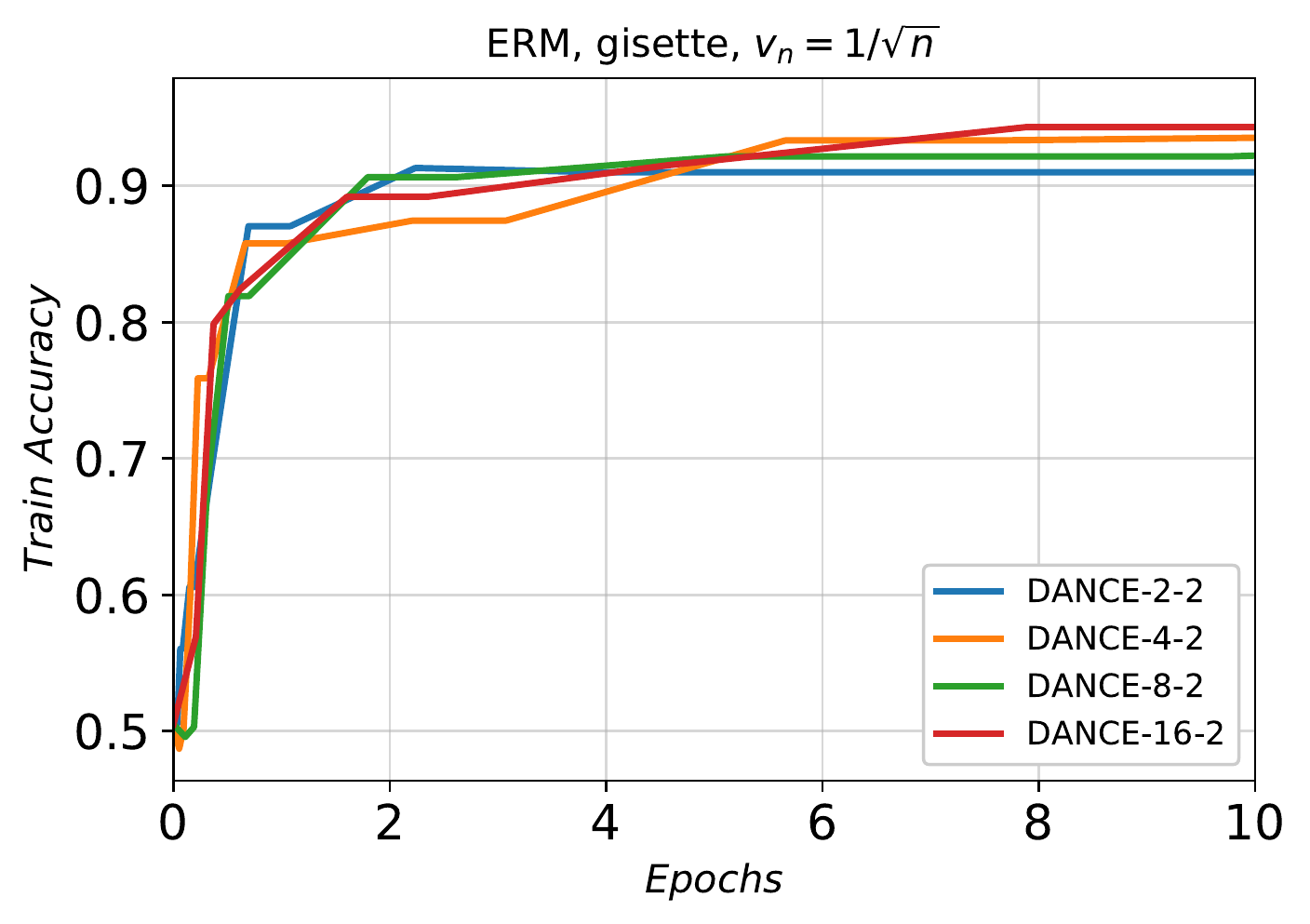}
	\includegraphics[width=0.30\textwidth]{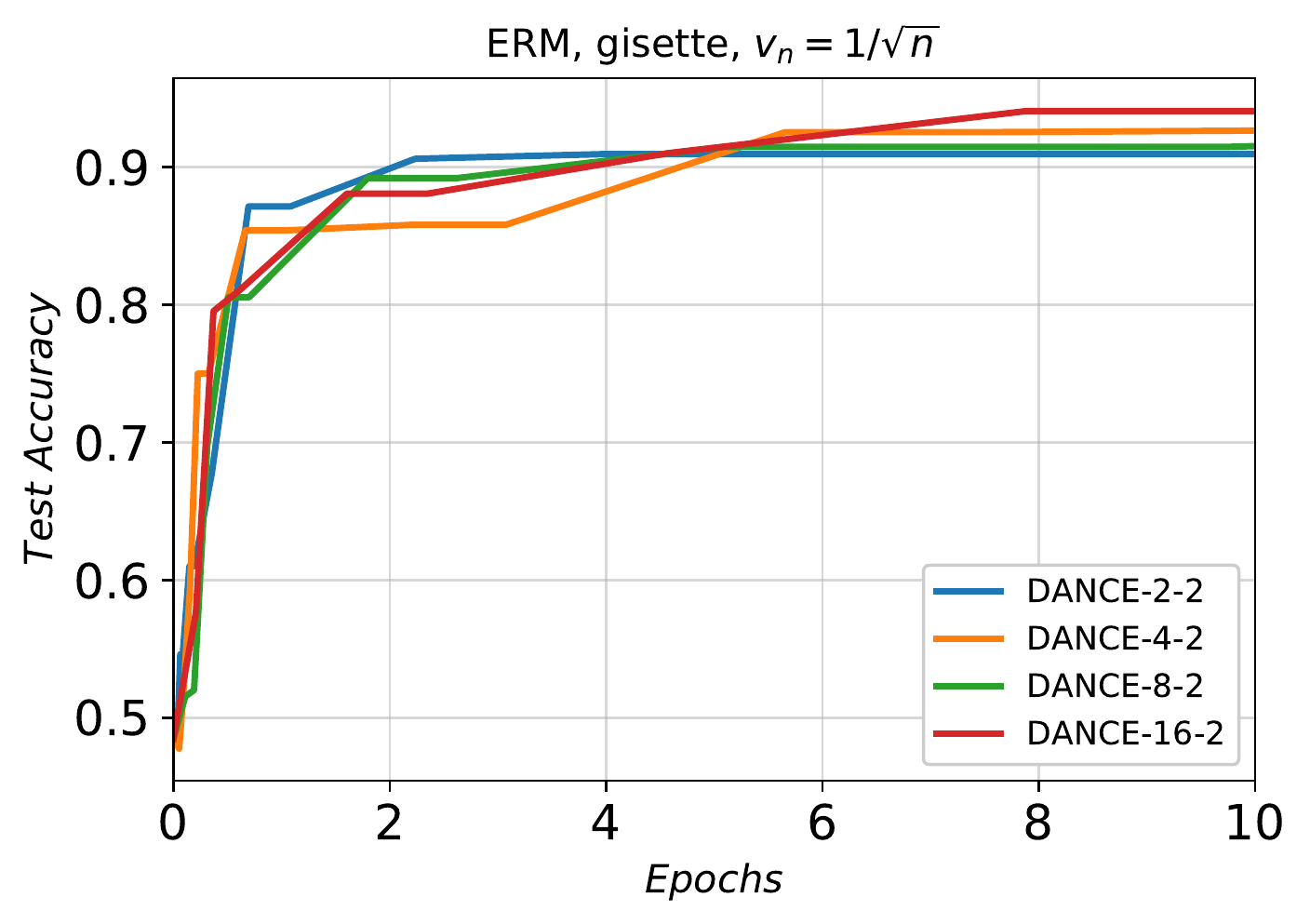}
	
	\includegraphics[width=0.30\textwidth]{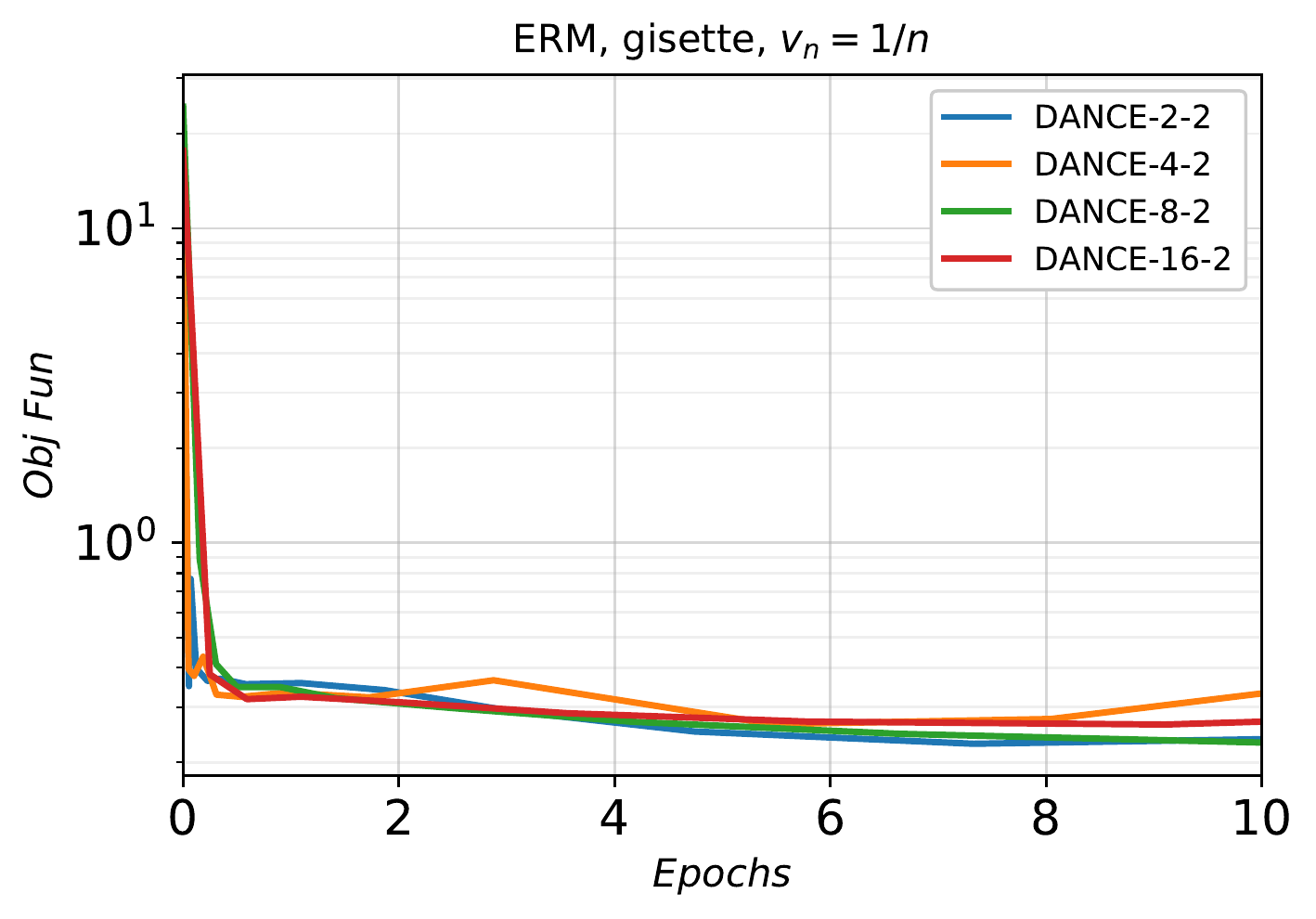}
	\includegraphics[width=0.30\textwidth]{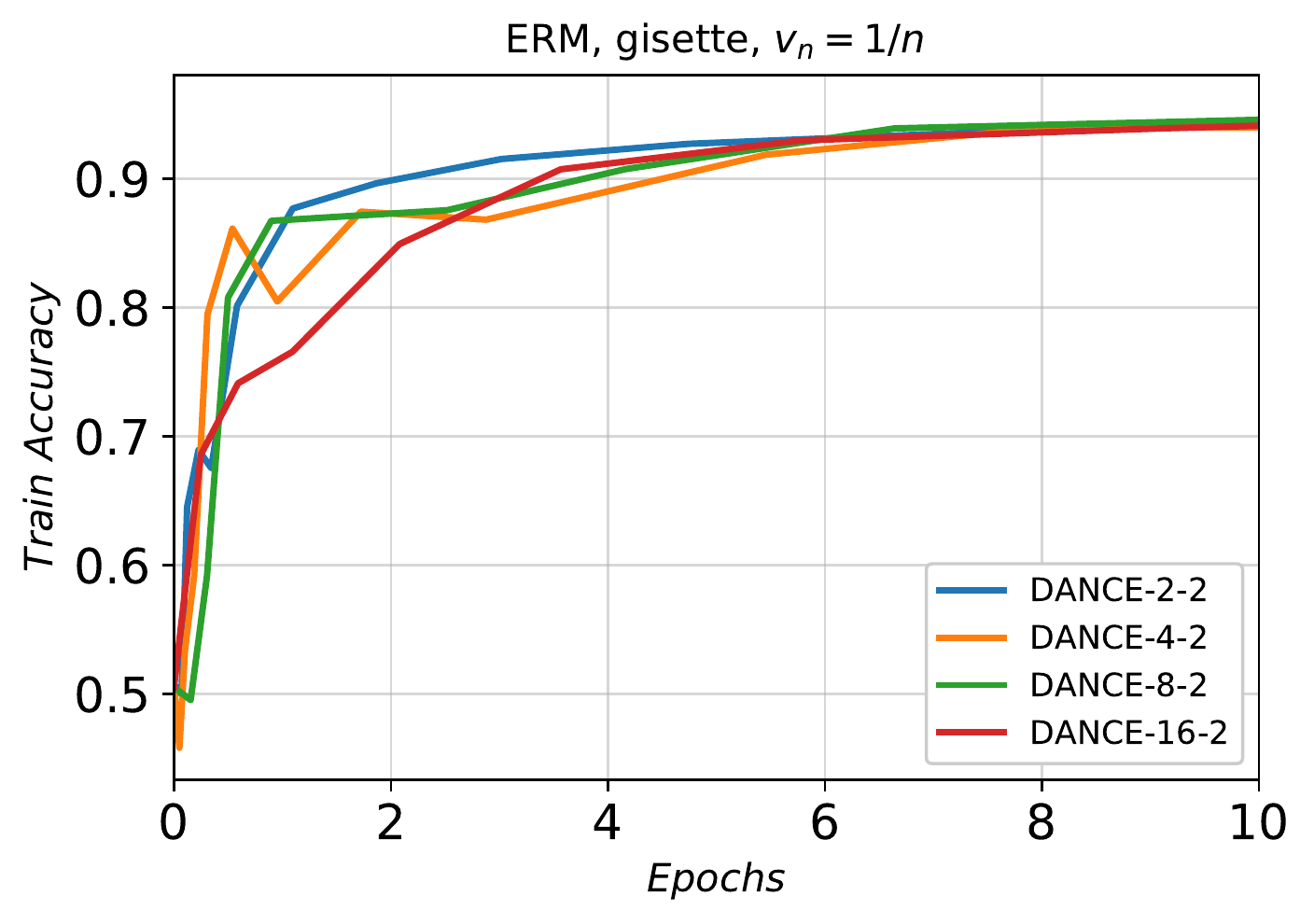}
	\includegraphics[width=0.30\textwidth]{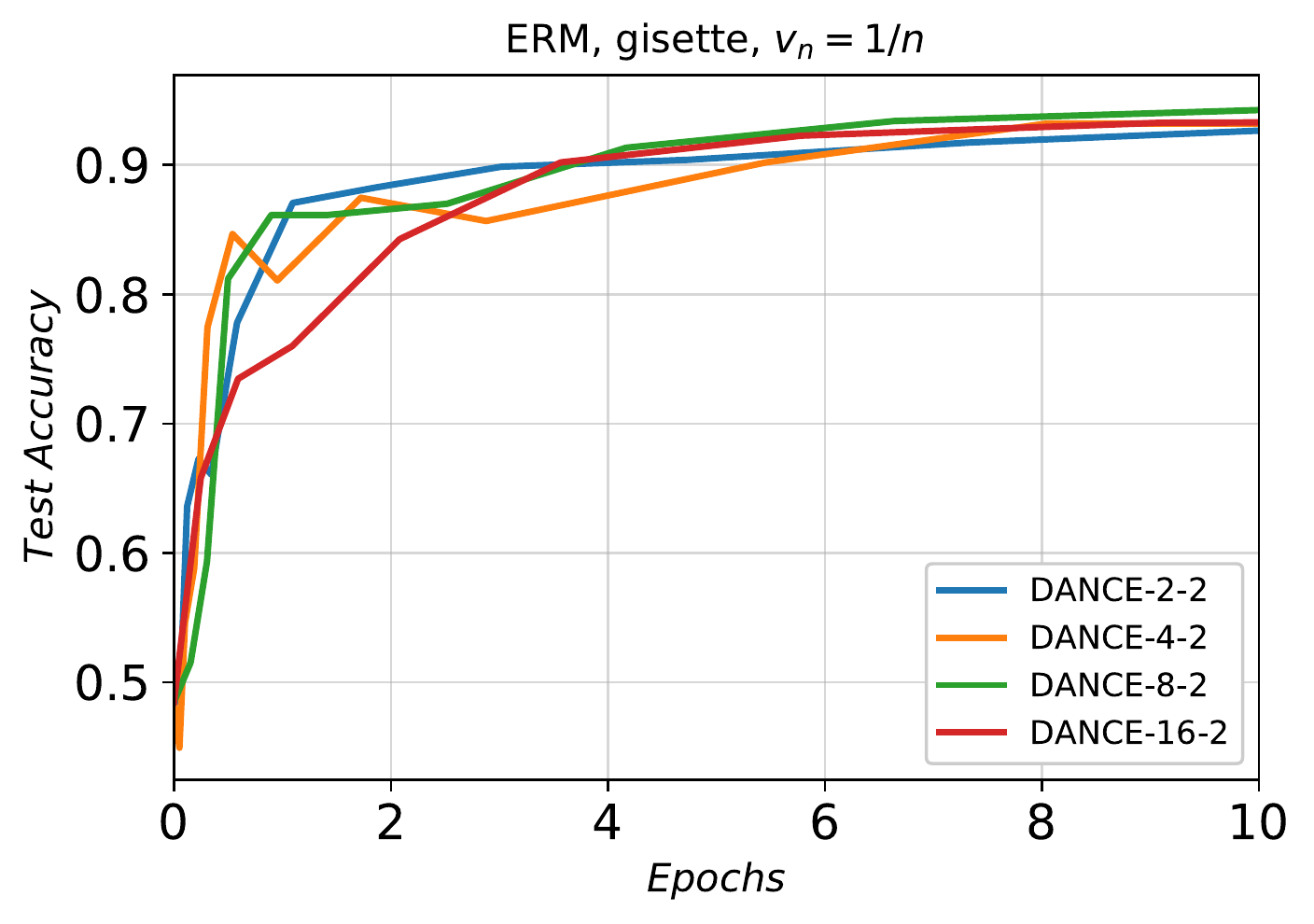}
	\caption{DANCE performance with respect to  fixed $\alpha$ and different values of initial samples.  The first row shows the results of DANCE for ``gisette" dataset when $V_n = \dfrac{1}{\sqrt{n}}$. The second row shows the results of DANCE for ``gisette" dataset when when $V_n = \dfrac{1}{n}$. }
	\label{fig:13}
\end{figure*}

\begin{figure*}[h]
	\centering
	\includegraphics[width=0.30\textwidth]{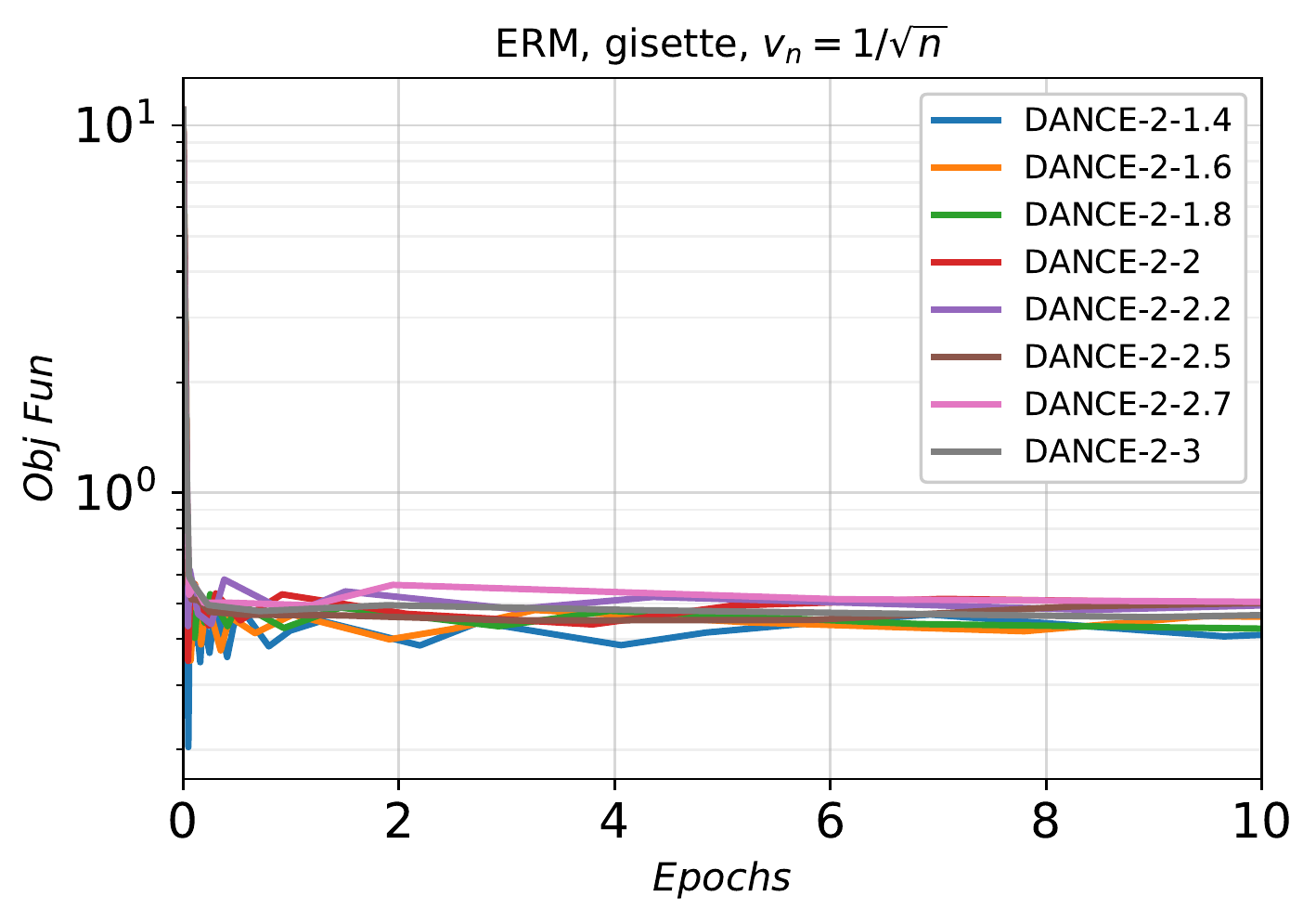}
	\includegraphics[width=0.30\textwidth]{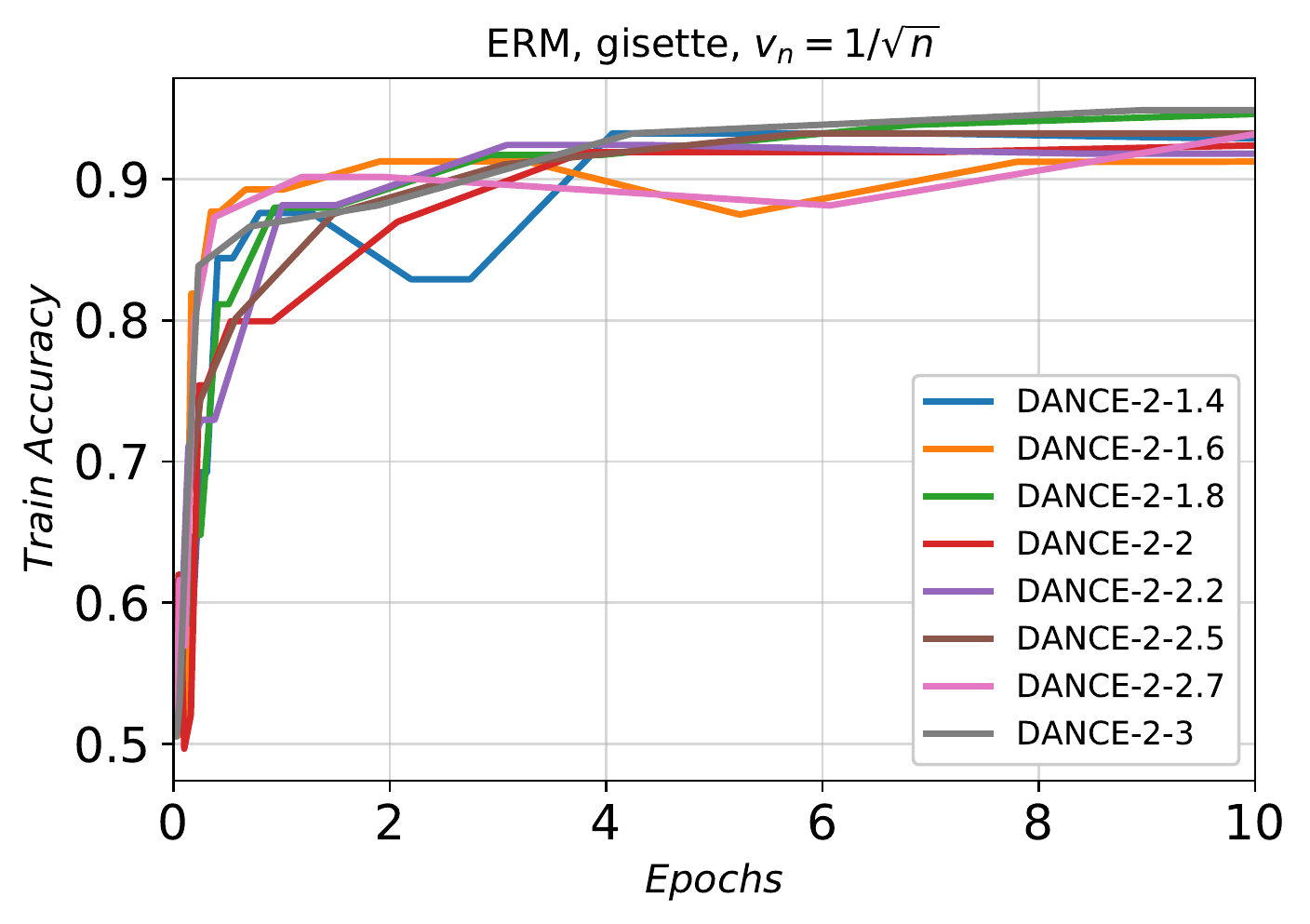}
	\includegraphics[width=0.30\textwidth]{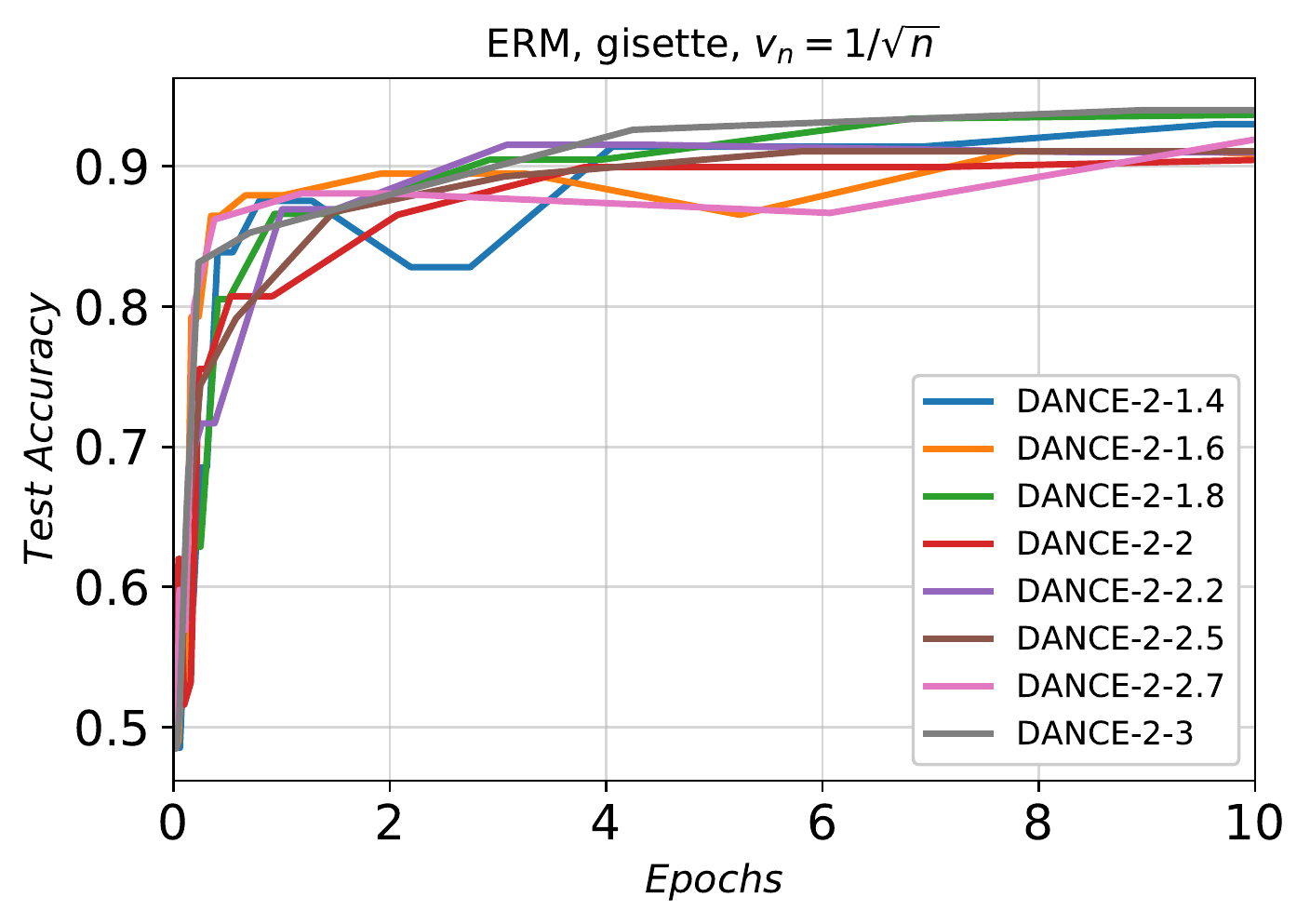}
	
	\includegraphics[width=0.30\textwidth]{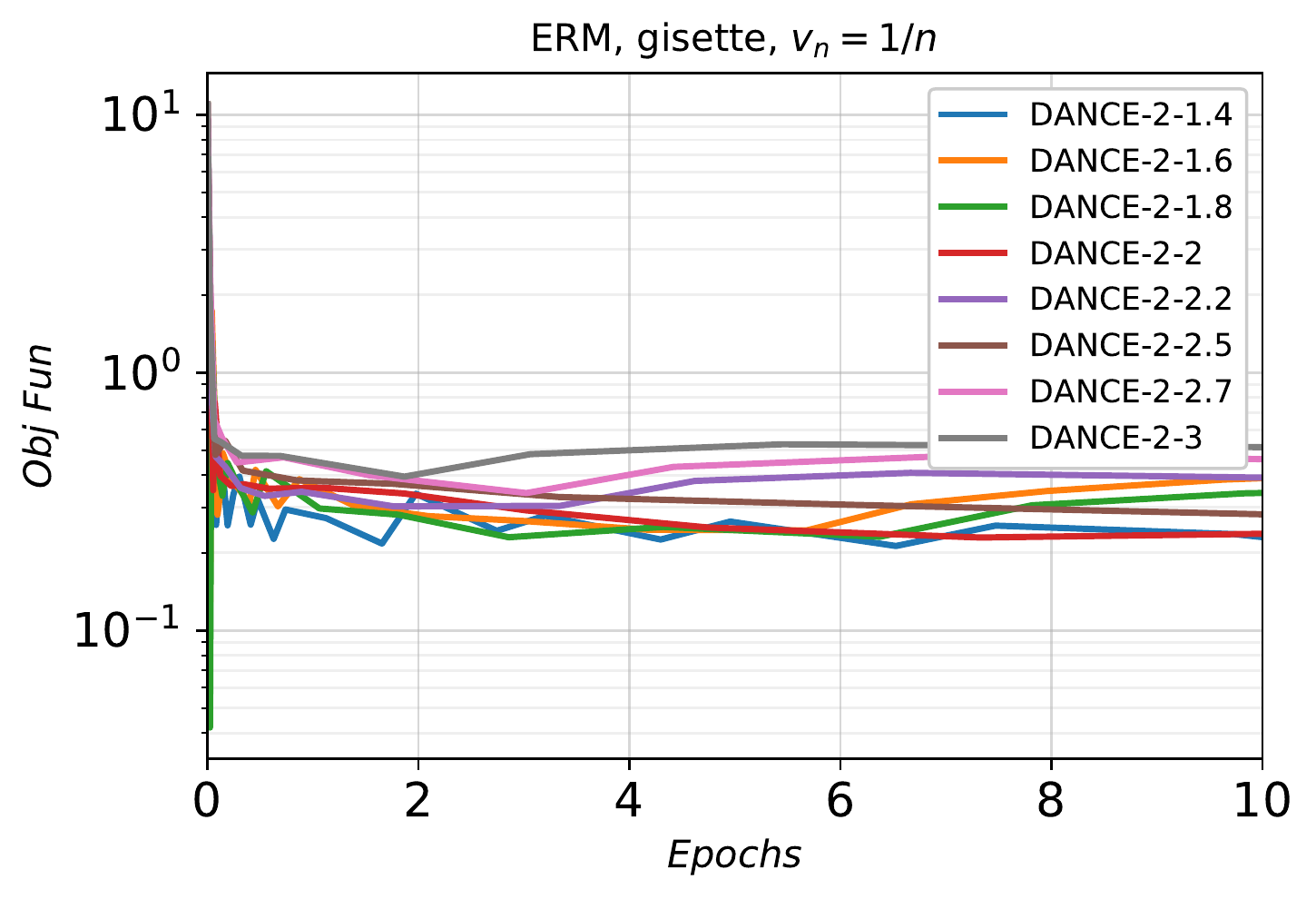}
	\includegraphics[width=0.30\textwidth]{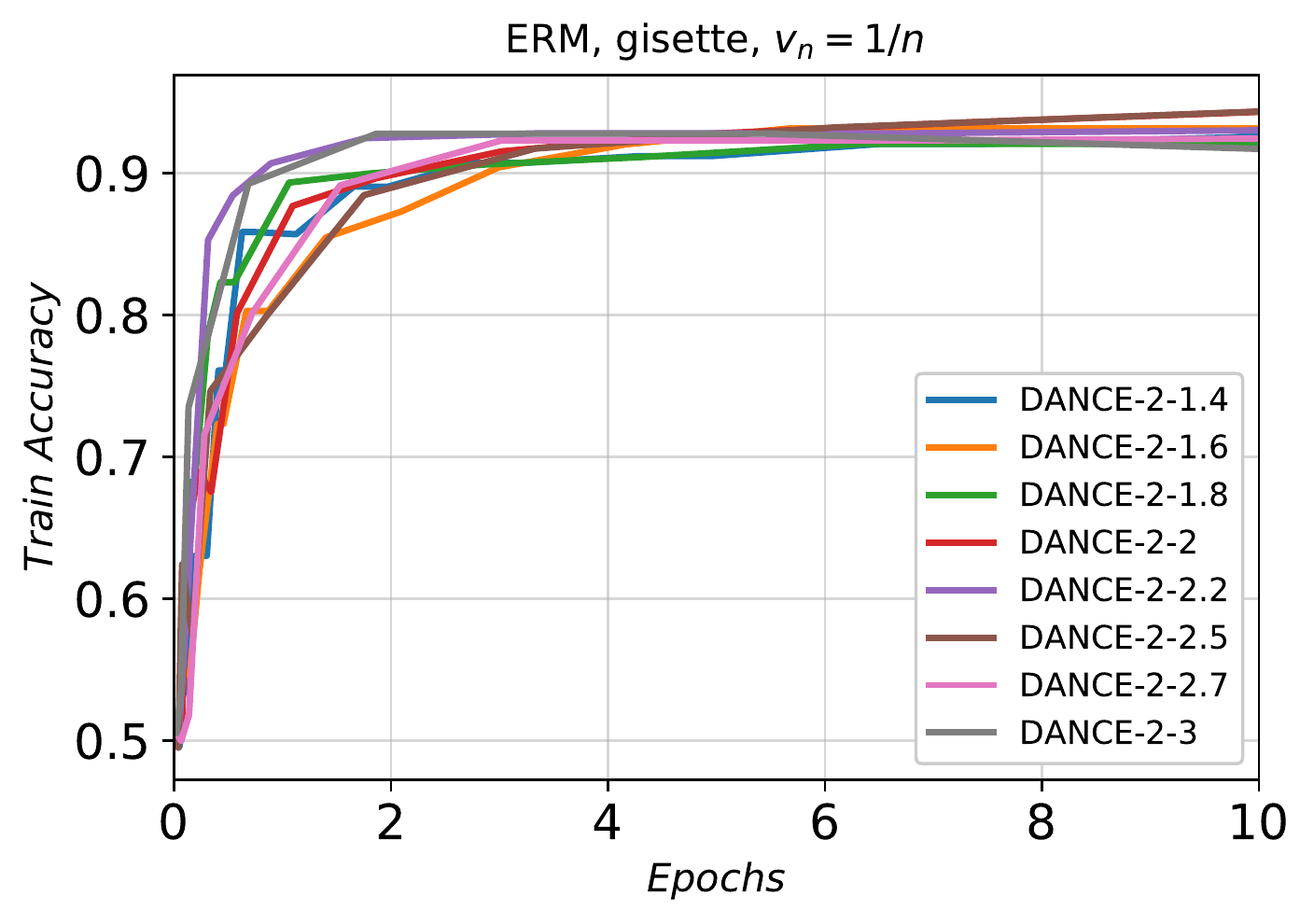}
	\includegraphics[width=0.30\textwidth]{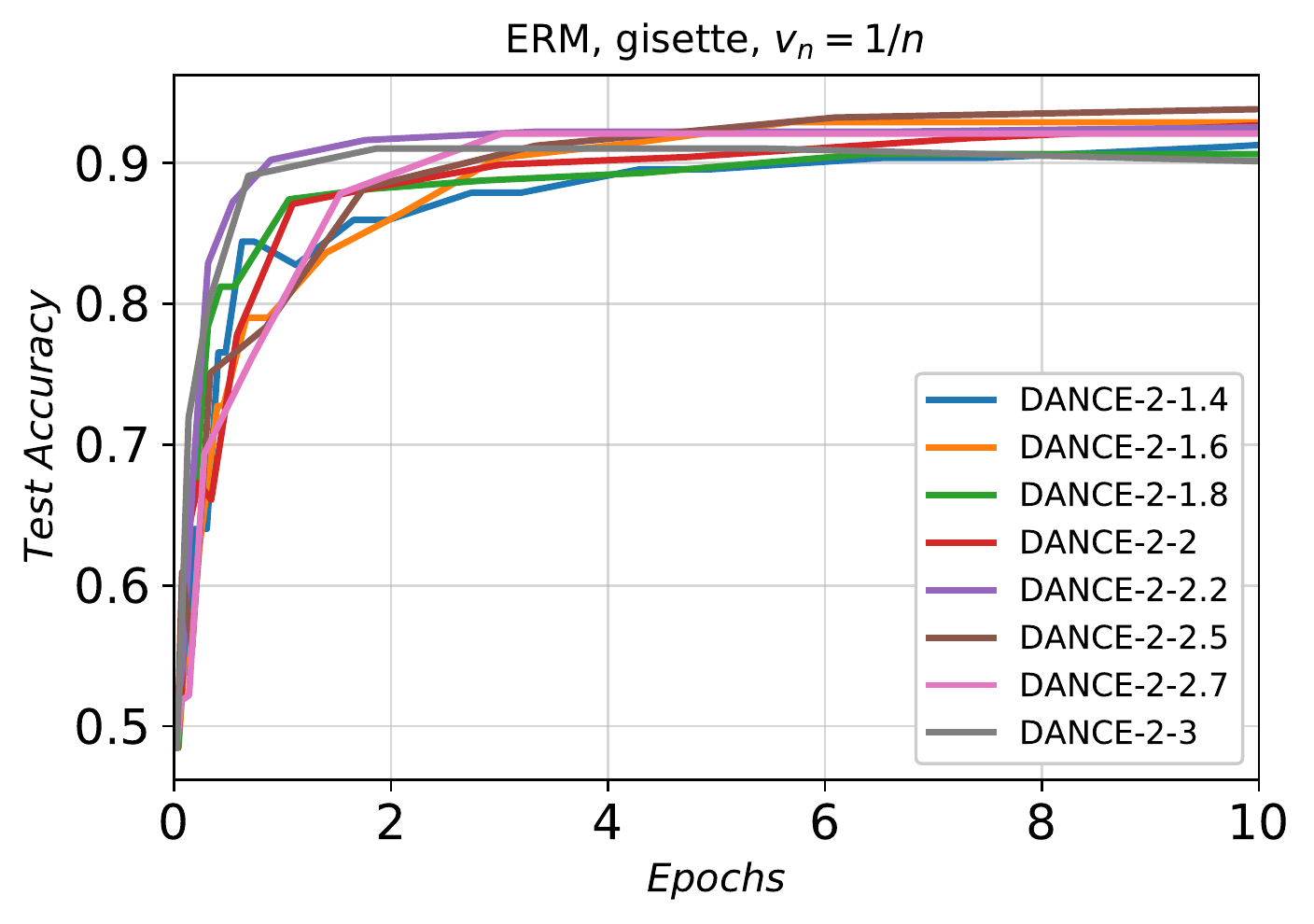}
	\caption{DANCE performance with respect to different values of $\alpha$ and fixed number of initial sample.  The first row shows the results of DANCE for ``gisette" dataset when $V_n = \dfrac{1}{\sqrt{n}}$. The second row shows the results of DANCE for ``gisette" dataset when when $V_n = \dfrac{1}{n}$. }
	\label{fig:14}
\end{figure*}

In this section, we consider different possible values of hyper-parameters for DANCE, and as it is clear from Figures \ref{fig:11}, \ref{fig:12}, \ref{fig:13} and \ref{fig:14}, DANCE behaves in a robust way when changes in the hyper-parameters happen. In other words, DANCE is not that much sensitive to the choice of hyper-parameters.

\begin{figure*}[h]
	\centering
	\includegraphics[width=0.30\textwidth]{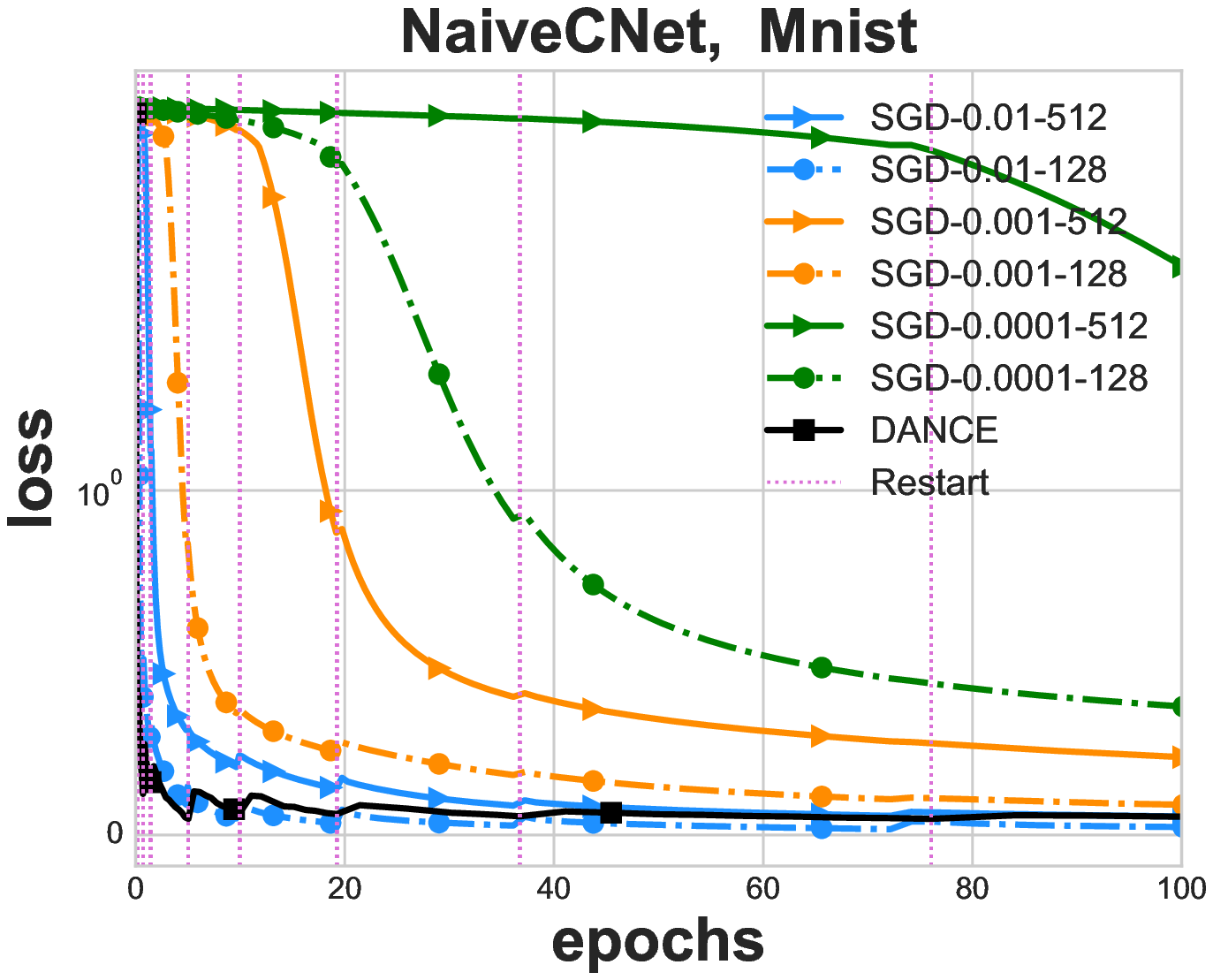}
	\includegraphics[width=0.30\textwidth]{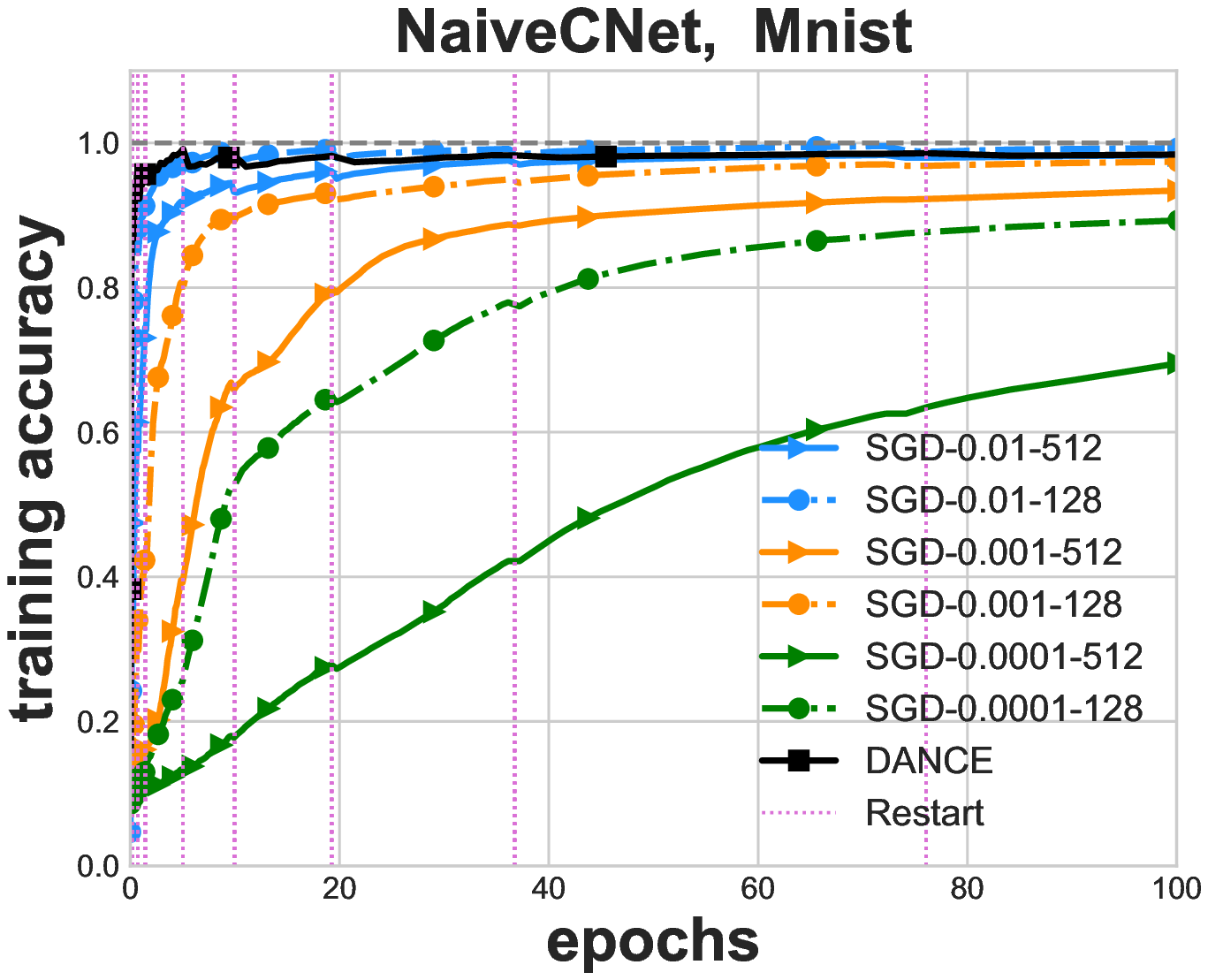}
	\includegraphics[width=0.30\textwidth]{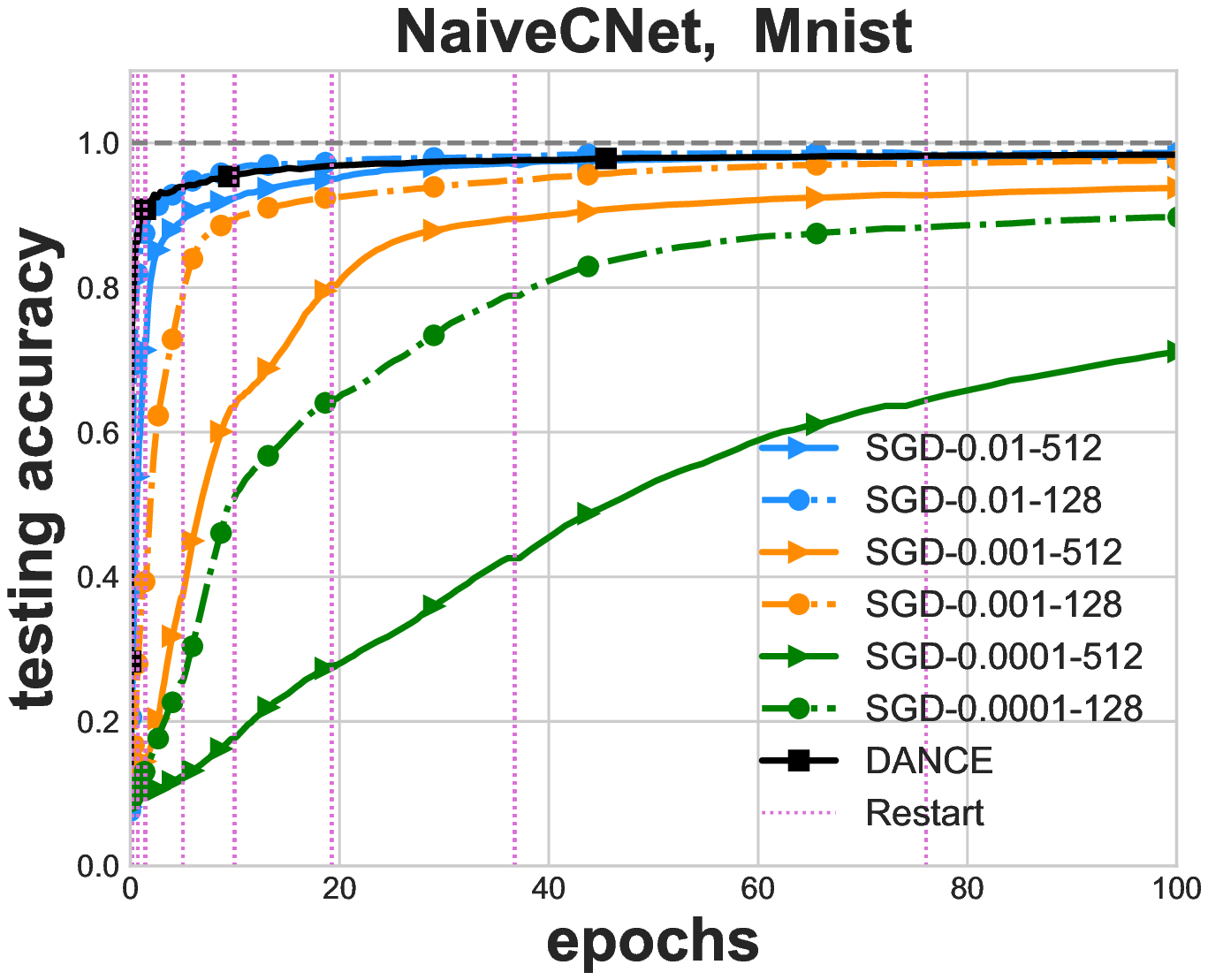}
	
	\includegraphics[width=0.30\textwidth]{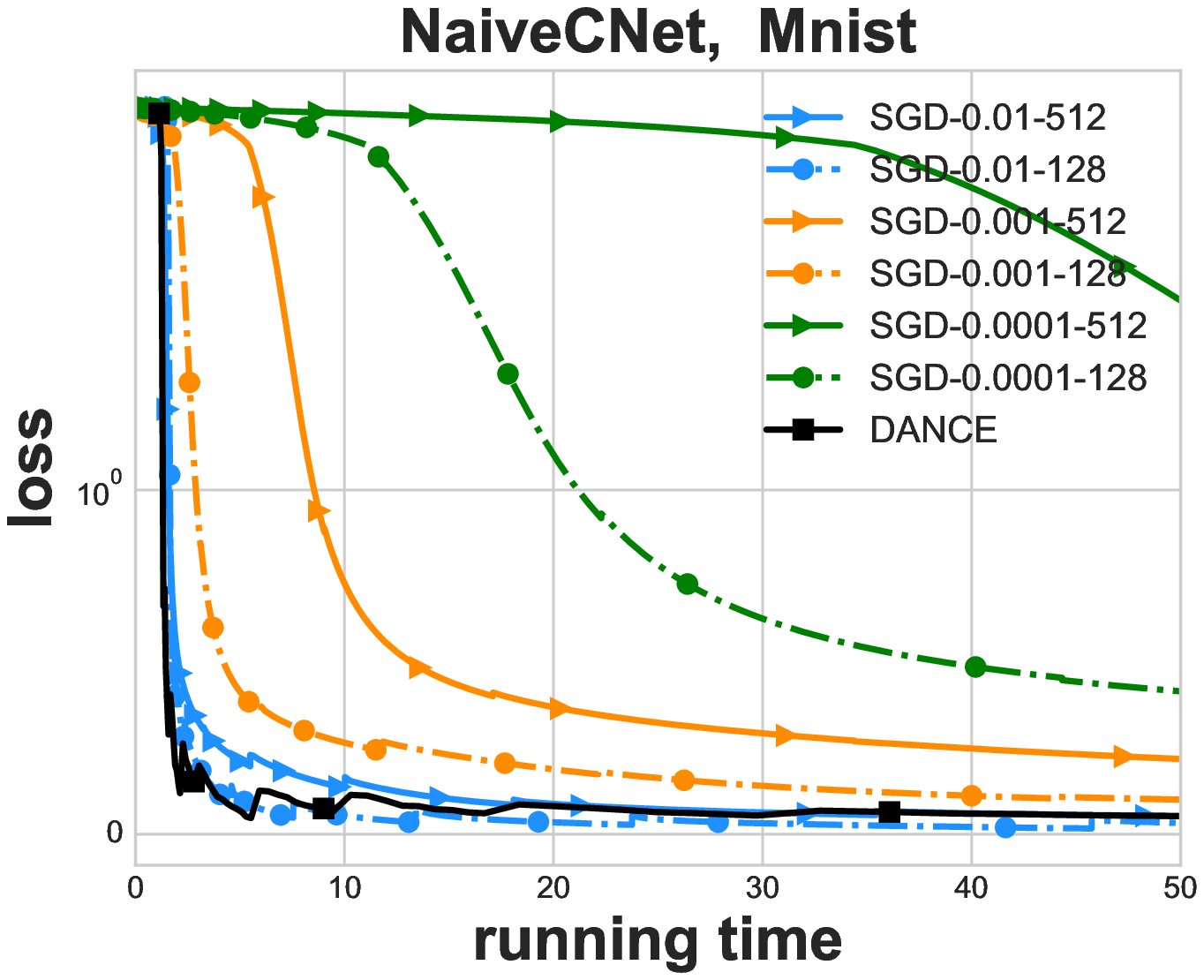}
	\includegraphics[width=0.30\textwidth]{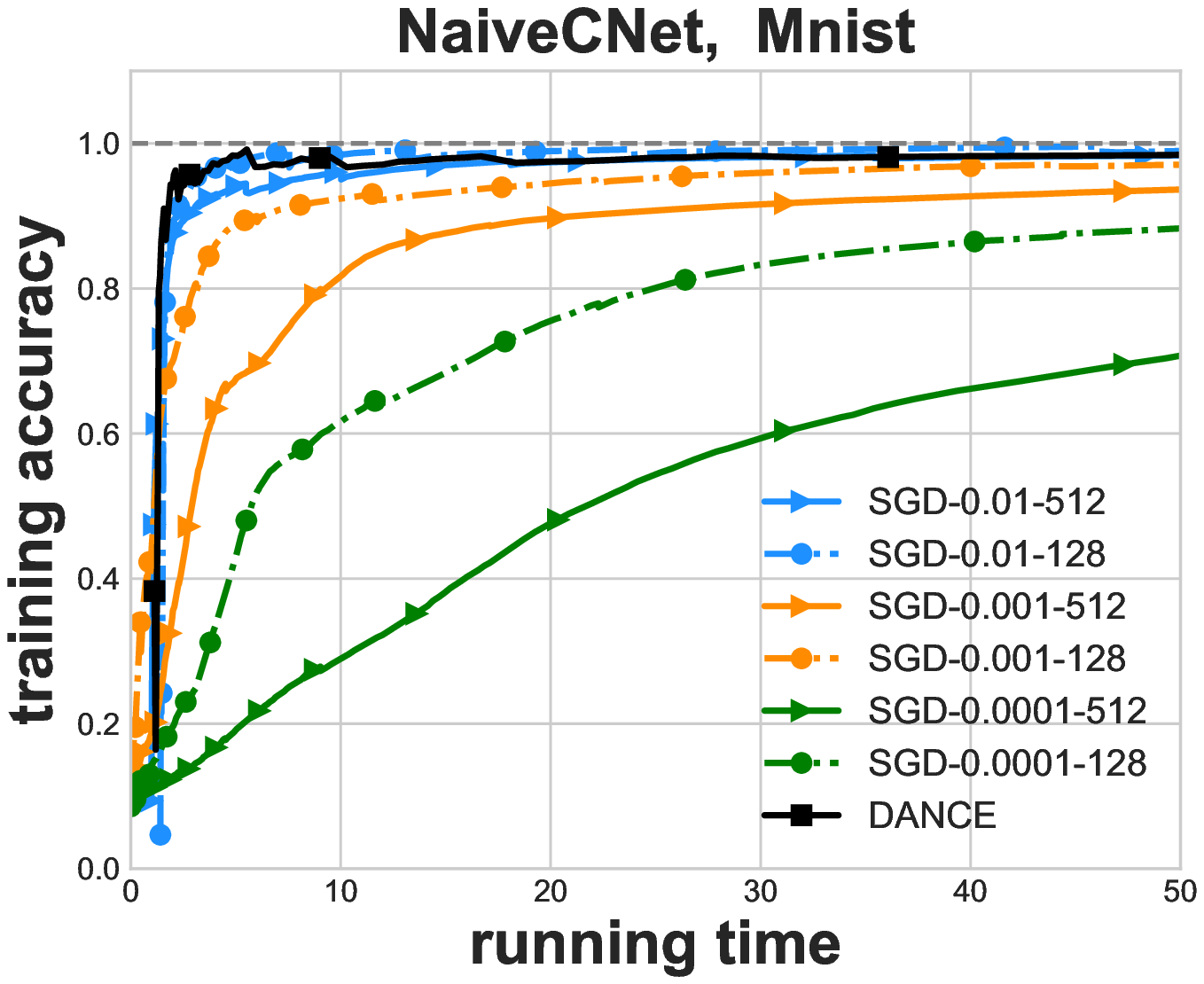}
	\includegraphics[width=0.30\textwidth]{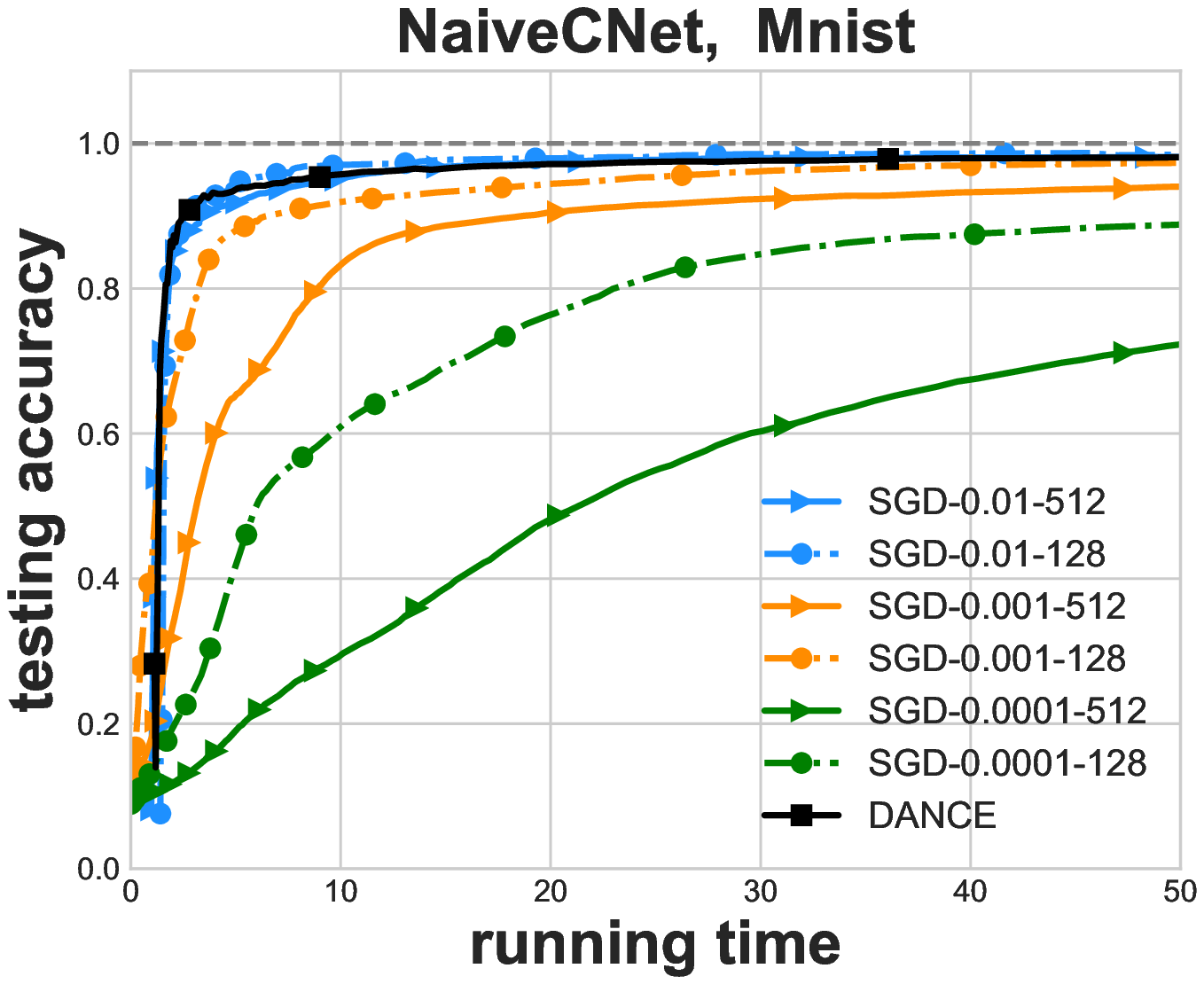}
	\caption{Comparison between DANCE and SGD with various hyper-parameters on Mnist dataset and NaiveCNet. NaiveCNet is a basic CNN with 2 convolution layers and 2 max-pool layers (see details at Appendix~\ref{sec: details concering experimental}). Figures on the top and bottom show how loss values, training  accuracy and test accuracy are changing with respect to epochs and running time. We force two algorithms to restart (double training sample size) after achieving the following number of epochs: $0.075, 0.2, 0.6. 1.6, 4.8, 9.6, 18, 36, 72$.  For SGD, we varies learning rate from $0.01, 0.001, 0.0001$ and batchsize from $128, 512$. One can observe that SGD is sensitive to hyper-parameter settings, while DANCE has few parameters to tune but still shows competitive performance. }
\end{figure*}



\begin{figure*}[t!]
	\centering
	\includegraphics[width=0.32\textwidth]{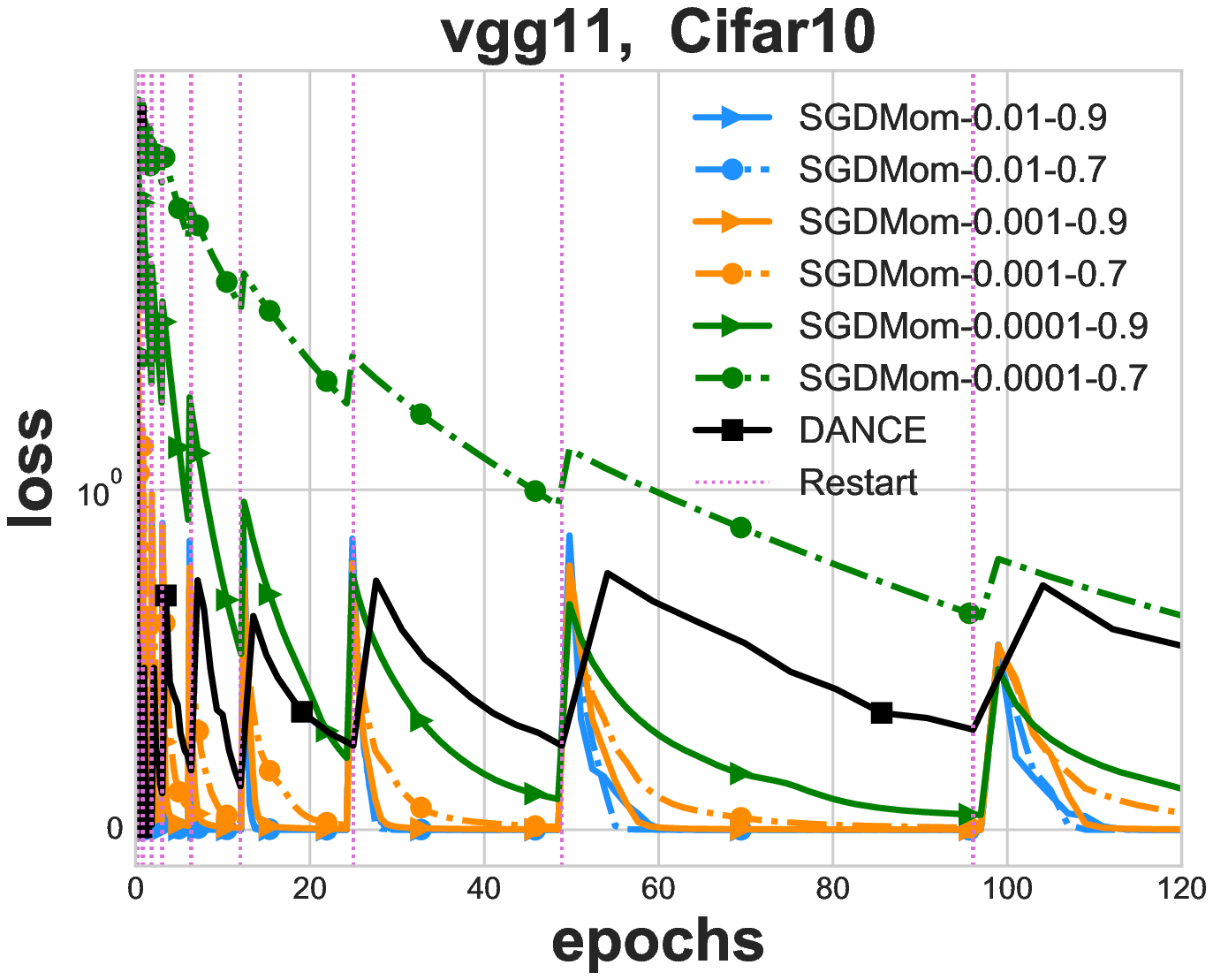}
	\includegraphics[width=0.32\textwidth]{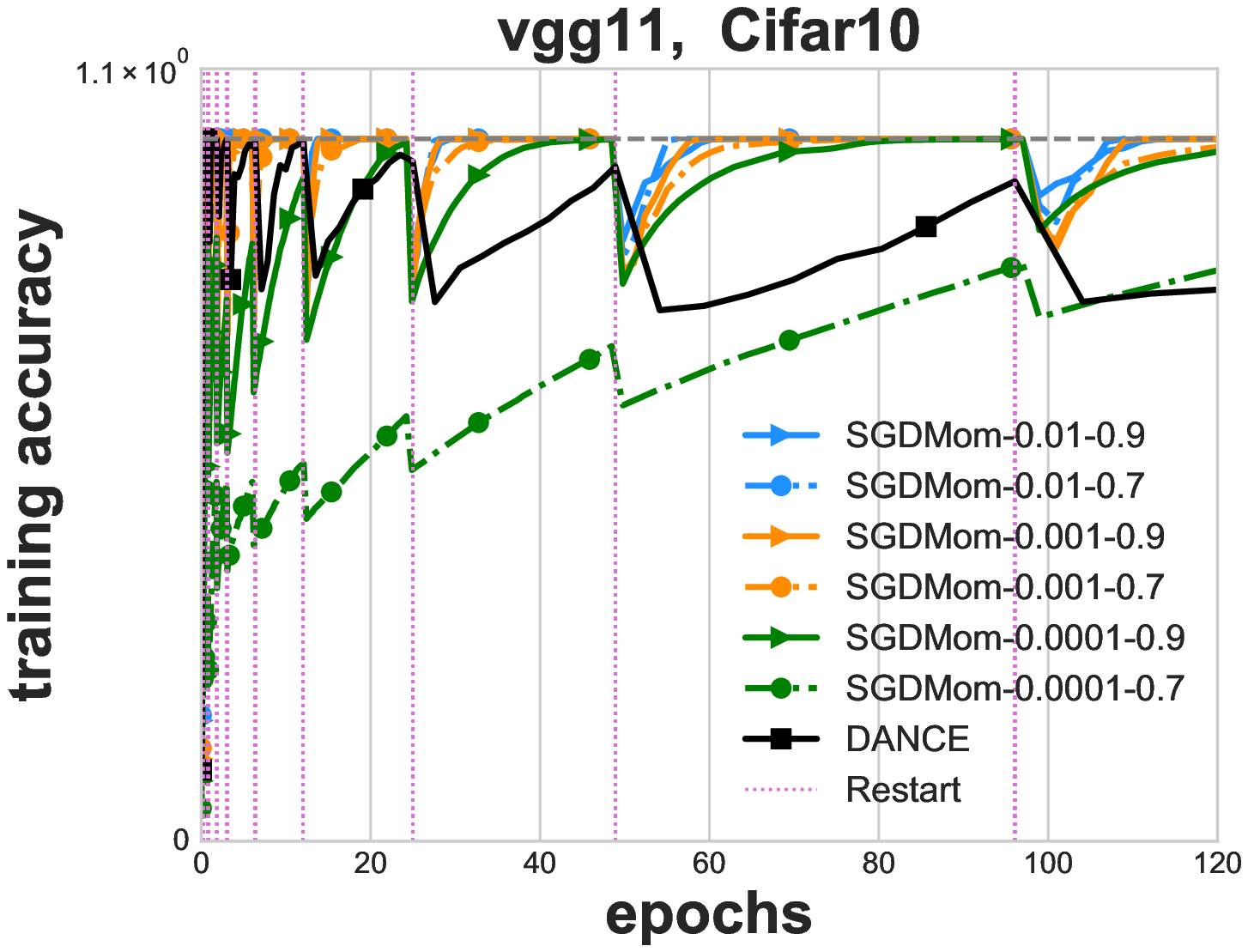}
	\includegraphics[width=0.32\textwidth]{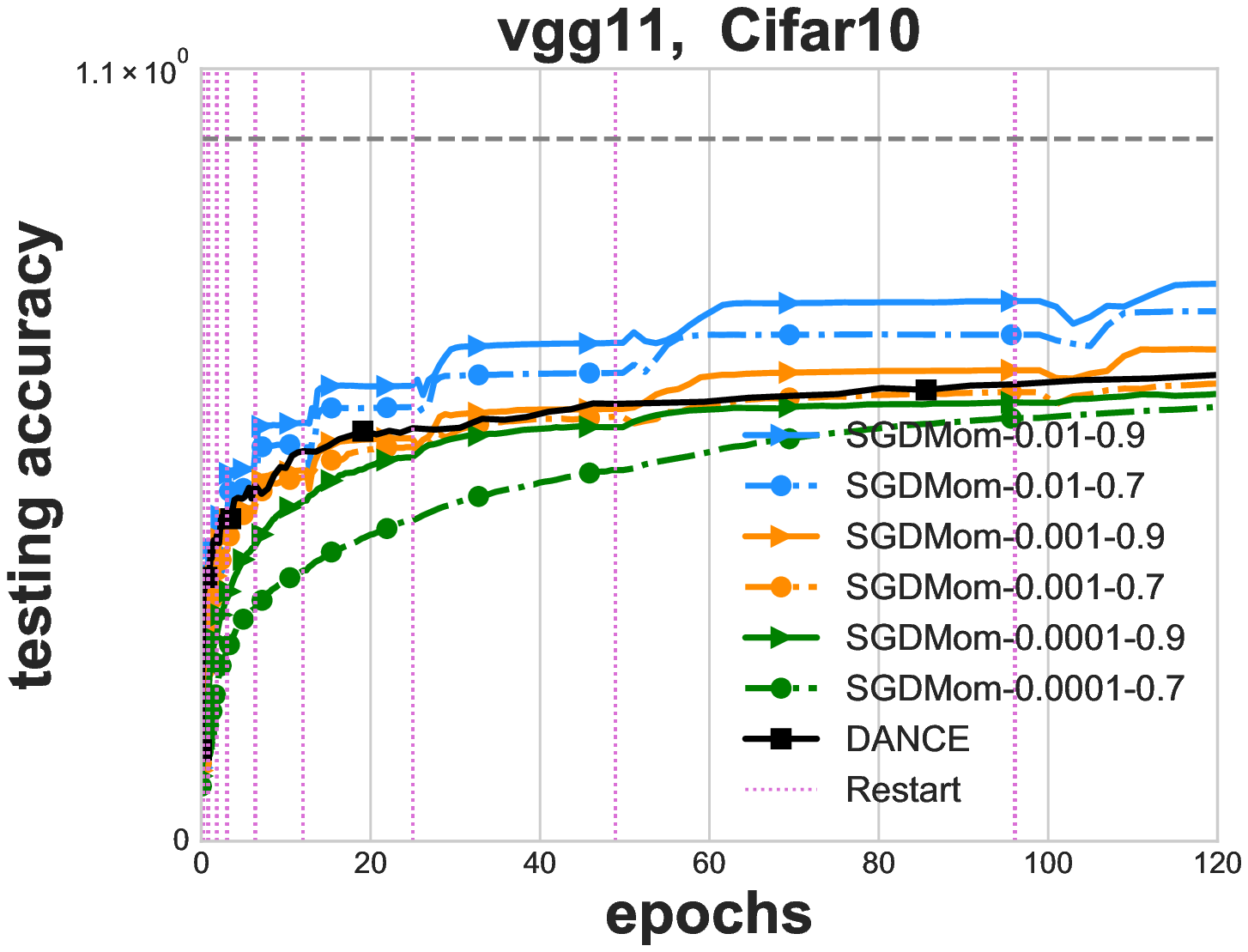}
	
	\includegraphics[width=0.32\textwidth]{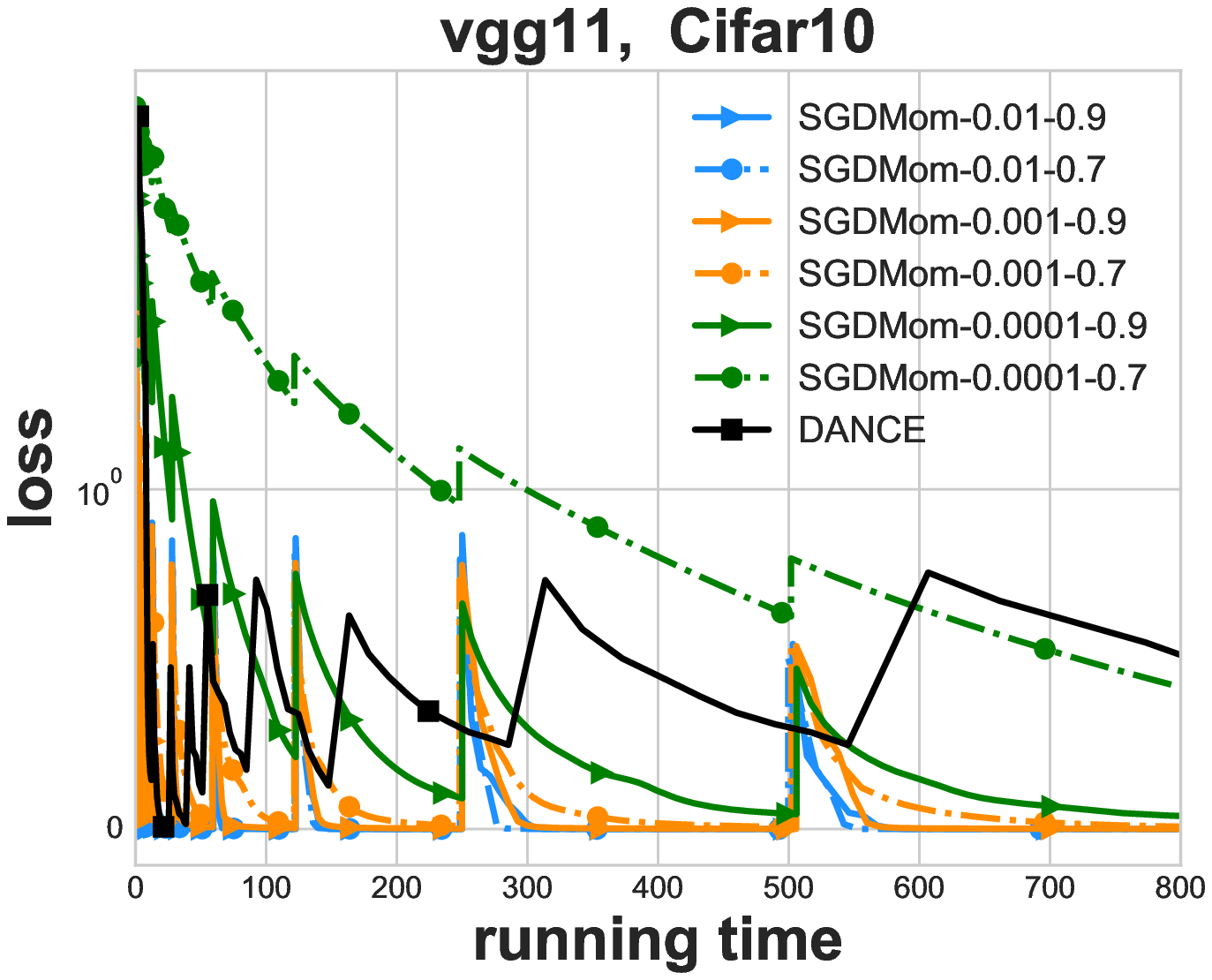}
	\includegraphics[width=0.32\textwidth]{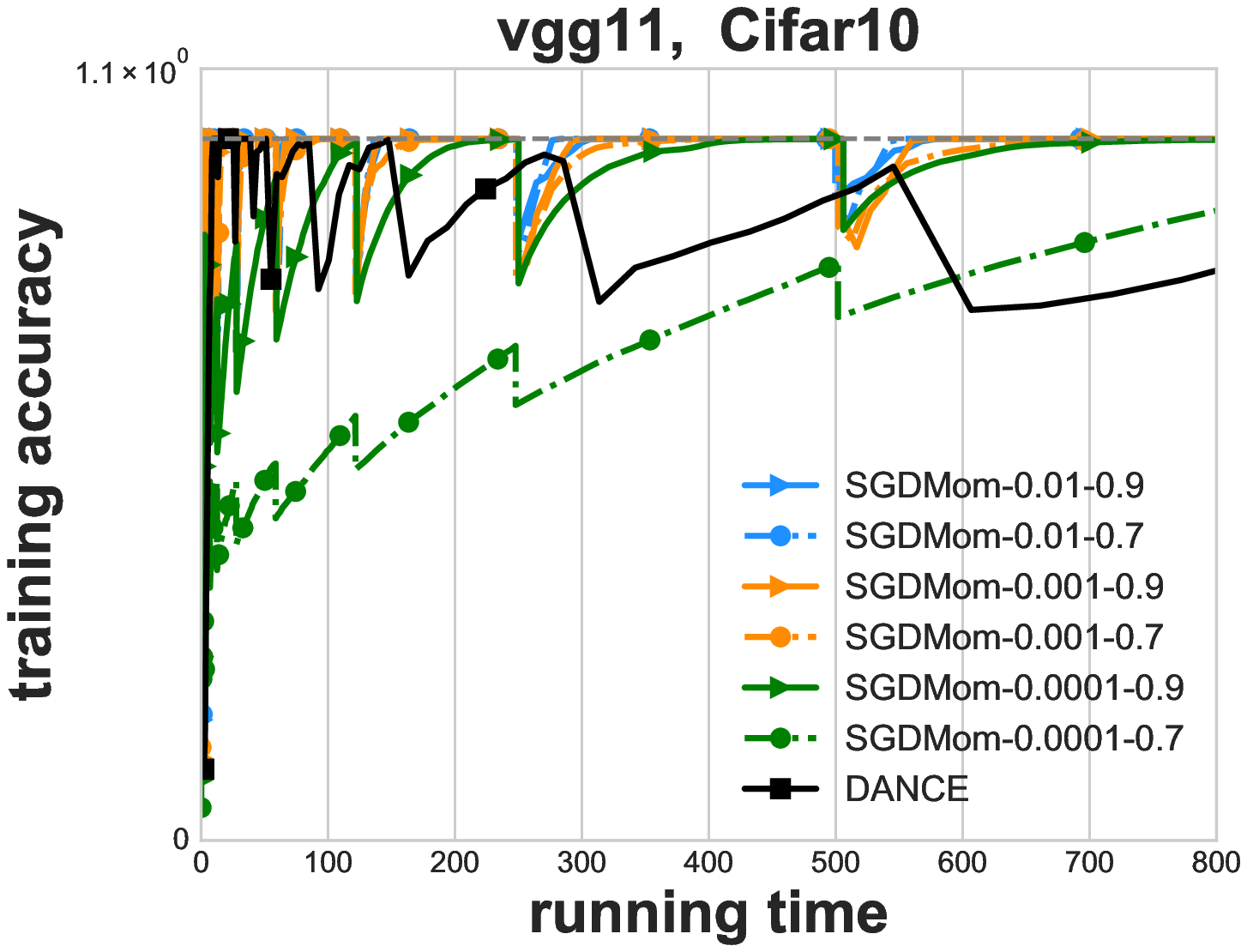}
	\includegraphics[width=0.32\textwidth]{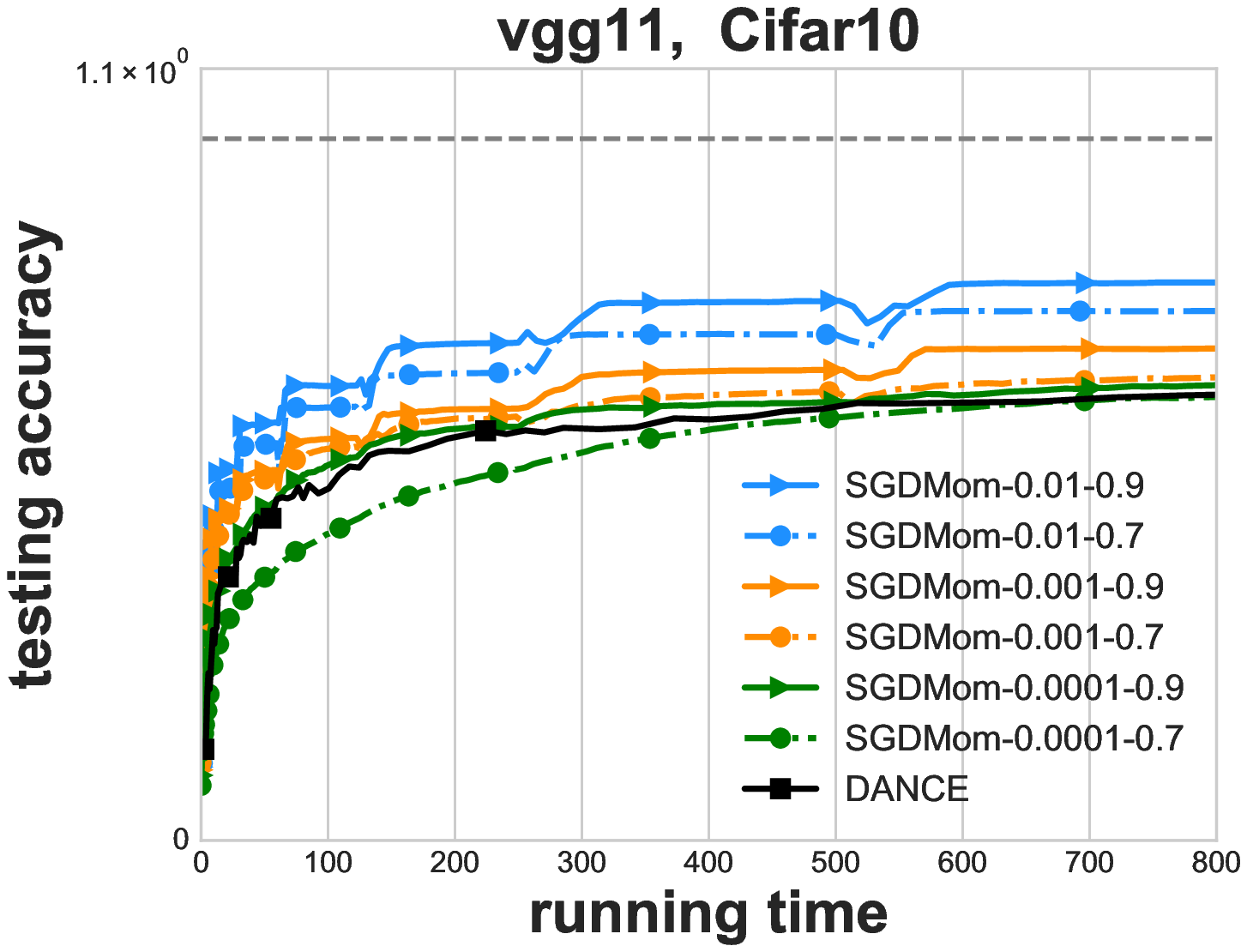}
	\caption{Comparison between DANCE and with momentum for various hyper-parameters on Cifar10 dataset and vgg11 network. Figures on the top and bottom show how loss values, training  accuracy and test accuracy are changing regarding epochs and running time, respectively. We force two algorithms to restart (double training sample size) after running the following number of epochs: $0.2, 0.8, 1.6. 3.2, 6.4, 12, 24, 48, 96$.  For SGD with momentum, we fix the batchsize to be $256$ and varies learning rate from $0.01, 0.001, 0.0001$ and momentum parameter from $0.7, 0.9$. One can observe that SGD with momentum is sensitive to hyper-parameter settings, while DANCE has few hyper-parameters to tune but still shows competitive performance. }
	\label{fig:vgg sgdmom}
\end{figure*}



\makeatletter
\setlength{\@fptop}{0pt}
\makeatother

\begin{figure*}[t!]
	\centering
	\includegraphics[width=0.30\textwidth]{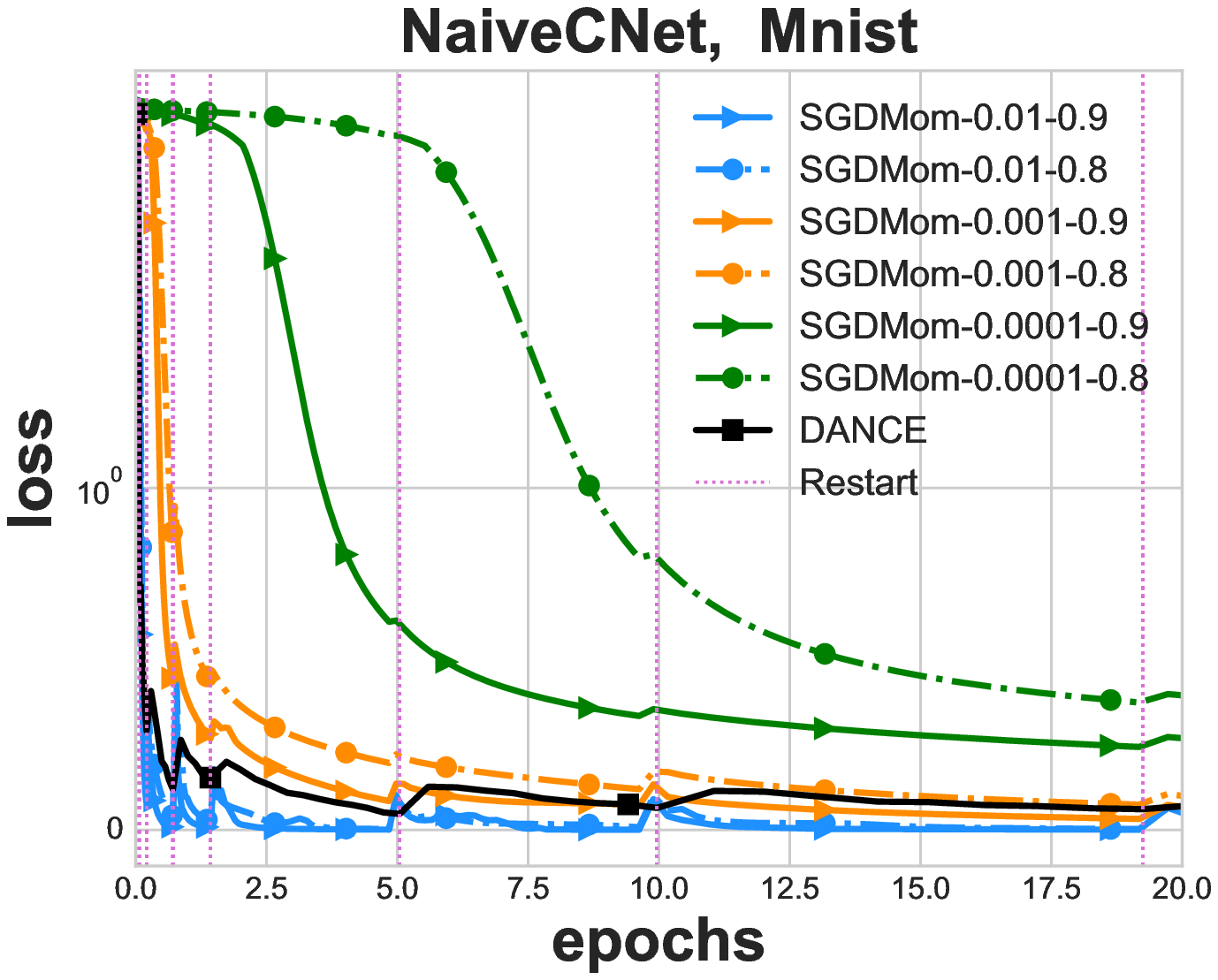}
	\includegraphics[width=0.30\textwidth]{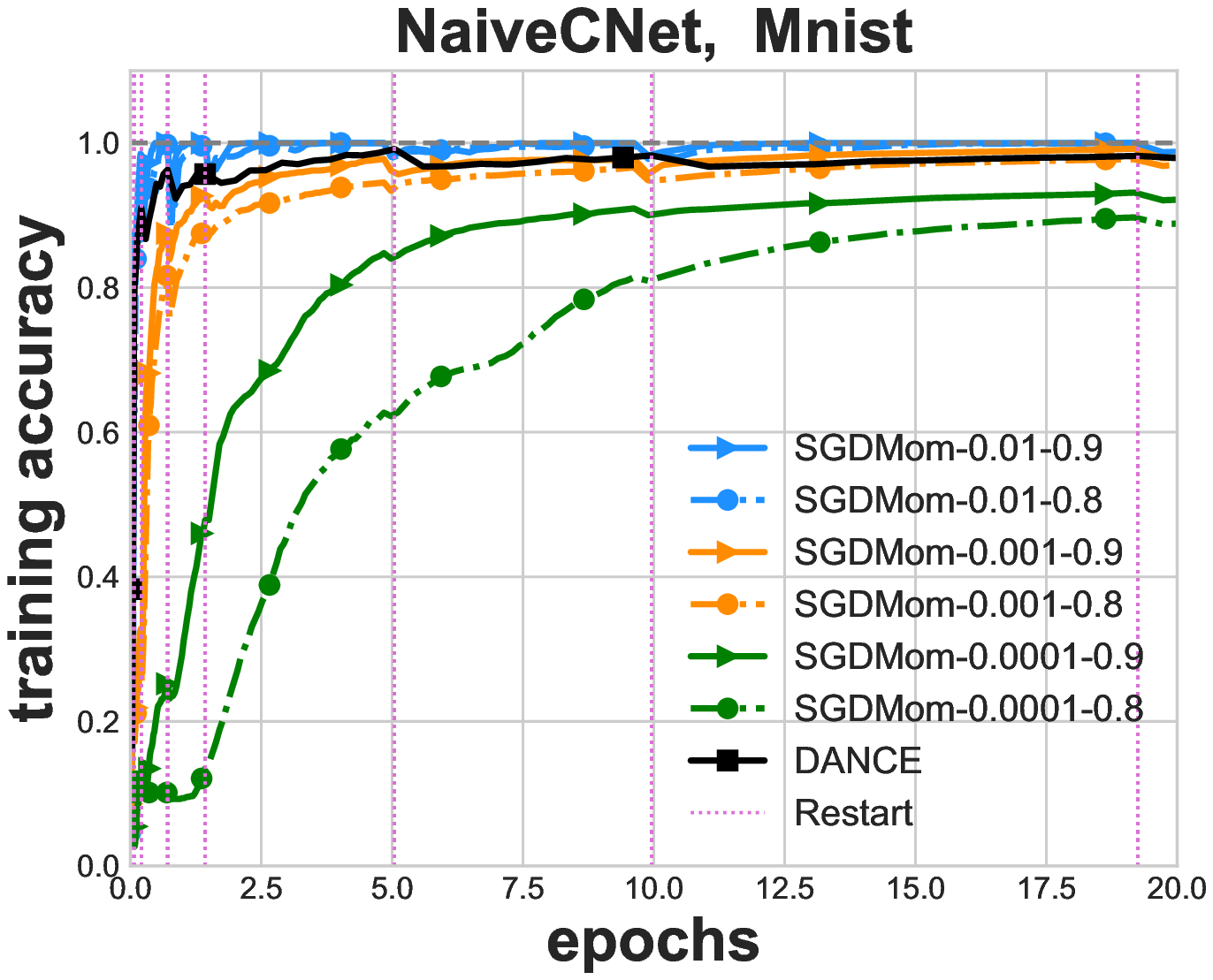}
	\includegraphics[width=0.30\textwidth]{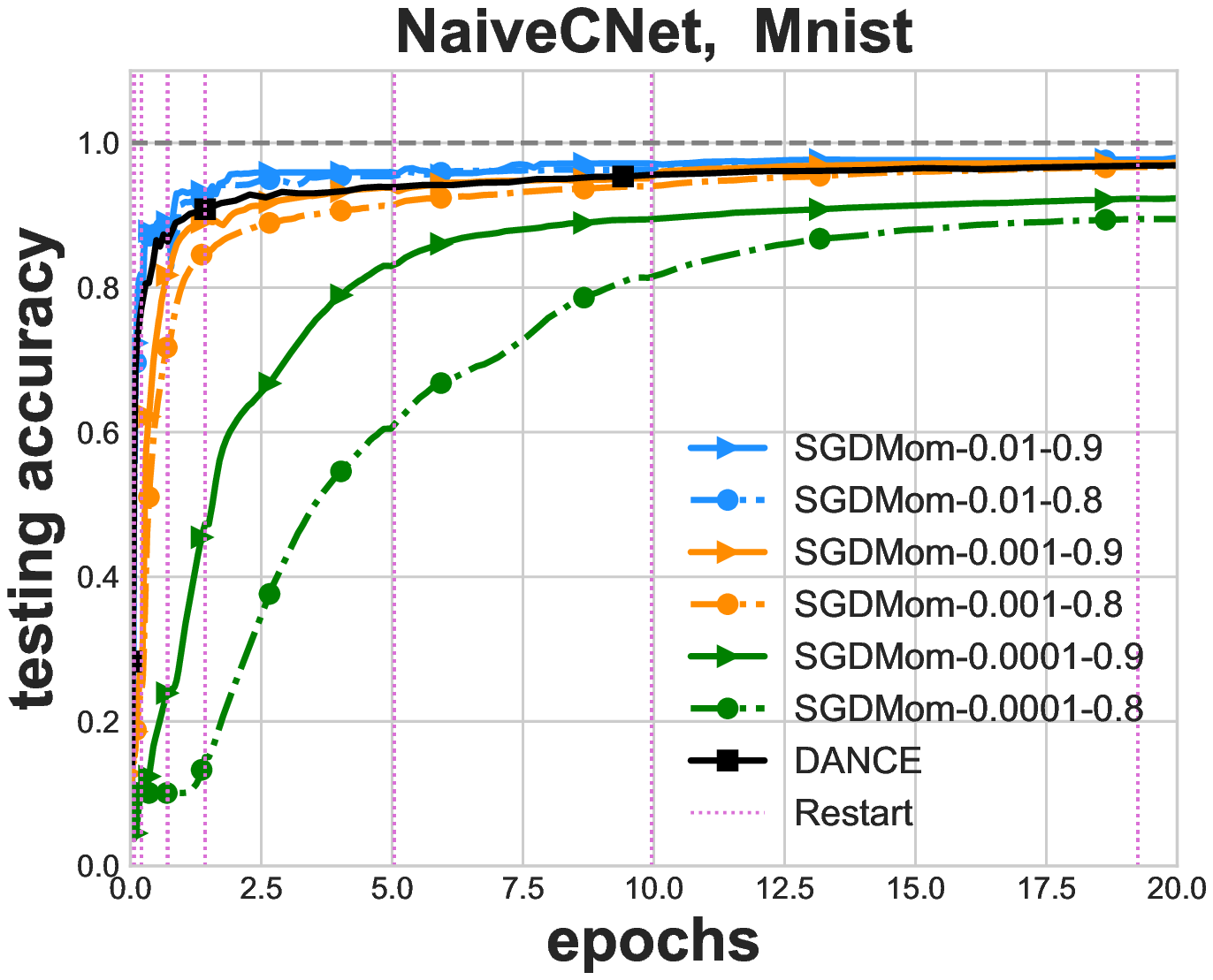}
	
	\includegraphics[width=0.30\textwidth]{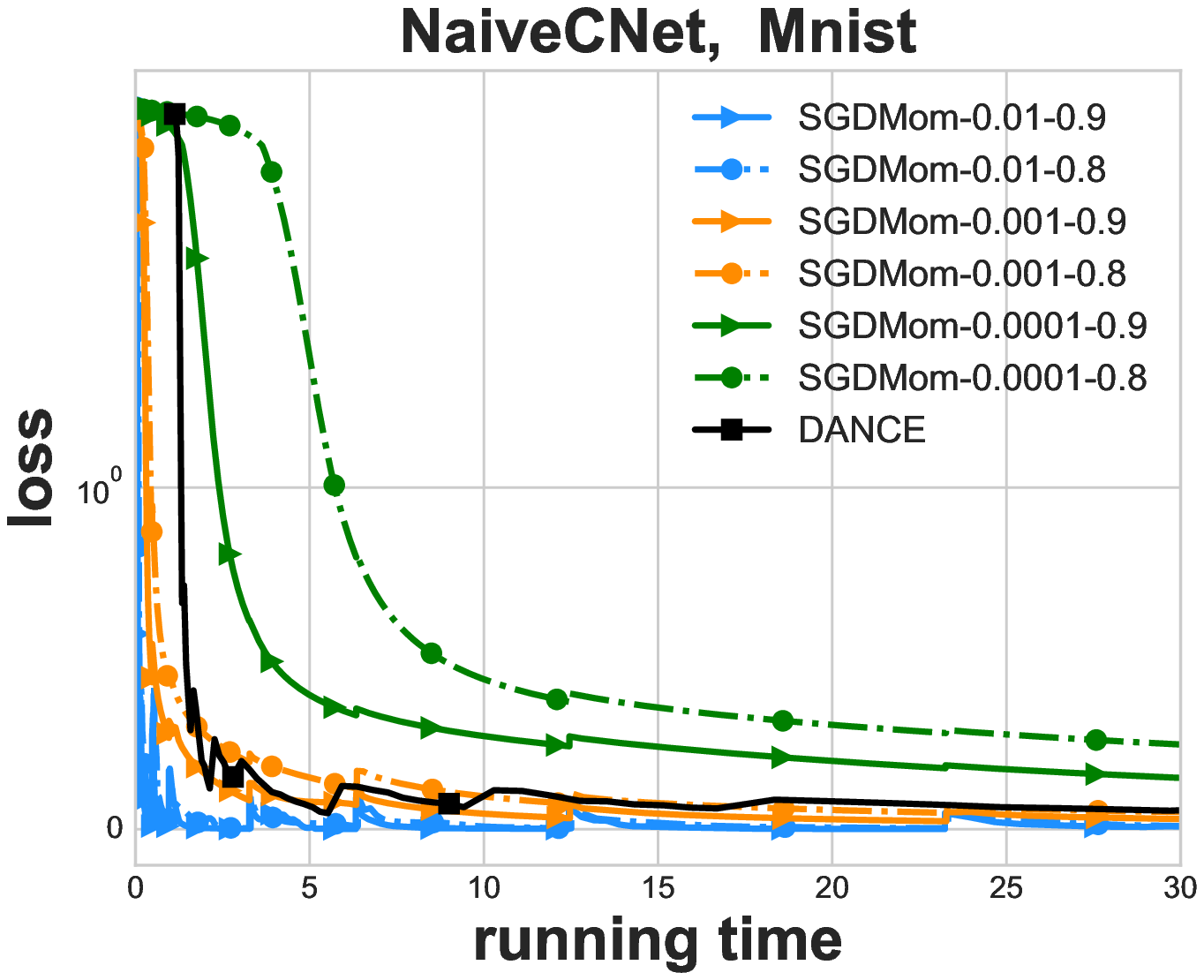}
	\includegraphics[width=0.30\textwidth]{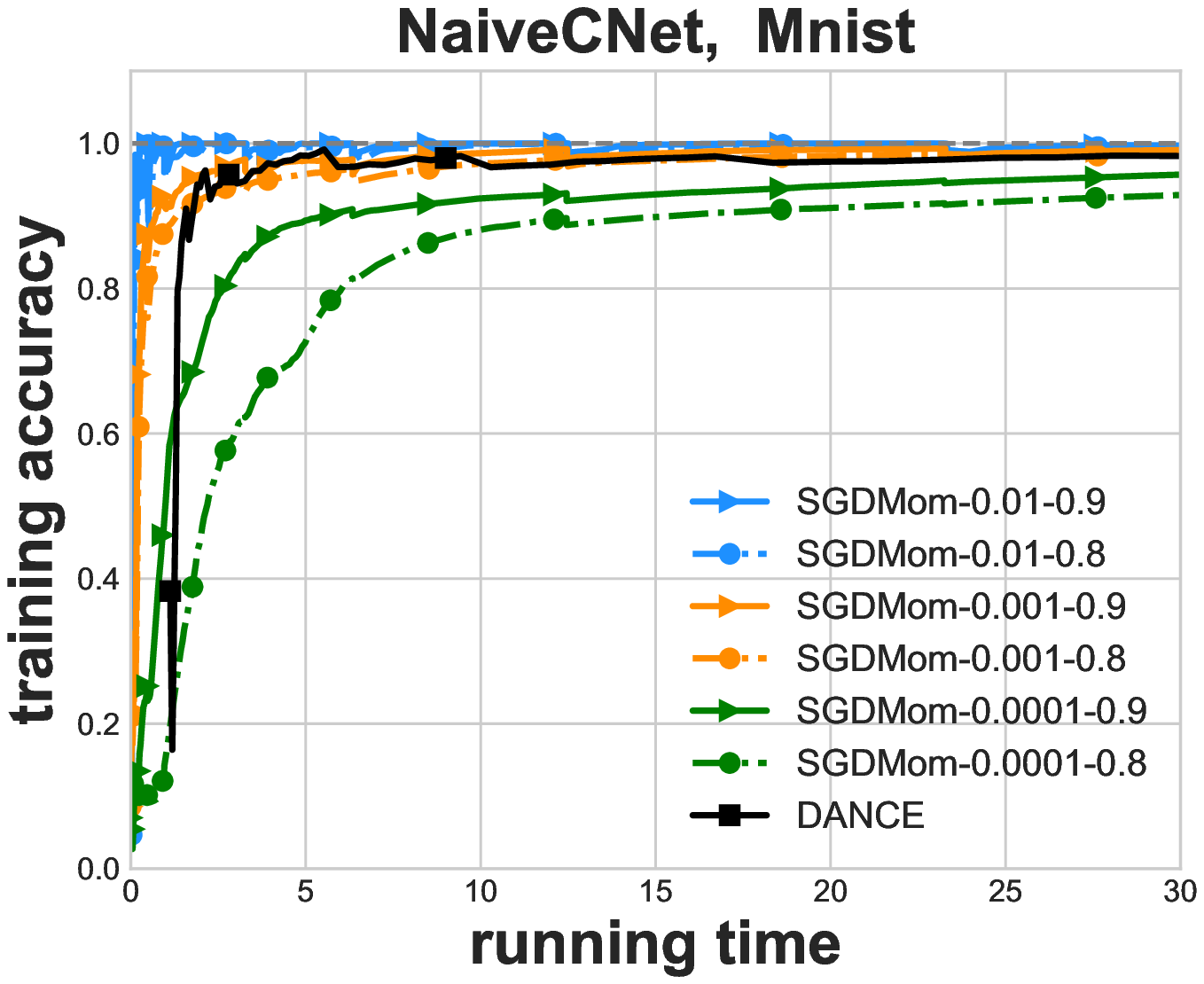}
	\includegraphics[width=0.30\textwidth]{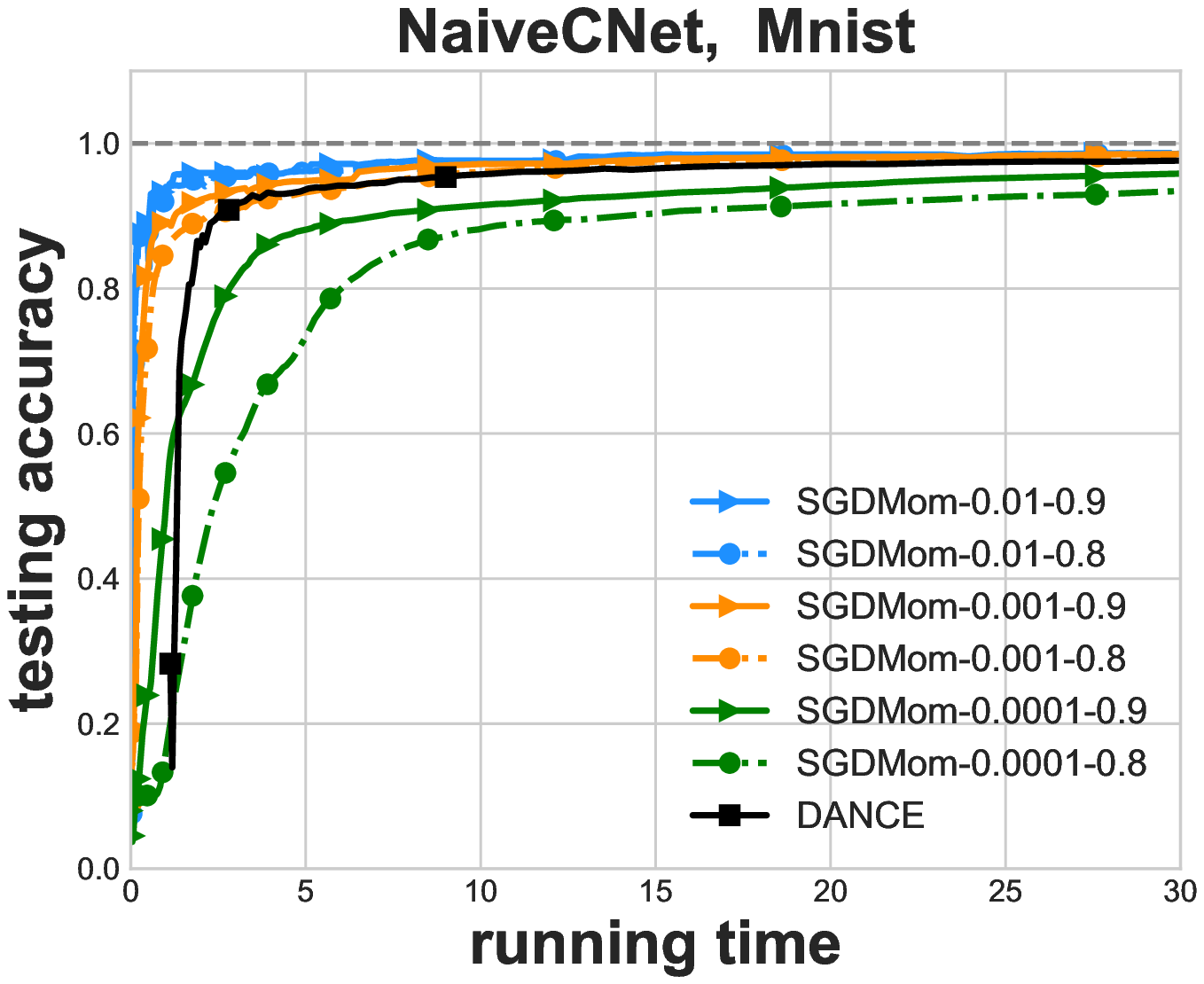}
	\caption{Comparison between DANCE and SGD with momentum for various hyper-parameters on Mnist dataset and NaiveCNet. Figures on the top and bottom show how loss values, training  accuracy and test accuracy are changing regarding epochs and running time, respectively. We force two algorithms to restart (double training sample size) after running the following number of epochs: $0.075, 0.2, 0.6. 1.6, 4.8, 9.6, 18, 36, 72$.  For SGD with momentum, we fix the batchsize to be $128$ and set learning rate to be $0.01, 0.001, 0.0001$ and momentum parameter to be $0.8, 0.9$. One could observe that SGD with momentum is sensitive to hyper-parameter settings, while DANCE has few hyper-parameters to tune but still shows competitive performance. }
	\label{fig:mnist sgdmom}
\end{figure*}

\end{document}